\documentclass[12pt]{article}
\usepackage{mathtools}
    \allowdisplaybreaks
\usepackage{amssymb}
\usepackage{bm}
\usepackage{physics}
\usepackage[mathscr]{euscript}
\usepackage{multirow}
\usepackage{enumitem}
\usepackage{xcolor}
\usepackage[colorlinks,citecolor=blue,linkcolor=blue,urlcolor=blue,pagebackref]{hyperref}
\usepackage{amsthm}
    \theoremstyle{plain} %
        \newtheorem{theorem}{Theorem}
        \newtheorem{corollary}{Corollary}
        \newtheorem{lemma}{Lemma}
        
    \theoremstyle{definition}
        \newtheorem{assumption}{Assumption}

    \theoremstyle{remark}
        \newtheorem{remark}{Remark}
\usepackage[margin=1in]{geometry}
\usepackage{booktabs}
\usepackage{microtype}
\usepackage{setspace}
\usepackage[ruled]{algorithm2e}
\usepackage[labelfont=bf, font=small]{caption}
\usepackage{natbib}
    \bibpunct[, ]{(}{)}{,}{a}{}{,}%

\DeclareMathOperator{\pr}{\mathbb{P}}
\DeclareMathOperator{\E}{\mathbb{E}}
\DeclareMathOperator{\Var}{Var}
\DeclareMathOperator{\Cov}{Cov}

\DeclareMathOperator{\ind}{\mathbb{I}}
\DeclareMathOperator*{\argmin}{argmin}
\DeclareMathOperator*{\argmax}{argmax}
\def\Real{\mathbb{R}}

\def\calA{\mathcal{A}}

\def\calN{\mathcal{N}}

\def\calS{\mathcal{S}}

\def\calW{\mathcal{W}}

\def\SFA{\mathsf{A}}

\def\SFC{\mathsf{C}}

\def\SFI{\mathsf{I}}

\def\SFP{\mathsf{P}}

\def\SFQ{\mathsf{Q}}

\def\SFR{\mathsf{R}}

\def\SFX{\mathsf{X}}

\DeclareMathOperator{\plim}{\mathrm{plim}}

\begin{document}

\begin{titlepage}
\title{Asymptotic Theory for IV-Based Reinforcement Learning with Potential Endogeneity\thanks{Contact information: Jin Li (\url{jli1@hku.hk}), Ye Luo (\url{kurtluo@hku.hk}), Faculty of Business and Economics, The University of Hong Kong, Pokfulam Road, Hong Kong SAR, Co-Director of Center for Social Science, Shenzhen Loop Area Institute;  Zigan Wang (\url{wangzigan@zju.edu.cn}), School of Economics, Zhejiang University, China;
Xiaowei Zhang (\url{xiaoweiz@ust.hk}), Department of Industrial Engineering and Decision Analytics, The Hong Kong University of Science and Technology, Clear Water Bay, Hong Kong SAR.
We thank
Chunrong Ai, Jose Blanchet, Nan Chen, Xinyun Chen, Victor Chernozhukov, Jim Dai, Yifan Feng, Ivan Fernandez-Val, Jean-Jacques Forneron,  Kay Giesecke, Peter W. Glynn, Wei Jiang, Hiroaki Kaido, Tom Luo, as well as seminar participants at Boston University, Chinese University of Hong Kong, Peking University, Shenzhen Research Institute of Big Data, Stanford University, and Mostly OM 2024 Workshop
for helpful discussions. We thank Mengxin Yang for her excellent research assistance. We thank for funding support from NSFC-RGC  N\_HKU7115/22 (T2261160400) and TRS T32-615\_24-R. All remaining errors are ours.}}
\author{Jin Li \and Ye Luo \and Zigan Wang \and Xiaowei Zhang}
\date{}
\maketitle
\vspace{-10pt}
\begin{abstract}
In the standard data analysis framework, data is collected (once and for all), and then data analysis is carried out. However, with the advancement of digital technology, decision-makers constantly analyze past data and generate new data through their decisions. We model this as a Markov decision process and show that the dynamic interaction between data generation and data analysis leads to a new type of bias---reinforcement bias---that exacerbates the endogeneity problem in standard data analysis.
We propose a class of instrument variable (IV)-based reinforcement learning (RL) algorithms to correct for the bias and establish their theoretical properties by incorporating them into a stochastic approximation framework. Our analysis accommodates iterate-dependent Markovian structures and, therefore, can be used to study RL algorithms with policy improvement.
We also provide formulas for inference on optimal policies of the IV-RL algorithms.
These formulas highlight how
intertemporal dependence of the Markovian environment affects the inference.

\medskip
\noindent\textbf{Keywords:} Endogeneity,
Measurement Error,
Markov Decision Process,
Instrumental Variable,
Reinforcement Bias,
Reinforcement Learning,
Q-learning,
Actor-Critic,
Stochastic Approximation

\medskip
\noindent\textbf{JEL Codes:} C10
\end{abstract}
\setcounter{page}{0}
\thispagestyle{empty}
\end{titlepage}
\pagebreak \newpage

\section{Introduction}\label{sec:intro}

With the advancement of digital technology, data-based management has become increasingly popular.
Human resource departments now use data analytics for recruitment, performance evaluation, promotion decisions, and terminations. Marketing teams optimize digital advertising campaigns using objective metrics like viewer counts and click-through rates. Operations managers leverage historical data to forecast demand, manage inventory, design worker incentives, and implement dynamic pricing strategies.

The typical data-based management paradigm assumes, implicitly, that the data is collected, once and for all, and then data analysis is carried out, and the decision is made. Increasingly, however, these decisions are made frequently, and they affect the generation of future data. When data generation and data analysis interact dynamically, how are the results of the data analysis affected in the long run? In particular, if the data collected is imperfect and contains various types of biases, e.g., omitted variable, measurement error, model misspecification, selection, and simultaneous equations, how are the magnitude of the biases affected by the dynamic interaction? And what can be done to correct these biases?

These challenges are increasingly relevant as algorithms begin to govern workplace and social interactions. Consider online delivery platforms such as Meituan and Uber Eats, where algorithms assist managers in determining appropriate incentive schemes for riders. The primary objective for these companies is to select an incentive structure that maximizes long-term value. However, measuring this long-term value poses significant difficulties, as many benefits—and costs—are realized in the future. Consequently, companies rely on an imperfect set of metrics, typically encompassing key performance indicators (KPIs) such as delivery time, short-term profit, gross merchandise value, and customer satisfaction ratings. Based on these KPIs, managers decide the extent to which rider compensation is linked to performance metrics.

A frequently discussed issue in this context is the potential for gaming performance metrics \citep{HolmstromMilgrom91}. Workers may engage in behaviors that artificially enhance their performance indicators. Thus, the performance metrics not only reflect an imperfect measure of true latent reward (latent reward later) of the firm due to classic measurement-error problems but also because riders have the capacity to manipulate outcomes. More critically, the extent to which riders can influence these metrics is directly related to the strength of the incentives provided.

For instance, when riders receive a strong incentive to perform well, they may encourage customers to provide favorable ratings. In such cases, while customers may leave positive feedback, their actual satisfaction with the service could be low, potentially leading to decreased platform usage over time. Generally speaking, stronger incentives push riders to take actions that can create negative externalities, thus diminishing the firm's overall cumulative value.

In summary, heightened incentive strength exacerbates the measurement problem by widening the gap between measured performance and true performance. This misalignment means that the measured performance consists of the firm's latent reward plus an error term that varies with the chosen incentive level. This scenario introduces a novel form of measurement challenge, further complicated by the fact that incentive strength is frequently adjusted based on historical performance metrics.

Specifically, consider a scenario where the manager implements such an optimal incentive policy that increases with the platform's monthly active users (MAUs). When the manager performs a regression analysis of KPIs against incentive levels using historical data, the estimated effectiveness of incentives shows an upward bias, possibly because of rider's incentive to influence the performance metric.

This upward bias leads the manager to conclude that the current incentive policy is suboptimal, prompting an increase in per-rider incentives.
Under this updated policy, new data is generated and analyzed.
The analysis of these new data may again suggest that the updated policy is suboptimal, triggering further policy adjustments.

Note that the manager serves dual roles as both data analyst and data generator.
As an analyst, she estimates the effectiveness of platform incentives using historical data.
As a generator, she uses these estimates to determine future incentive levels.
This process continues \emph{ad infinitum} and stops only when the estimated effectiveness parameter becomes consistent with the parameter guiding the incentive policy, reaching a fixed point.

Being a fixed point, however, is not the same as being the actual parameter. This paper demonstrates that additional biases arise when data generation and analysis interact. The additional bias arises because the biases from the previous analysis affect the new data generated and can amplify the original bias in a fashion similar to the multiplier effect in macroeconomics. We denote such bias as the \emph{reinforcement bias} (R-bias).

To study this new bias and how to correct it, we consider a Markov decision process (MDP) in which an agent makes a sequence of decisions over time to maximize her discounted rewards.
In each period, the \emph{latent} reward is an unknown function of an observable current state variable and the agent's decision.
However, the \emph{observed} proxy reward is mis-measured and potentially biased. Based on the biased data, the agent makes decisions over time. Our findings are summarized as follows.

\subsection{Main Contributions}

First, we demonstrate the existence of the R-bias. We analyze an example where the state variables are Markovian, and
the noise (in the observed reward) is correlated with the state variable or action. We calculate how the R-bias arise because of the interaction between learning and decision-making and illustrate how the Markovian structure amplifies the bias.

Second, we show how to correct this bias using an instrumental variable (IV) approach. We introduce IV-based reinforcement learning (RL) algorithms, IV-Q-learning and IV-AC, based respectively on two leading RL algorithms: Q-learning and Actor-Critic.
Our use of IVs enables the agent to \emph{causally explore} in her decisions.
The notion of exploration in the conventional sense in RL algorithms helps the agent better learn about the rewards by choosing seemingly inferior actions.
But when observed  proxy rewards are mis-measured and biased, knowledge from these explorations is also biased, leading to suboptimal policies. By contrast, by perturbing the data-generating process, our IV-based approach helps the agent mitigate measurement issue and learn optimal policies.

Our third (and technical) contribution is to analyze IV-RL algorithms through the lens of stochastic approximation (SA)
and establish general results concerning asymptotic performances of SA.
In contrast to RL literature that usually
assumes independent observations to exploit a martingale-difference structure in the SA formulation, we study SA with iterate-dependent Markovian structures (i.e., the underlying Markov chain has a transition distribution depending on the current SA iterate). The Markovian aspect is essential for many practical applications since the transition of state variables (e.g., the MAU in the incentive scheme design example) is typically better described as Markovian.
And the iterate-dependent aspect is key to accommodating RL algorithms that feature policy improvement: state transitions depend on the policy used.
We establish a central limit theorem (CLT) for SA under primitive assumptions.
Applying the general result, we then develop inference methods for optimal policies of IV-RL algorithms.
Our results highlight how intertemporal dependence in the Markovian environment affects the inference.

\subsection{Related Works}

Our paper contributes to methodological advances in RL.
For a general introduction to RL,
see \cite{SuttonBarto18}; for  applications in economics and finance, see \cite{Charpentier23} and \cite{Bai24}.
Despite extensive theoretical analysis of RL algorithms in recent years (see, e.g., \cite{BhandariRussoSingal21} and references therein), statistical inference on optimal policies remains significantly underexplored in the RL literature. We address this gap by establishing CLTs for the IV-RL algorithms.

A closely related methodological literature studies nonparametric
SA under feedback and dependent data.
\citet{ChenWhite98} formulate adaptive learning with feedback as
Hilbert-valued SA and establish almost-sure
convergence under near-epoch dependence. \citet{ChenWhite02} study
 Robbins--Monro procedures on finite-dimensional
subspaces of increasing dimension and derive almost-sure convergence,
a functional CLT, and convergence-rate results under
Hilbert-valued weakly dependent errors. 
Our setting differs because policy updates change the Markov
transition kernel governing future data. We derive a CLT under primitive conditions on these iterate-dependent kernels
and use it to conduct inference on policies learned by IV-RL algorithms.

Our problem setting where adaptive data collection may involve biases from omitted variables, measurement error, and other sources can be viewed through the lens of partially observable MDPs (POMDPs), in which state variables are only observed through noisy proxies.
Although POMDPs can be converted to MDPs by introducing belief states to capture historical information about the original state variables, solving these resulting MDPs remains challenging due to the curse of dimensionality.
For an introduction to POMDPs, see \cite{Krishnamurthy16}, and for recent advances in RL algorithms for POMDPs, see \cite{Subramanian22approximate} and references therein.
Our paper potentially offers an alternative approach to addressing such problems.

Our paper also contributes to the literature on IV methods.\footnote{This is a vast literature, and we refer to \cite{Hausman83}, \cite{Imbens94}, and \citet{Chen05} for theoretical analyses of IVs in linear models, and refer to \cite{Hansen82}, \cite{Ai03}, \cite{Newey03}, and
\cite{Chernozhukov07} for the use of IVs in nonlinear models. See \cite{Imbens14} for a
broader review of identification and IV methods.} 
Most studies in this literature focus on the causal effect of a single
decision.
Few use IVs to learn the causal effect of a policy in RL
settings\footnote{Economists have recently been interested in the use of RL in business decision processes; see, e.g.,
\cite{BlakeNoskoTadelis15,CalvanoCalzolariDenicoloPastorello20,CowgillTucker20,LiRaymondBergman26} and \cite{JohnsonRhodesWildenbeest23}.
}
or other sequential decision problems.\footnote{A separate approach to
causal RL uses directed acyclic graph (DAG) models. This approach
requires detailed knowledge of the causal links among states and
actions, which can be difficult to obtain in economic settings with
complex causal channels \citep{Imbens20}.}
\citet{Nambiar19} study a dynamic pricing problem with contextual
information, independent state variables, and a misspecified demand
model. They use randomized experiments to construct IVs and identify
optimal pricing policies. Our setting instead allows a general class of
IVs and Markovian state dynamics.
\citet{LiLuoZhang21_bandit} study contextual bandits with biased reward
observations and identify \emph{self-fulfilling bias}, a form of
selection bias created by interaction with the decision environment.
Because it arises from such interaction, self-fulfilling bias is closely
related to R-bias. Their correction uses IV-based bandit algorithms that
do not take the SA form studied here and therefore require different
analysis. Moreover, contextual-bandit states are not affected by
previous actions \citep[Ch.~2]{SuttonBarto18}. We instead study the
intertemporal dependence introduced by Markovian state transitions and how this dependence
affects R-bias and inference for IV-based RL algorithms.

Related work also studies recursive IV and GMM estimation.
\citet{Bradtke96} propose a least-squares IV approach to
temporal-difference learning, a standard RL algorithm for policy evaluation, and provide convergence results.
\citet{Chen23} introduce stochastic generalized method of moments, an SA method for overidentified moment
models with streaming data. They establish almost-sure convergence and
functional CLTs for online 2SLS and efficient SGMM.
Our paper instead integrates recursive IV estimation with policy
learning and inference in Markovian environments.

Our paper also relates to the emerging literature on policy evaluation in adaptive experiments, e.g., through contextual bandit algorithms \citep{Bottou_etal13,HadadHirshbergZhanWagerAthey21,ZhanHadadHirshbergAthey21}. This literature deals with a related bias due to insufficient sampling of certain actions in an adaptive data collection process  \citep{ShinRamdasRinaldo19}. Unlike the present paper, however,
this bias is not caused by biased reward observations and is fixed differently, for example, by inverse propensity weighting. Another key difference is that this literature focuses on offline learning in which data is collected in advance and then fixed for subsequent analysis. By contrast, we consider an interactive decision-making environment where new data are continuously generated and optimal policies are learned in an online fashion.

The rest of the paper is organized as follows.
Section \ref{sec:RBias} presents an illustrative example of R-bias, followed by Section \ref{sec:RL} which introduces the IV-Q-learning algorithm. Section \ref{sec:analysis} embeds this algorithm within the SA framework and establishes its theoretical properties. Section \ref{sec:numerical} provides numerical evaluations of the IV-RL algorithms, Section \ref{sec:empapplication} introduces an empirical example of stock repurchase with analysis using IV-Q-learning, and Section \ref{sec:conclusion} concludes.
The IV-AC algorithm and its theoretical analysis appear in the Appendix.
The main theorems' proofs are provided in the Online Supplemental Material, while proofs of auxiliary results are available upon request.

\section{Reinforcement Bias}\label{sec:RBias}

\subsection{Problem Setup}\label{subsec:framework}
An agent is engaged in an infinite-horizon MDP with state space $\calS$, action space $\calA$, and a discount factor $\gamma\in[0,1)$.\footnote{We treat $\gamma$ as a fixed parameter chosen by the user rather than estimated from the data.
This treatment is standard in RL, where \(\gamma\) is part of the learning objective  \citep{SuttonBarto18}, and in structural econometrics, where researchers typically fix the discount factor because it is generally not identified from choice data without additional exclusion restrictions
\citep{Rust87}. In applications, $\gamma$ should reflect both the length of the decision period and the target discount rate at that frequency.
In our empirical application, we set \(\gamma=0.95\) and show that the results are stable across values ranging from \(0.85\) to \(0.99\).}
At each time $t=0,1,\ldots$, the agent observes the current state $S_t =s\in\calS$ of the environment and follows a policy $\pi$ to take an action $A_t=a\in\calA$.
In general, the policy can be random; that is, for each $s\in\calS$, $\pi(\cdot|s)$ is a probability distribution on $\calA$ from which $a$ is sampled.
Following the action $A_t=a$, a reward $r(s,a)\in \Real$ is realized, and the state transitions to $S_{t+1} =s'$ with probability $\SFP(s'|s, a)\coloneqq \pr(S_{t+1} = s'|S_t=s, A_t=a)$.
Given an initial state $s$,
the goal of
the agent is to learn an optimal policy $\pi^*$ that maximizes the expected cumulative discounted reward:
\begin{equation}\label{eq:Case2}
\max_{\pi} \E_{A_t\sim\pi(\cdot|S_t):t\geq 0}\biggl[\sum_{t=0}^\infty \gamma^t r(S_t, A_t) \Big| S_0 = s \biggr].
\end{equation}

For a policy $\pi$, we define the \emph{state-action value function}, i.e., the ``Q-function'', i.e., as
the expected cumulative discounted reward if the initial state-action pair is $(s,a)$ and the agent follows policy $\pi$ afterward:
\[
Q^\pi(s,a) \coloneqq \E_{A_t\sim\pi(\cdot|S_t):t\geq 1}\biggl[\sum_{t=0}^\infty \gamma^t r(S_t, A_t)\Big|S_0=s,A_0=a\biggr].
\]
For simplicity,
we write the optimal Q-function (i.e., the Q-function associated with $\pi^*$) as $Q^*(s,a)$.
It is well known that $Q^*$ satisfies the Bellman optimality equation
\begin{align}\label{eq:bellman}
    Q^*(s,a) =  r(s,a)+\gamma \E \Bigl[\max_{a'\in\mathcal{A}} Q^*(S',a')\Bigr],
\end{align}
where the expectation is taken with respect to $S' \sim \SFP(\cdot|s,a)$.
Given $Q^*$, an optimal policy can be calculated via $\pi^*(s)=\argmax_{a\in\mathcal{A}} Q^*(s,a)$ for all $s\in \mathcal{S}$.

The term ``model'' is typically used to refer to the state transition distribution $\mathsf{P}(\cdot|s,a)$ and the reward function $r(s,a)$.
If both are known, one can directly solve the Bellman optimality equation (at least numerically, in principle) for the optimal policy via dynamic programming techniques, including value iteration and policy iteration \citep{BertsekasDP}.
This dynamic programming method is known as the ``model-based'' approach. However, in complex environments, the state transition distribution or the reward function are typically unknown, making dynamic programming method infeasible. One then resorts to RL methods, which are ``model-free.'' RL methods allow the agent to adjust her policies towards optimality without knowing the model specification  \citep{SuttonBarto18}.

In RL, the agent adjusts her policies based on observations of the state-action-reward tuple $(S_t, A_t, R_t)$, where $R_t \coloneqq r(S_t,A_t)+\epsilon_t$ is an observed proxy reward of the ``true" latent reward $r(S_t,A_t)$. Standard RL methods assume that $R_t$ is \emph{unbiased}, that is, $\E[\epsilon_t|S_t,A_t] = 0$. This assumption is important for the agent's policies to converge to the optimal policy in the long run.

In many applications, however, the observed proxy reward is biased, that is, $\E[\epsilon_t|S_t,A_t]\neq 0$. In the platform incentive policy design example above, if the firm forgets to measure the indirect cost of over-incentivization, such as riders' safety and other externalities that may affect company's long run profit, then the observed proxy reward becomes higher than the true reward. In addition, if the evaluator of the KPI design gains from its success, she may sugarcoat the data, again causing the observed proxy reward to be higher. When $\E[\epsilon_t|S_t,A_t]\neq 0$, because the agent adjusts her policies based on the biased observed reward, it creates a new bias, which we illustrate below.

\subsection{An Illustrative Example of R-Bias}\label{subsec:R-bias}

Consider an online delivery platform that designs rider incentive schemes each period to maximize its expected cumulative discounted reward.
The problem takes the form of equation~\eqref{eq:Case2}, where $S_t$ represents the platform's MAUs and $A_t$ denotes the incentive strength  such as performance bonus levels, delivery time targets, and associated bonuses or penalties.

Assume that the latent reward function $r(S_t,A_t):=\theta^* S_t A_t - \frac{1}{2}A_t^2$, where $ \theta^* S_t A_t$ is the true long run profit, and $ \frac{1}{2}A_t^2$ is the hidden cost of externalities.
Here, $\theta^*$ is an unknown parameter that measures the effectiveness of incentive.
The product price is normalized to one so that
$r(S_t, A_t)$ is the firm's latent reward in period $t$. The state transition takes the following form:
$S_{t+1} =  c_1 S_t + c_2 A_t +\eta_t$,
where $c_i\geq 0$, $i=1,2$, and $\eta_t$ is random noise independent of $S_t$ and $A_t$ with mean $0$ and variance $\sigma_\eta^2$.
The \emph{observed proxy reward} $R_t=r(S_t,A_t)+\epsilon_t$ and for deriving closed form solutions, assume that $\E[\epsilon_t|S_t,A_t] = \beta A_t^2$ for some $\beta>0$.
Namely, the observed proxy reward is biased. (If $\epsilon_t$ is unbiased, this problem is a linear-quadratic optimal control problem \citep{BertsekasDP}).

To show why a new form of bias arises from the platform adjusting its policies based on the observed proxy reward, we first examine the simple case with $c_2=0$. That is, the platform's action $A_t$ does not affect the future states.
If the action does affect future states ($c_2>0$),
the magnitude of this new bias further increases, which we discuss later in this section.

If $c_2=0$, let us first assume that
the platform uses a \emph{fixed} linear policy $A_t = \theta_{\mathsf{pol}} S_t$, where $\theta_{\mathsf{pol}}$ is
exogenously given and
not necessarily the same as the true parameter value $\theta^*$.
To estimate $\theta^*$, a common strategy is to regress $R_t + \frac{1}{2}A_t^2 = \theta^* S_t A_t + \epsilon_t$ on $S_t A_t$, provided that the cost $\frac{1}{2}A_t^2$ is known.
The standard regression formula shows that,
in the long run, the estimate of $\theta^*$ is given by
\begin{equation}\label{eq:static_est}
\plim \hat\theta_{\mathsf{est}} ( \theta_{\mathsf{pol}})
 =
\theta^*+\frac{\Cov(S_t A_t, \epsilon_t)}{\Var(S_t A_t)} =  \theta^* + \frac{\Cov(\theta_{\mathsf{pol}} S_t^2, \beta \theta_{\mathsf{pol}}^2 S_t^2)}{\Var(\theta_{\mathsf{pol}} S_t^2)} = \theta^*+\beta \theta_{\mathsf{pol}}.
\end{equation}

Therefore, the long-run bias of the estimate is $\beta\theta_{\mathsf{pol}}$.
This bias reveals the endogeneity in the setting because the noise $\epsilon_t$ is correlated with the action taken.
Notice that even if the platform chooses the optimal policy $A_t = \theta^* S_t$, such bias remains, and it equals $\theta^*+\beta\theta^*$.

Next, let the platform adjust its policy based on existing data. In particular, suppose the firm starts with a policy $A_t = \hat{\theta}_0 S_t$, for some $\hat\theta^0$.
If this policy is used for an extended period of time,
then following \eqref{eq:static_est} with $\theta_{\mathsf{pol}} = \hat\theta^0$, the estimate of $\theta^*$ from the data generated under this policy is given by
$\hat{\theta}_{\mathsf{est},0}
= \theta^* + \beta \hat\theta^0$.

Suppose the platform then uses this estimate as a guide and
adjusts her policy to $A_t = \hat{\theta}_1 S_t$, where
$\hat{\theta}_1 = \hat{\theta}_{\mathsf{est},0}$.
After this adjusted policy is used for an extended period of time,
the new estimate of $\theta^*$ is then given by
$\hat{\theta}_{\mathsf{est},1}
= \theta^* + \beta \hat{\theta}_1$.
Continuing this process, the estimate of $\theta^*$ after $k$ iterations is given by
\[
\hat{\theta}_{\mathsf{est},k} =  \theta^* + \beta \hat{\theta}_{k-1}  = \cdots = \theta^* + \beta \theta^* + \cdots + \beta^{k-2}\theta^* + \beta^{k-1}\hat{\theta}_0 \to \frac{\theta^*}{1-\beta},
\]
as $k\to\infty$, provided that $\beta \in (-1,1)$.
We call $\frac{\theta^*}{1-\beta} - \theta^* = \frac{\beta \theta^*}{1 - \beta}$ the \emph{R-bias}. More generally, we use R-bias to refer to the additional bias that arises because the agent adjusts her policy.

The R-bias can also be calculated via a fixed-point argument.
Suppose the long-run policy converges to $A_t = \hat{\theta} S_t$.
Then, consistency between the estimate and the decision requires $\hat\theta_{\mathsf{est}} = \hat\theta_{\mathsf{pol}}$, and consequently we have that:
$\plim \hat\theta_{\mathsf{est}}(\hat{\theta}_{\mathsf{est}}) = \hat{\theta}_{\mathsf{est}}$, 
i.e., $\theta^* + \beta \hat{\theta} = \hat{\theta}$.
This
implies that $\hat{\theta} = \frac{\theta^*}{1-\beta}$,
which, again, gives the same R-bias.
Above, we demonstrate the emergence of R-bias when $c_2=0$.
When $c_2>0$ (so that the high incentive scheme increases service quality and therefore, the number of future users), R-bias also arises, although its value is more involved to calculate.
Table~\ref{table:1} shows numerical values of R-bias for different values of $c_2$ and $\beta$. Notice that R-bias is increasing in $c_2$, suggesting that the Markovian dependence structure may amplify the R-bias.

\begin{table}[t]
\begin{center}
\caption{R-Bias for The Incentive Scheme Design Example ($\theta^*=0.25$, $c_1=0.3$ and $\gamma = 0.95$) \label{table:1}}
\footnotesize{
\begin{tabular}{cccccccc}
\toprule
        & \multicolumn{7}{c}{$c_2$}                               \\
        \cmidrule(lr){2-8}
                           $\beta$     & 0.0    & 0.1    & 0.2    & 0.3    & 0.4    & 0.5    & 0.6    \\
\cmidrule{1-1}
\cmidrule(lr){2-2}
\cmidrule(lr){3-3}
\cmidrule(lr){4-4}
\cmidrule(lr){5-5}
\cmidrule(lr){6-6}
\cmidrule(lr){7-7}
\cmidrule(lr){8-8}
 0.1   & 0.0278 & 0.0278 & 0.0280 & 0.0282 & 0.0285 & 0.0289 & 0.0295 \\
                          0.3   & 0.1071 & 0.1075 & 0.1084 & 0.1100 & 0.1124 & 0.1156 & 0.1198 \\
                          0.5   & 0.2500 & 0.2518 & 0.2572 & 0.2672 & 0.2835 & 0.3104 & 0.3620 \\
                        \bottomrule
\end{tabular}
}
\end{center}
\end{table}

We conclude this section with several remarks.
First, the state variable (the MAUs) can depend on the last period's action, i.e., the incentive scheme, since service quality affects future MAUs while the incentive scheme affects the service quality. As the decision-maker adjusts her policy, the transition of the state variable is also affected.
In this sense, the stochastic environment of the example, and many economic and business problems in general, is iterate-dependent Markovian, i.e.,
the transition distribution of state variables iteratively changes in response to policy adjustments.

Second, the noise in the observed reward can depend on the state-action pair in various ways.
In the platform incentive design example, the noise $\epsilon_t$ correlates only with the action $A_t$, and the dependence structure takes a specific form. In general, the noise can also correlate with the state variable $S_t$.\footnote{To see how the R-bias may arise when the noise is correlated with the state variable, suppose, for example, that $S_{t+1} = c_1 S_t + \eta_t$ and
$
\epsilon_t = \beta S_t^2 + o_t,
$
where, again, $\eta_t$'s and $o_t$'s are i.i.d. random variables with mean zero.
If the firm uses a policy $A_t = \tilde\theta S_t$,
the standard regression formula implies that the long-run estimate of $\theta$ is given by
$
\plim \hat\theta_{\mathsf{static}} (\tilde \theta)
 =
\theta^* +\frac{\Cov(S_t A_t, \epsilon_t)}{\Var(S_t A_t)} =  \theta^* + \frac{\Cov(\tilde\theta S_t^2,\beta  S_t^2)}{\Var(\tilde\theta S_t^2)} = \theta^* +\frac{\beta}{\tilde\theta}.
$
Consequently, to calculate the estimate of $\theta$ in the interactive environment,
the fixed-point argument used earlier implies that the long-run estimate should satisfy
$\theta^* + \frac{\beta}{\hat\theta} =  \hat\theta$.
Thus, the R-bias is given by $\hat\theta - \theta^*=\frac{1}{2}(\sqrt{(\theta^*)^2+4\beta}-\theta^*)$.}
The method and theory developed in this paper allow the noise to be correlated with $S_t$ and $A_t$  simultaneously with a general dependence structure.

Finally, while we demonstrate R-bias in a simple example, R-bias also arises in complex MDPs with state transitions and reward functions unknown to the agent, and the policy of the agent is updated over time. Therefore, when applying RL methods to these complex MDP problems, the resulting ``optimal policies'' are also subject to R-bias. The rest of the paper discusses how to incorporate IVs into RL methods to correct the bias.

\section{IV-Q-learning Algorithm}\label{sec:RL}
In this section,
we first introduce Q-learning, a standard RL algorithm, and then describe how to incorporate IVs into it.
We will also discuss how to incorporate IVs into Actor-Critic, another standard RL algorithm, in the Appendix.

\subsection{Q-learning}\label{sec:q-learning}
Q-learning \citep{WatkinsDayan92} is an iterative algorithm that incrementally estimates the long-term discounted reward and, through doing so, identifies the best action for each state.
A random behavior policy $\pi$ is fixed\footnote{The behavior policy is fixed to facilitate theoretical analysis of Q-learning, a standard treatment in RL literature.
Practical implementations of Q-learning often allow the behavior policy to be ``improved'' based on the current estimate of the optimal Q-function \citep[Ch.~6.5]{SuttonBarto18}.
The improved behavior policy has the advantage of generating data that bears a closer resemblance to the data that the optimal policy would generate,
making the subsequent estimation of the optimal Q-function more efficient.
However, these modifications lack convergence guarantees from a theoretical viewpoint \citep{CalvanoCalzolariDenicoloPastorello20}.
In this paper, we do not examine policy improvement in Q-learning, but rather in Actor-Critic for which the behavior policy is constantly adjusted.}  to generate data from which the algorithm learns optimal policies.
Suppose, for now, that the state and action spaces are both finite.
The optimal Q-function $Q^*$ then takes the form of a look-up table representation; that is, its information can be stored in a matrix, where the entry indexed by $(s,a)$ is $Q^*(s,a)$.

In each period $t$, the agent observes the state $S_t$, selects action $A_t$ according to policy $\pi$, observes an immediate proxy reward $R_t$ and the subsequent state $S_{t+1}$, and finally, updates the estimate of $Q^*$ according to
\begin{equation}\label{eq:tab-Q}
      \widehat{Q}_{t+1}(s, a) =
      \widehat{Q}_t(s, a) + \alpha_t \Bigl[R_t + \gamma \max_{a' \in \calA} \widehat{Q}_t(S_{t+1}, a') - \widehat{Q}_t(s, a) \Bigr] \mathbb{I}_{\{(S_t,A_t) = (s,a)\}},
\end{equation}
for all $s\in\calS$ and $a\in\calA$, where $\mathbb{I}$ denotes the indicator function.
Equation~\eqref{eq:tab-Q} implies that if $ (s,a) \neq  (S_t, A_t)$,
the estimate  of $Q^*(s,a)$ is unchanged: $\widehat{Q}_{t+1}(s,a) = \widehat{Q}_t(s,a)$.
For $(s,a)=(S_t, A_t)$, the updated value $\widehat{Q}_{t+1}(s,a)$ is a weighted average of the previous value $\widehat{Q}_t(s,a)$ and $R_t+\gamma \max_{a' \in \calA} \widehat{Q}_t(S_{t+1}, a')$, which is the sum of the observed immediate reward and the best that the agent thinks she can do from the subsequent state $S_{t+1}$.
The weight $\alpha_t$ is the learning rate that affects the speed of convergence.
With this updating rule, $\widehat{Q}_t$ converges to $Q^*$, provided that the behavior policy permits the agent to visit each state-action pair infinitely often \citep{WatkinsDayan92}.

This ``tabular'' Q-learning algorithm, where Q-functions can be represented by a table indexed by state-action pairs, extends naturally to infinite and continuous state-action spaces.
Assume the value function $Q^*(s,a)$ lies in the linear span of $\phi(s,a)\coloneqq (\phi_1(s,a),\ldots,\phi_p(s,a))^\intercal$,
a vector of $p$ appropriately chosen basis functions, such that
$Q^*(s,a) = \phi(s,a)^\intercal \theta^*$ for some unknown parameter $\theta^*\in \mathbb{R}^p$.
Then, one learns $Q^*(s,a)$ through  $\theta^*$ by iteratively updating its estimate $\theta_t$.
To understand this update, consider the original Q-learning update rule \eqref{eq:tab-Q}, where the estimate $\widehat{Q}_t$ moves toward $Q^*$ in the direction of $\Delta_t I(S_t,A_t)$.
Here,
\[\Delta_t\coloneqq R_t + \gamma \max_{a' \in \calA} \widehat{Q}_t(S_{t+1}, a') - \widehat{Q}_t(S_t, A_t),\]
and $I(S_t, A_t)$ denotes a matrix of indicator functions $\mathbb{I}{{(S_t, A_t) = (s,a)}}$ for all $(s, a)$.

Following this principle, we update $\theta_t$ by stepping in the direction of $\Delta_t \nabla_{\theta}\widehat{Q}_t(S_t, A_t) = \Delta_t \phi(S_t, A_t)$, yielding:
\begin{equation}\label{eq:Q-learning-algorithm}
\theta_{t+1}=\theta_t +\alpha_t \Bigl[R_t + \gamma \max_{a'\in \calA} (\phi(S_{t+1},a')^\intercal\theta_t) - \phi(S_t,A_t)^\intercal\theta_t\Bigr] \phi(S_t,A_t).
\end{equation}
When indicator functions serve as basis functions (mapping each state-action pair to a basis function), equation \eqref{eq:Q-learning-algorithm} reduces to \eqref{eq:tab-Q}.

In this paper, we study inference on $\theta^*$ and therefore, the optimal policy.
Under the linear model,
the optimal policy,  derived from the optimal Q-function, is $\pi^*(s) = \argmax_{a\in\calA} \phi(s,a)^\intercal \theta^*$. This allows us to draw inference on both the estimated optimal policy $\argmax_{a\in\calA} \phi(s,a)^\intercal \theta_t$ and its associated value function $\phi(s,a)^\intercal \theta_t$.

Consider the  platform incentive design example, we can select the basis functions as $\phi(s,a) = (1,s,a, sa, s^2, a^2)^\intercal$ for which $Q^*(s,a) = \phi(s,a)^\intercal \theta^*$ with $\theta^* \in \mathbb{R}^6$. Then, $\phi(S_{t+1},a')^\intercal\theta_t$ is a quadratic function in $a$ and thus can be easily maximized. One can then update the estimate of $\theta^*$ according to ~\eqref{eq:Q-learning-algorithm}. Notice that the basis functions $\phi(s,a)$ are non-linear in $(s,a)$, we can estimate the heterogeneous treatment effect $\Delta Q^*(s):=Q^*(s,1)-Q^*(s,0)$ using the estimate of $Q^*$ if the action $a$ is binary.

\begin{remark}

Throughout this paper, we focus on the case where the optimal Q-function is correctly specified as a linear model with a fixed set of $p$ basis functions. The basis functions are usually chosen based on domain knowledge of the specific problem or extracted using feature engineering techniques.
For finite state-action spaces, using indicator functions as basis functions ensures no model misspecification.

If $Q^*$ is misspecified, then the limit of the iterate $\theta_t$, should it converge, is interpreted as follows.
Let $T^*$ denote the Bellman optimality operator so that the Bellman optimality equation~\eqref{eq:bellman} can be rewritten as
$Q^* = T^*Q^*$.
Let $\Pi_{\phi}$ be the projection operator onto
$\mathrm{span}(\phi)$, the linear space spanned by $\phi$ with respect to the $L_2$ norm.
Define $Q_\theta(s,a) \coloneqq  \phi(s,a)^\intercal \theta$.
Then, $Q_{\theta^*}(s,a) = \phi(s,a)^\intercal \theta^*$
satisfies the ``projected Bellman optimality equation''
$Q_{\theta} = \Pi_{\phi} T^* Q_{\theta}$,
which is equivalent to
\begin{equation}\label{eq:bellman-proj}
\E_{\nu}\Bigl[ \Bigl( r(S_t, A_t) + \gamma \max_{a'\in \calA} (\phi(S_{t+1},a')^\intercal\theta) - \phi(S_t,A_t)^\intercal\theta \Bigr) \phi(S_t, A_t) \Bigr ] =0,
\end{equation}
provided that $R_t = r(S_t, A_t) + \epsilon_t$ is unbiased (i.e.,
$\E[R_t|S_t,A_t] =  r(S_t, A_t)$),
where $\nu$ is the stationary distribution of $(S_t, A_t)$ induced by the behavior policy $\pi$.
If $Q^*$ is correctly specified, then equation \eqref{eq:bellman-proj} is identical to the Bellman optimality equation, and its solution $\theta^*$ recovers the optimal Q-function, i.e., $Q^*(s,a) = \phi(s,a)^\intercal \theta^*$.

To address model misspecification, one may use a sieve approximation
in which the basis dimension grows with the time horizon.
\citet{ChenWhite02} study SA with growing dimensionality.
In the RL setting, \citet{ChenQi22} formulate Q-function estimation for
off-policy evaluation as a nonparametric IV problem and develop sieve
2SLS estimators for beta-mixing trajectories, which include Markov data as a special case.
Their results provide a natural reference for extending our linear
Q-function approximation to a sieve setting.
Our analysis, however, keeps the basis dimension fixed.
A formal sieve extension of our IV-RL framework would require joint
control of the approximation error and the SA error.
We leave this extension for future research.

\end{remark}

\begin{remark}\label{comment:Q-convergence}
The convergence of Q-learning with linear value functions for general state and action spaces requires significantly stronger conditions than that of original Q-learning \citep{MeloMeynRibeiro08}. Loosely speaking, the behavior policy $\pi$ should be ``close'' to $\pi^*$, i.e., the optimal policy that we are supposed to identify in the first place, unless the discount factor $\gamma$ is sufficiently small. Otherwise, counterexamples can be constructed in which the estimate $\theta_t$ in \eqref{eq:Q-learning-algorithm} diverges \citep[Ch.~11.2]{SuttonBarto18}.
\end{remark}

\subsection{Incorporating IVs into Q-learning} \label{subsec:IV}
We now describe and motivate an IV-based approach to address the endogeneity issue in RL algorithms when the observed rewards are biased.
Under the updating rule \eqref{eq:Q-learning-algorithm}, the policies converge to a limit that satisfies
\begin{equation}\label{eq:Q-learning-biased}
\E_{\nu}\Bigl[ \Bigl( \underbrace{R_t  + \gamma \max_{a'\in \calA} (\phi(S_{t+1},a')^\intercal\theta)}_{Y_t} - {\underbrace{\phi(S_t,A_t)}_{X_t}}^\intercal \theta  \Bigr) \phi(S_t, A_t) \Bigr ] =0.
\end{equation}

When the observed reward $R_t = r(S_t, A_t) + \epsilon_t$ is unbiased, i.e., $\E[\epsilon_t|S_t, A_t]=0$, the limiting policy satisfies the projected Bellman optimality equation~\eqref{eq:bellman-proj}. When the observed reward is biased, i.e., $\E[\epsilon_t|S_t, A_t]\neq 0$, an extra term $\E_{\nu}[\epsilon_t \phi(S_t, A_t)]$ appears, causing the limiting policy from Q-learning to differ from the optimal policy. Similar problems also occur to the Actor-Critic algorithm as shown in Appendix~\ref{app:IV-AC}.

To address the problem, note that equation~\eqref{eq:Q-learning-biased} is in the form of $\E[(Y_t - X_t^\intercal \theta) X_t] =0$. This is similar to the moment condition for estimating a linear regression model via OLS, except here, the ``dependent variable'' $Y_t$ and the ``explanatory variables'' $X_t$ are generated from an MDP. When there is endogeneity problem in OLS, a standard approach is to introduce instrument variables and correct the bias via 2SLS. This motivates the following bias-correcting idea that is akin to 2SLS.

Specifically, let $Z_t$ be a vector of IVs so that $Z_t$ is correlated with $\phi(S_t, A_t)$ but uncorrelated with $\epsilon_t$. Following the idea of 2SLS,
we first use OLS to regress $\phi(S_t, A_t)$ on $Z_t$ to
estimate the projection matrix $\theta_{\SFI}^*$, which satisfies the equation
\begin{equation}\label{eq:IV-moment-condition}
\E_{\nu}[(\phi(S_t,A_t) - \theta_{\SFI} Z_t) Z_t^\intercal ] = 0,
\end{equation}
where $\nu$, with a slight abuse of notation, is the stationary distribution of $(S_t,A_t,Z_t)$.
We then use the projected value $\theta_{\SFI}^* Z_t$ in lieu of $\phi(S_t,A_t)$ in equation~\eqref{eq:Q-learning-biased} so that the aforementioned extra term $\E_{\nu}[\epsilon_t \phi(S_t, A_t)] = \E_{\nu}[\epsilon_t \theta_{\SFI}^* Z_t] = 0$.
This reduces equation~\eqref{eq:Q-learning-biased} to the projected Bellman optimality equation~\eqref{eq:bellman-proj}, thereby solving the endogeneity problem.

In the context of Q-learning, this idea leads to the following procedure. First, we apply SA (see Section~\ref{sec:analysis} for its definition) to solve equation~\eqref{eq:IV-moment-condition}, which yields the updating rule for $\theta_{\SFI,t}$ as an estimate of $\theta_{\SFI}^*$:
\begin{equation}\label{eq:IV-iterate}
    \theta_{\SFI,t+1} =  \theta_{\SFI,t} + \alpha_{\SFI,t} \bigl(\phi(S_t,A_t) - \theta_{\SFI,t} Z_t \bigr) Z_t^\intercal,
\end{equation}
where $A_t$ is generated by the behavior policy $\pi$ of Q-learning.
Having an estimate of $\theta_{\SFI}^*$ in period $t$,
we modify the updating rule \eqref{eq:Q-learning-algorithm} so that the updating is done in the direction of $\theta_{\SFI,t} Z_t$ rather than $\phi(S_t, A_t)$.
Specifically,
\begin{equation}\label{eq:Q-learning-debiased}
    \theta_{\SFQ, t+1}=\theta_{\SFQ, t} +\alpha_{\SFQ,t} \Bigl[R_t + \gamma \max_{a'\in \calA} (\phi(S_{t+1},a')^\intercal\theta_{\SFQ,t}) - \phi(S_t,A_t)^\intercal\theta_{\SFQ,t}\Bigr] \theta_{\SFI,t} Z_t,
\end{equation}
where the subscript $\SFQ$ indicates quantities associated with  Q-function approximation.

The detailed updating rules for IV-Q-learning are presented in Algorithm \ref{algo:IV-Q}. The algorithm incorporates two features that are standard in machine learning but may be unfamiliar to economists.

First, the algorithm permits the use of \emph{minibatches}, that is, the parameters are updated once every $B$ time periods (where $B$ can be finite or increase with time horizon $T$). Within each minibatch, a total number of $B$ state transitions occur, and the resulting state-action-reward tuples are recorded to provide a more accurate updating direction.
The use of minibatches, therefore, helps stabilize the convergence of algorithms and is commonly used in RL in practice \citep[Ch.~16]{SuttonBarto18}.

Second, the algorithms contain a projection step, defined as $\Pi_{\Theta}(\theta)\coloneqq \argmin_{\tilde{\theta}\in \Theta  } \|\theta-\tilde{\theta}\|$, between parameter updates.
Widely used in practice \citep{KushnerYin03}, it prevents occasional large random fluctuations in the wrong direction in early iterations, in which case it may take the iterates a long time to return to a convergent trajectory.

\begin{algorithm}[t]
\caption{IV-Q-learning}\label{algo:IV-Q}
\SetAlgoLined
\DontPrintSemicolon
\SetKwInOut{Input}{Input}
\SetKwInOut{Output}{Output}
\small{
\Input{Features $\phi$, behavior policy $\pi$, minibatch size $B$,
and learning rates $(\alpha_{\SFQ,k},\alpha_{\SFI,k})$ }
Initialize $\theta_{\SFQ,0}$, $\theta_{\SFI,0}$, and $S_0$ at random, and take action $A_0 \sim \pi(\cdot|S_0)$.\;
\For{all $k=0,1,\ldots$}{
\For{all $t=k{B},\ldots,(k+1)B-1$}{
Observe $(Z_t, R_t, S_{t+1})$
and take action $A_{t+1} \sim \pi(\cdot|S_{t+1})$.}
Update $\theta_{\SFQ,k}$ and $\theta_{\SFI,k}$ via
\vspace{-1ex}
\begin{align*}
    \theta_{\SFQ,k+1}={}& \theta_{\SFQ,k} + \frac{\alpha_{\SFQ,k}}{B} \sum_{t=kB}^{(k+1)B-1} \Bigl(R_t + \gamma \max_{a'\in \calA} (\phi(S_{t+1},a')^\intercal\theta_{\SFQ,k}) - \phi(S_t,A_t)^\intercal\theta_{\SFQ,k}\Bigr) \theta_{\SFI,k} Z_t, \\
    \theta_{\SFI,k+1} ={}& \theta_{\SFI,k} + \frac{\alpha_{\SFI,k}}{B} \sum_{t=kB}^{(k+1)B-1} \bigl( \phi(S_t,A_t)-\theta_{\SFI,k} Z_t \bigr) Z_t^\intercal.
\end{align*}\;
\vspace{-3ex}
Perform projection: $\theta_{\SFX,k+1} \leftarrow \Pi_{\Theta}(\theta_{\SFX,k+1})$ for $\SFX\in \{\SFQ,\SFI\}$.
}
}
\end{algorithm}

Let $\{\theta_{\SFQ}^*, \theta_{\SFI}^*\}$
be the limit of the IV-Q-learning algorithm, when it converge.
We will show in Section \ref{sec:analysis} that
for Algorithm~\ref{algo:IV-Q},
$\{\theta_{\SFQ}^*, \theta_{\SFI}^*\}$ satisfies
\begin{equation}\label{eq:iv-1}
\left\{
\begin{aligned}
& \E_{\nu}\Bigl[ \Bigl(R_t  + \gamma \max_{a'\in \calA} (\phi(S_{t+1},a')^\intercal\theta_{\SFQ}) - \phi(S_t,A_t)^\intercal \theta_{\SFQ} \Bigr) \theta_{\SFI} Z_t \Bigr ] =0,  \\
& \E_{\nu}[(\phi(S_t,A_t) - \theta_{\SFI} Z_t) Z_t^\intercal ] = 0,
\end{aligned}
\right.
\end{equation}
where $\nu$ is the stationary distribution of $(S_t,A_t,Z_t)$ induced by the behavior policy $\pi$.

\subsection{IVs in MDP Settings}

We take stock repurchase as an example, and later an empirical study is presented in this context. Let $S_t$ represent the firm’s state variables in period $t$, such as financial conditions, earnings levels (e.g., EPS or earnings per share), recent stock performance, and balance sheet characteristics. The action $A_t$ corresponds to the firm’s repurchase decision, including the timing and intensity of share buybacks. Importantly, in the corporate finance literature, there is no evidence that earnings levels or other state variables mechanically trigger repurchase decisions. Instead, repurchases are typically driven by \emph{marginal shocks}—such as negative pre-repurchase earnings surprises relative to salient thresholds—rather than by the level of earnings itself.

Endogeneity can arise when repurchase decisions respond to short-term valuation shocks that also affect observed stock returns, leading to biased reward signals.\footnote{Repurchase decisions in the data may reflect managerial private incentives, tax considerations, or other institutional factors that are not necessarily aligned with shareholder-value maximization. In the empirical application, these considerations are treated as determinants of the observed repurchase action choice rather than as components of the firm’s value-creation objective, and earnings-threshold-related incentives are used as an empirically tractable source of variation to study the valuation consequences of repurchase actions.} In this context, valid IVs are factors that generate exogenous variation in such marginal incentives without directly affecting the firm's latent per-period value creation (i.e., latent reward). For example, earnings threshold crossings or near-miss events can induce discrete changes in managerial incentives to repurchase shares, even when underlying firm fundamentals remain smooth.

While such IVs may appear similar to ``exploration'' in reinforcement learning algorithms, their roles are fundamentally distinct. Exploration refers to deliberately choosing suboptimal actions to learn about the environment, whereas IVs perturb the data-generating process itself in a structured and exogenous manner. In the stock repurchase context, standard exploration would involve experimenting with different repurchase intensities, while IV-based causal exploration operates through exogenous variation in marginal repurchase incentives that is not driven by managerial beliefs or market sentiment.

When observed rewards are biased due to short-term market noise or valuation sentiment, exploration alone is insufficient to recover the optimal policy. In such settings, identifying optimal repurchase strategies requires exogenous variation that shifts repurchase decisions independently of the reward measurement error, thereby enabling consistent estimation of the underlying value-creation process.

Our framework highlights a novel role for earnings thresholds and other institutional features in corporate finance. Rather than implying a direct causal link from states to actions, these features generate exogenous marginal incentives that facilitate causal identification in dynamic decision problems and help recover optimal repurchase policies.

\section{Theoretical Analysis via Stochastic Approximation}\label{sec:analysis}

In this section, we analyze SA methods in a general environment and then apply the results to the IV-RL algorithms. Because of the technical nature of the analysis, we first provide an overview of the main results in Section~\ref{subsec:mainResults}. The formal statement of the assumptions, results, and the discussion of the proof techniques are contained in later subsections.

\subsection{Overview of Main Results}\label{subsec:mainResults}

We analyze a general formulation encompassing both IV-Q-learning and IV-AC (see Appendix \ref{app:IV-AC}) as special cases.
Both algorithms can be expressed as SA iterations that solve equations of the form
$\E[G(W,\theta)] = 0$,
where $W$ is a random variable with support $\calW$, $\theta\in\mathbb{R}^d$ is an unknown parameter of interest, and $G: \calW\times \mathbb{R}^d\mapsto \mathbb{R}^d$ is a deterministic function.
We consider the following general form of SA that permits projection and minibatch-updating:
\begin{equation}\label{eq:update-scheme-projection}
    \theta_{k+1} = \Pi_{\Theta}\Biggl(\theta_k + \frac{\alpha_k}{B}\sum_{t=kB}^{(k+1)B-1}G(W_t, \theta_k)+ M_{k}\Biggr),
\end{equation}
where
$\alpha_k \propto k^{-\delta}$ for some $\delta\in (\frac{1}{2}, 1]$,\footnote{This choice of learning rate is standard in RL literature. Other values of $\delta$ typically make the algorithm diverge.}
$W_t$ is a random variable approximating $W$ in distribution and $M_k$ is a residual term that arises to account for the estimation error of $\theta_{\SFI}^*$.

For example, the IV-Q-learning algorithm aims to solve equation \eqref{eq:iv-1}.
In addition,
if $\theta_{\SFI}^*$ is known,\footnote{When $\theta_{\SFI}^*$ needs to be estimated, replacing $\theta_{\SFI}^*$ with $\theta_{\SFI,k}$ introduces additional terms that are absorbed by $M_k$; see Appendix~\ref{app:SA} for details.} the update rule for $\theta_{\SFQ,k}$  is reduced to the form of \eqref{eq:update-scheme-projection} with
$\theta_k=\theta_{\SFQ,k}$, $\alpha_k = \alpha_{\SFQ,k}$, $W_t = (S_t, A_t, R_t, Z_t, S_{t+1})$, $M_k = 0$, and
\begin{equation*}%
G(W_t, \theta_{k}) = \Bigl(R_t + \gamma \max_{a'\in \calA} (\phi(S_{t+1},a')^\intercal\theta_{k}) - \phi(S_t,A_t)^\intercal\theta_{k}\Bigr) \theta_{\SFI}^* Z_t.
\end{equation*}

Our first main result (Theorem~\ref{theo:asymp} in Section~\ref{subsec:normality}) establishes the asymptotic normality of $\theta_k$:
with $\rightsquigarrow$ denoting weak convergence,
\begin{equation}\label{eq:SA-CLT}
    (B^{1-\delta} T^\delta)^{\frac{1}{2}}(\theta_{\lfloor T/B \rfloor}-\theta^*)\rightsquigarrow \calN(0,\Gamma^*)\quad \mbox{as } T\to\infty,
\end{equation}
for some positive definite matrix $\Gamma^*$ that is increasing in a measure (described in Section \ref{subsec:normality}) of the intertemporal dependency of
the Markov chain $\{G(W_t,\theta^*):t\geq 0\}$,
provided that $B/T^\frac{\delta}{1+\delta}\rightarrow 0$ as $T\to\infty$. (This condition means $B$ either is fixed or does not grow too fast relative to $T$, because otherwise $\theta_k$ would not be updated enough times.)

The CLT \eqref{eq:SA-CLT} differs from previous convergence rate analyses of SA \citep{KushnerYin03,KondaTsitsiklis04,mokkadem2006} in two aspects.
First, our result accommodates the use of  minibatches and characterizes its effect on the convergence rate.
Second, our analysis  extends the existing studies by allowing the environment to be \emph{iterate-dependent Markovian}. This extension accommodates applications where the state variables are Markovian (instead of i.i.d.) because they are affected by the past. Moreover, by allowing the Markovian structure to be iterate-dependent, our analysis accommodates settings where the behavior policies from RL algorithms improve over time, enabling adaptive data collection. This is a common feature in practice, especially when the state space is large. Without policy improvement and adaptive data collection, the costs associated with identifying the optimal policy can be prohibitive \citep{Bottou_etal13}.

Our second main result involves applying Theorem~\ref{theo:asymp} to IV-Q-learning. As we focus on IV-Q-learning in the maintext, a set of result in the Appendix~\ref{subsec:IV-AC-result} is available for the IV-Actor Critic (AC) algorithm.
Corollary \ref{corollary:CLT-Q} (Section~\ref{sec:apply_SA}) and Corollary \ref{corollary:CLT-AC} (Appendix~\ref{subsec:IV-AC-result}) show that the two algorithms eliminate the R-bias under conditions that correspond to those in Theorems~\ref{theo:asymp}.
These results also allow us to make inferences on
the optimal policy $\pi^*$ and its value function $Q^*$. This contrasts with prior studies on policy evaluation with adaptively collected data that focus on the inference of the value function $Q^\pi$ of a given policy $\pi$ \citep{HadadHirshbergZhanWagerAthey21,ZhanHadadHirshbergAthey21}, especially for the case of the IV-AC algorithm.

\subsection{Key Assumptions}\label{subsec:assumptions}

We now describe and discuss the assumptions used in Theorem \ref{theo:asymp}, starting with those related to the stochastic environment.

\begin{assumption}\label{assump:ergodic}
Let $\Theta \subset \Real^d$ be a compact set.
For each $\theta\in\Theta$, let
$\SFP_\theta$ be a Markov transition distribution on state space $\calW$. For each $t\geq 0$, let $\mathscr{F}_t$ denote the $\sigma$-algebra generated by $\{\theta_0, \theta_1,...,\theta_t, W_0, W_1, \ldots,W_t\}$.  Let $U(w) \coloneqq 1+\|w\|^4 $ and let $f$ be an arbitrary function such that $|f(w)|\leq U(w)$ for all $w\in\calW$.
\begin{enumerate}[label=(\roman*), noitemsep, topsep=0pt]
\item  \label{asp:ergodic_MC}
$\pr(W_{t+1}\in\cdot |\mathscr{F}_t) = \SFP_{\theta_k}(W_{t+1}\in\cdot |W_t)$ for each $k\geq 0$ and $kB\leq t < (k+1)B$.
\item \label{asp:ergodic_geom}
    $\SFP_\theta$ is geometrically ergodic with respect to $U$, uniformly in $\theta\in\Theta$.
    \item \label{asp:ergodic_Lip}
    $\bigl| \int f(w') \bigl( \SFP_{\theta}(\dd{w'}|w) - \SFP_{\widetilde{\theta}}(\dd{w'}|w) \bigr) \bigr|  \leq L_\nu \|\theta-\widetilde{\theta}\| U(w)$
    for some $L_\nu>0$ and all  $\theta,\widetilde{\theta}\in \Theta$, $w\in\calW$.
    \item \label{asp:boundedness}
     $\bigl|\int f(w_t)\Pi_{l=0}^{t-1} \SFP_{\widetilde{\theta}_l}(\dd{w_{l+1}}|w_{l}) \bigr| \leq K_U U(w_0)$, for all $t>0$ and  $\widetilde{\theta}_0,...,\widetilde{\theta}_{t-1}\in \Theta$.
\end{enumerate}
\end{assumption}

Before discussing Assumption~\ref{assump:ergodic} in detail, we note that if $\SFP_\theta$ is independent of $\theta$ (i.e., $\SFP_\theta \equiv \SFP$ for some $\SFP$), which responds to the situation where the behavior policy of an RL algorithm is fixed such as Q-learning. Then, 
Assumption~\ref{assump:ergodic} is simply reduced to that $W_t$ is a geometrically ergodic (with respect to $U$) Markov chain. (In this case, condition~\ref{asp:ergodic_Lip} becomes redundant and is no longer needed.)
Loosely speaking, this assumption states that the Markov chain $W_t$ admits a unique stationary distribution $\nu$, and
the $t$-step transition distribution induced by $\SFP$
converges to $\nu$ at an exponential rate of $t$.
This is a standard condition for analyzing general state-space Markov chains \citep[Ch.~15]{MeynTweedie09}.

To accommodate adaptive behavior policies,
Assumption~\ref{assump:ergodic} extends the condition above as follows. Assumption~\ref{assump:ergodic}\ref{asp:ergodic_MC} defines the iterate-dependent Markovian structure: the Markov chain transition depends on the policy.
Assumption~\ref{assump:ergodic}\ref{asp:ergodic_geom} is a technical condition requiring  uniform geometric ergodicity of $\mathsf{P}_\theta$ across $\theta$, which is also used in
\cite{KondaTsitsiklis03} to prove convergence of standard AC algorithms.
A key implication of this condition is that $\mathsf{P}_\theta$ possesses a unique stationary distribution $\nu_\theta$ for each $\theta$.
Assumption~\ref{assump:ergodic}\ref{asp:ergodic_Lip} is a Lipschitz condition that guarantees the behavior policies are updated gradually by the step of policy improvement.
Assumption~\ref{assump:ergodic}\ref{asp:boundedness} regularizes the tails of multi-step transition distributions to ensure that, loosely speaking, when the data-generating process is changing,
the data $W_t$ may still converge to a stationary distribution.

Next, we state the conditions on $G(W,\theta)$ in \eqref{eq:update-scheme-projection} and $\bar{G}(\theta)\coloneqq \E_{\nu_\theta}[G(W,\theta)] $. These conditions are the standard regularity conditions in SA but are stated at a general level. In Section~\ref{sec:apply_SA}, we discuss how to translate these conditions to specific conditions on the primitives (reward function and so on) for specific RL algorithms.

\begin{assumption}\label{assump:SA_Lip} %
    $\sup_{\theta\in\Theta} \|G(w,\theta)\|\leq L(1+\|w\|)$ and  $\| G(w, \theta) - G(w, \widetilde{\theta}) \| \leq L(1+\|w\|)  \|\theta-\widetilde{\theta}\|$
    for some $L>0$, all $w\in\calW$, and  all $\theta,\widetilde{\theta} \in\Theta$.
\end{assumption}

\begin{assumption}\label{assump:G-bar}
\begin{enumerate}[label=(\roman*), noitemsep, topsep=0pt]
    \item \label{asp:SA_unique} $\bar{G}(\theta)=0$ has a unique solution $\theta^*$ in the interior of $\Theta$.

    \item \label{asp:SA_nsd}
    $(\theta-\theta^*)^\intercal \bar{G}(\theta) \leq - \psi \norm{\theta-\theta^*}^2$
    for some $\psi>0$ and all  $\theta\in\Theta$.

    \item \label{asp:SA_taylor} $\bar{G}(\theta)$ is twice continuously differentiable
    and
$\|\bar{G}(\theta) - \nabla \bar{G}(\theta^*) (\theta - \theta^*)  \| \leq C_r \|\theta - \theta^*\|^2  $ for some $C_r>0$ and all $\theta\in\Theta$.
\end{enumerate}
\end{assumption}

Assumption~\ref{assump:SA_Lip} lists the common linear growth and the Lipschitz conditions.
Assumption~\ref{assump:G-bar}\ref{asp:SA_unique} is a standard, albeit strong, condition for theoretical analysis of SA \citep{KushnerYin03}.
Although we only require $\bar{G}(\theta)$ to have a unique root in the subset $\Theta$ rather than in all of $\mathbb{R}^d$,
identifying such a region a priori may still pose practical challenges for practitioners.
Extending our theory to cases without solution uniqueness in $\Theta$ lies outside this paper's scope and is left for future research.
Assumption~\ref{assump:G-bar}\ref{asp:SA_nsd} is a classical condition \citep[Ch.~6]{NevelsonHasminskii76} that guarantees $\bar{G}$ is a contraction mapping. Finally, Assumption~\ref{assump:G-bar}\ref{asp:SA_taylor} enables Taylor's expansion of $\bar{G}$ around $\theta^*$,
a standard condition for establishing CLTs for nonlinear SA \citep{mokkadem2006}.

Finally, we describe the condition on the residual term $M_k$ in equation~\eqref{eq:update-scheme-projection}.

\begin{assumption}\label{assump:SA_error} %
    $M_k$ is $\mathscr{F}_{(k+1)B-1}$-measurable; $\E_w [\| M_{k}\| ] \leq C_M \alpha_k^2 $,
    $\E_w [\|M_k\|^2]\leq C_M (\frac{\alpha_k^3}{B}+\alpha_k^4)$, and $\E[\|M_k\||\mathscr{F}_{kB}]\leq C_M\alpha_k (1+\|W_{kB}\|)$  for some fixed constant $C_M>0$, all  $k\geq 0$, and all $w\in\calW$,
    where $\E_w[\cdot] \coloneqq \E[\cdot | W_0=w]$.
\end{assumption}

Assumption~\ref{assump:SA_error} ensures that the residual term $M_k$ from the proposed IV-RL algorithms vanishes sufficiently fast so that its effect on the rate of convergence of $\theta_k$ is asymptotically negligible relative to the learning rate $\alpha_k$. Note that $M_k$ is similar in form to the residual term in the classical Kiefer--Wolfowitz algorithm, which accounts for the error caused by using finite differences to estimate gradients. We stress that it is crucial to allow the residual term $M_k$ in the analysis of iterate-dependent Markovian process such as the Actor-Critic algorithm. For details, see Appendix \ref{app:SA}.

\subsection{Asymptotic Normality}\label{subsec:normality}

\begin{theorem}\label{theo:asymp}
Under the Assumptions \ref{assump:ergodic} $\sim$ \ref{assump:SA_error}, let $\{\theta_k: k\geq 1 \}$ be the SA iterates following \eqref{eq:update-scheme-projection}, where $\alpha_k = \alpha_0 k^{-\delta}$ for some $\alpha_0>0$ and $\delta \in (\frac{1}{2},  1]$.
Suppose that $\psi\alpha_0>2$ in the case of $\delta=1$. 
Then, for any $w\in\calW$,  $C>0$, and $\delta_c\in[0, \delta-\frac{1}{2})$,
if $B/T^\frac{\delta}{1+\delta}\rightarrow 0$,
then $(B^{1-\delta} T^\delta)^{\frac{1}{2}}(\theta_{\lfloor T/B \rfloor}-\theta^*)\rightsquigarrow \calN(0,\Gamma^*)$ as $T\to\infty$,
where $\Lambda^*\coloneqq  \nabla \bar{G}(\theta^*)$, $\Sigma^* \coloneqq  \Sigma^*(0)+\sum_{l= 1}^\infty (\Sigma^*(l)+\Sigma^*(l)^\intercal)$, $\Sigma^*(l)\coloneqq \Cov_{\nu_{\theta^*}}[G(W_0,\theta^*), G(W_l,\theta^*)]$, and
\begin{gather}
    \Gamma^* \coloneqq  \left\{
    \begin{array}{ll}
        \alpha_0\int_0^\infty \exp(u \Lambda^*)\Sigma^*  \exp(u \Lambda^*)^\intercal \dd{u}, & \mbox{ if } \delta \in (\frac{1}{2}, 1), \\
        \alpha_0\int_0^\infty \exp(u/\alpha_0) \exp(u \Lambda^*)\Sigma^*  \exp(u \Lambda^*)^\intercal \dd{u},  &  \mbox{ if } \delta = 1.
    \end{array} \right. \label{eq:Sigma}
\end{gather}
\end{theorem}

To prove the asymptotic normality, we follow a common, martingale method:
(i)
Express
$(\theta_{\lfloor T/B \rfloor} - \theta^*)$ as the sum of a martingale and a remainder term,
(ii) invoke a martingale CLT,
and (iii) show the remainder term is asymptotically negligible. However, steps (ii) and (iii) require calculating moments of various orders of $G(W_t, \theta_k)$, and the iterate-dependent Markovian structure dramatically complicates these calculations.

Our key idea lies in how to construct the martingale. In contrast to the standard approach of creating one martingale difference for each data point $W_t$, we partition the data $W_t$'s into groups and construct a sequence of martingale differences for each group. Note that when the minibatch size $B$ is large enough relative to the order of $\ln T$, one can simply use a minibatch as a group. When the group size $N$ is large, the data of one group are mostly separated from another group's data by a long time interval, implying that the dependence between the groups is weak.
This weak dependence allows us to apply the results developed for finite-time performance analysis to verify the conditions for a martingale CLT.
However, the remainder term, i.e., the difference between the martingale and the quantity of interest $(\theta_{\lfloor T/B \rfloor} - \theta^*)$, grows with $N$. By choosing a proper value of $N$, one can strike a balance between the two competing forces and, therefore, obtain results in Theorem \ref{theo:asymp}.
See Section~\ref{app:proof-SA} of the online supplemental material for the details.

Theorem \ref{theo:asymp} highlights the impact of the Markovian dependence structure:
$\Sigma^*$ will be large,
if $G(W_t,\theta^*)$ exhibits a strong intertemporal dependence structure, meaning $\Sigma^*(l)$'s are large.
Therefore,
Theorem~\ref{theo:asymp} sheds light on why
``long-term strategies'' are very challenging to learn.
It also suggests that to learn $\theta^*$ more efficiently,
one should perform
more exploration to overcome intertemporal dependencies.
 
\begin{remark}
In stochastic gradient descent literature, the batch size often increases with the iterate number $t$ \citep{ChenLeeTongZhang20}.
Our focus differs in two key aspects: (1) we consider settings where the time horizon $T$ is known beforehand and the agent seeks inference on the estimates after algorithm termination, and (2) we use a fixed batch size, which is standard in RL literature. Our theory can be extended to cases with increasing batch sizes, but this lies beyond our paper's scope.
\end{remark}

\begin{remark}
    The sieve interpretation of the linear Q-function approximation raises
a natural question about inference on $\theta^*$. The results of \citet{AiChen07}
suggest that, when a misspecified function enters the moment
restrictions additively, the misspecification may not affect the
first-order asymptotic variance of a finite-dimensional parameter.
A similar robustness property may hold for our IV-RL algorithms.

The intuition is clearest for IV-Actor-Critic (a popular RL algorithm introduced in Appendix). Conditional on the actor, the
critic estimates the Q-function through a Bellman equation that is
linear in the critic parameter. The estimated Q-function also enters
the actor's policy-gradient equation linearly. The critic's sieve
approximation error therefore enters the estimating equation for the
finite-dimensional actor parameter $\theta_{\SFA}^*$ as an additive
nuisance term. Under suitable rate and orthogonality conditions, this
error may not affect the first-order asymptotic variance of
$\theta_{\SFA}^*$.

For IV-Q-learning, the same argument is more subtle. The Q-function
enters the Bellman optimality equation through the maximization
operator in \eqref{eq:iv-1}. An approximation error can therefore
change both the maximized value and the action that attains the
maximum. It does not enter the estimating equation in a purely
additive way. 
Our current theory keeps the basis dimension fixed. 
Formal robustness
results for both IV-AC and IV-Q-learning in the Sieve context require substantial further work.
\end{remark}

\subsection{Applying SA Theory to IV-RL Algorithms} \label{sec:apply_SA}

We now apply our general results for SA to IV-Q-learning.
We show that $\theta_{\SFQ,k}$ in IV-Q-learning can achieve the parametric rate of convergence when the learning rates are proportional to $k^{-1}$.
A similar result is derived for IV-AC in Appendix~\ref{app:IV-AC}.
We first describe the conditions required by both algorithms, starting with the assumptions on IVs.

\begin{assumption}\label{assump:IV}
For each $t\geq 0$, there exists a random vector $Z_t\in \Real^q$ such that $q\geq p$ and the following conditions are satisfied.
    \begin{enumerate}[label=(\roman*), noitemsep, topsep=0pt]
        \item \label{asp:IV-id}(Identification) $\E[\epsilon_t Z_t|S_t,A_t] = 0$; $\phi(S_t,A_t) = \theta_{\SFI}^* Z_t + \eta_t$, where $\theta_{\SFI}^*\in\Real^{p\times q}$ is a deterministic full-rank matrix, and $\eta_t\in\Real^p$ is a random vector with $\E[\eta_t Z_t|\mathcal{F}_{t-1}] = 0$, where $\mathcal{F}_{t-1}$ is the filtration that contains all information at time $t-1$.
        \item \label{asp:IV-wellpose}(Wellposedness)
        There exists a constant $\psi_{\SFI}>0$ such that
        \begin{itemize}[noitemsep, topsep=0pt]
            \item $\lambda_{\min}(\E_{\nu}[Z_t Z_t^\intercal]) \geq \psi_{\SFI}$ in the case of IV-Q-learning (Algorithm~\ref{algo:IV-Q}), and
            \item  $\lambda_{\min}(\E_{\nu_{\theta_{\SFA}}}[Z_t Z_t^\intercal]) \geq \psi_{\SFI}$ for all  $\theta_{\SFA}\in \Theta_{\SFA}$ in the case of IV-AC (Algorithm~\ref{algo:IV-AC}),
        \end{itemize}
         where $\lambda_{\min}$ denotes the minimum eigenvalue of a square matrix.
        \item \label{asp:IV-tran} (Transition  Irrelevance) $S_{t+1}$ is conditionally independent of $Z_t$ given $(S_t,A_t)$.
    \end{enumerate}
\end{assumption}

Assumption~\ref{assump:IV} describes IV validity conditions in MDP settings.\footnote{In the case of over-identification (i.e., $q > p$), $\theta_{\SFI,k}$ converges to the projection of $\phi(S_t,A_t)$ on $Z_t$.
}
Assumptions~\ref{assump:IV}\ref{asp:IV-id} and \ref{assump:IV}\ref{asp:IV-wellpose} are standard.
Assumption~\ref{assump:IV}\ref{asp:IV-tran} arises from the Markovian environment. This condition requires that the IV does not affect state transitions, which would introduce additional bias. The condition is satisfied when $Z_t$ influences the future only through its effect on $S_t$. For example, consider a state transition written as $S_{t+1} = f(S_t,A_t,\xi_t)$ for some deterministic function $f$ and random variable $\xi_t$. A common form is the linear transition model $S_{t+1}:= \beta_0 +\beta_1 S_t + \beta_2 A_t + \xi_t$, where $\xi_t$ is an i.i.d. random noise. If $\xi_t \perp Z_t$, the IV affects $S_{t+1}$ only through $(S_t, A_t)$, ensuring $S_{t+1} \perp Z_t|(S_t,A_t)$.

One sufficient condition for IVs to exist and satisfy the transition irrelevance condition is that $\xi_t$'s are independent shocks (e.g., productivity shocks).
In this case, lagged terms $S_{t-l}$ or $A_{t-l}$, $l=1,2,\ldots,$ would satisfy the transition irrelevance condition while maintaining correlation with $S_t$, $A_t$ through the Markovian property.
Nevertheless, a trade-off exists between IV strength and potential bias when using lag terms.
As variables become more lagged, their correlation with current variables $S_t$, $A_t$ weakens, reducing IV strength.
Adding more lagged terms can improve total IV strength and estimation efficiency for the optimal Q-function and optimal policy. However, this may lead to weak IV or many IV problems, resulting in biased estimates \citep{Hansen01102008}.

The following assumptions pertain specifically to IV-Q-learning.
Corresponding assumptions for IV-AC are presented in Appendix \ref{subsec:IV-AC-result}.

\begin{assumption}\label{assump:phi}
    \begin{enumerate}[label=(\roman*), noitemsep, topsep=0pt]
        \item \label{asp:phi-linear-indep} $\phi(S_t,A_t)$ is a vector of linearly independent random variables
        under the distribution $\nu$ in the case of IV-Q-learning (Algorithm~\ref{algo:IV-Q}).
        \item \label{asp:linear_phi} $\|\phi(s,a)\| \leq C(\norm{s}^{\ell_\phi}+\norm{a}^{\ell_\phi}+1)$ for some $C,\ell_\phi>0$, and  all $s\in\calS$, $a\in\calA$.
    \end{enumerate}
\end{assumption}

Assumption~\ref{assump:phi}\ref{asp:phi-linear-indep} is a natural generalization of the standard linear independence condition on the explanatory variables of linear regression.
It ensures that the projection from a Q-function to $\mathrm{span}(\phi)$ is unique, so $\theta_{\SFQ}^*$ is the unique solution to  equation~\eqref{eq:iv-1}  for IV-Q-learning.
Assumption~\ref{assump:phi}\ref{asp:linear_phi} is used to verify the growth condition in Assumption~\ref{assump:SA_Lip}.
Note that
\cite{KondaTsitsiklis03} use
a stronger growth condition on $\phi$ to prove the convergence of standard Actor-Critic algorithms.

\begin{assumption}\label{assump:IV-Q}
Let $\theta \coloneqq (\theta_{\SFQ}^\intercal, \mathrm{vec}{(\theta_{\SFI}})^\intercal)^\intercal$,  $\Theta \coloneqq \Theta_{\SFQ} \times \Theta_{\SFI} \subset \Real^{p+pq}$.
\begin{enumerate}[label=(\roman*), noitemsep, topsep=0pt]
    \item \label{asp:W_def_Q}
    $W_t^{\mathsf{IVQ}}$ (defined in \eqref{eq:W_IVQ}) and $\Theta$ satisfy Assumption~\ref{assump:ergodic} with $\SFP_\theta\equiv \SFP$ for some $\SFP$.
    \item \label{asp:linear_Q}
    $\max_{a\in\mathcal{A}}\|\phi(s,a)\|\leq C(\norm{s}^{\ell_\phi}+1)$ for some $C,\ell_\phi>0$ and all $s\in\calS$.
    \item \label{asp:contraction_Q} There exists a constant $\psi_\SFQ>0$ such that for all $\theta_\SFQ\in\Theta_{\SFQ}$, with $\widetilde{\theta}_\SFQ \coloneqq \theta_\SFQ-\theta_\SFQ^*$,
    \begin{equation}\label{eq:contraction-Q}
        \gamma^2\E_\nu\Bigl[\max_{a'\in\calA}(\phi(S_{t+1},a')^\intercal\widetilde{\theta}_\SFQ)^2\Bigr] - \E_\nu\bigl[\widetilde{\theta}_\SFQ^\intercal (\theta_{\SFI}^* Z_t)^\intercal (\theta_{\SFI}^* Z_t)\widetilde{\theta}_\SFQ \bigr] \leq -  2\psi_\SFQ \|\widetilde{\theta}_\SFQ\|^2.
    \end{equation}
    \item \label{asp:2diff-Q}  $\E_\nu[\max_{a'\in \calA}(\phi(S_{t+1},a')^\intercal\theta_{\SFQ})]$ is twice continuously differentiable for all $\theta_{\SFQ} \in \Theta_{\SFQ}$.

    \end{enumerate}
\end{assumption}

We impose Assumption~\ref{assump:IV-Q} to verify the conditions of Theorem~\ref{theo:asymp}.
In Assumption~\ref{assump:IV-Q}\ref{asp:W_def_Q}, $W_t^{\mathsf{IVQ}}$ can be simplified to $\{S_t, A_t, R_t, Z_t, S_{t+1}\}$ if these random variables are all bounded.
Assumptions~\ref{assump:IV-Q}\ref{asp:linear_Q}, \ref{assump:IV-Q}\ref{asp:contraction_Q}, and \ref{assump:IV-Q}\ref{asp:2diff-Q} correspond to Assumptions~\ref{assump:SA_Lip}, \ref{assump:G-bar}\ref{asp:SA_nsd}, and  \ref{assump:G-bar}\ref{asp:SA_taylor}, respectively.
These conditions are self-explanatory except Assumption~\ref{assump:IV-Q}\ref{asp:contraction_Q}, particularly the inequality~\eqref{eq:contraction-Q}.
It is critical for verifying the contraction condition (Assumption~\ref{assump:G-bar}\ref{asp:SA_nsd}) for the SA recursion.
This inequality means that the discount factor $\gamma$ cannot be arbitrarily close to one unless the behavior policy $\pi$ is sufficiently close to the optimal policy $\pi^*$ and the IV  $Z_t$ is strong enough.
\cite{MeloMeynRibeiro08} use a similar condition to prove the convergence of Q-learning with linear value functions (see Remark~\ref{comment:Q-convergence}).
This condition potentially narrows the scope of application of IV-Q-learning to practical economic problems,
which is essentially due to the algorithm's lack of policy improvement (see more discussion in our numerical experiments in Section~\ref{sec:numerical}).

\begin{corollary}\label{corollary:CLT-Q}
In Algorithm~\ref{algo:IV-Q},
let $\alpha_{\SFX,k} = \alpha_{\SFX,0} k^{-\delta}$ for some $\alpha_{\SFX,0}>0$, $\delta\in(\frac{1}{2}, 1]$, and all $k\geq 1$, $\SFX\in \{\SFQ,\SFI\}$.
Suppose that (i) $\{\theta_{\SFQ}^*, \theta_{\SFI}^*\}$ is the unique solution to equation~\eqref{eq:iv-1} in the interior of $\Theta$, (ii) Assumptions~\ref{assump:IV}, \ref{assump:phi}, and \ref{assump:IV-Q} hold, and (iii)  $\min\{\psi_{\SFQ} \alpha_{\SFQ,0}, \psi_{\SFI} \alpha_{\SFI,0}\}>2$ when $\delta=1$.
If $B/T^\frac{\delta}{1+\delta}\rightarrow 0$,
then
\[(B^{1-\delta} T^\delta)^{\frac{1}{2}}\bigl(\theta_{\SFQ,\lfloor T/B\rfloor}-\theta_{\SFQ}^*\bigr)\rightsquigarrow \calN(0,\Gamma^*_{\SFQ})\quad \mbox{as } T\to\infty.\]
Furthermore, suppose $\phi(s,a)$ is twice-differentiable, $\max_{a\in \mathcal{A}}(\phi(s,a)^\intercal \theta_Q^*)$ has a solution $\pi^*(s)$, and $\Sigma(s):=\frac{\partial^2 \phi(s,a)^\intercal \theta_Q^*}{\partial a \partial a^\intercal}$ has full rank. 
Let $\pi_{\lfloor T/B\rfloor}(s) =  \argmax_{a\in \mathcal{A}}(\phi(s,a)^\intercal \theta_{Q,\lfloor T/B\rfloor})$. Then, 
\[
(B^{1-\delta}T^{\delta})^{\frac{1}{2}}(\pi_{\lfloor T/B\rfloor}(s) - \pi^*(s)) \rightsquigarrow \mathcal{N}(0,\Sigma(s)^{-1}\Gamma^*_{\SFQ}\Sigma(s)^{-1}) \textrm{ as } T\rightarrow \infty.
\]
\end{corollary}

Corollary~\ref{corollary:CLT-Q} makes clear that IV-Q-learning can correct for R-bias and learn optimal Q-functions. The asymptotic covariance matrix $\Gamma^*_{\SFQ}$ is defined in Appendix~\ref{app:Cov}. Note that when $\delta=1$, the algorithm converges at a rate of $T^{-\frac{1}{2}}$ (i.e., the parametric rate with respect to the sample size),
although the number of iterations is not necessarily of order $T$.
(For instance, if $B \approx T^{\frac{1}{3}}$, then the number of iterations is approximately $T/B \approx T^{\frac{2}{3}}$.)

\section{Simulation Experiments}\label{sec:numerical}

In this section, we numerically evaluate the two IV-RL algorithms using the MDP problem in the platform incentive design example in Section~\ref{subsec:R-bias}.
The problem is specified as follows.

\begin{enumerate}[label=(\roman*), noitemsep, topsep=0pt]
\item The state transition follows
$S_{t+1} = c_0 + c_1 S_t +  c_2 A_t +\eta_t$, where $ c_0$, $c_1$, and $c_2$ are some constants, and $\eta_t$ is i.i.d. noise.
\item The behavior policies are normal distributions parameterized by $\theta_{\SFA} = (\theta_{\SFA}^{(0)}, \theta_{\SFA}^{(1)})^\intercal \in \Theta_{\SFA} \subset \Real^2$; namely,
    $\pi_{\theta_{\SFA}}(a|s) \propto \exp\Bigl(-\frac{(\theta_{\SFA}^{(0)} + \theta_{\SFA}^{(1)} s -a)^2}{2\sigma_{\SFA}^2}\Bigr)$, where $\sigma_{\SFA}^2$ is a known constant; see \citet[Ch.~13.7]{SuttonBarto18}.
This is equivalent to independently generating
    $A_{t,1} \sim \mathcal{N}(\theta_{\SFA}^{(0)}+\theta_{\SFA}^{(1)}S_t,\sigma_{\SFA}^2/2)$ and $A_{t,2}\sim \mathcal{N}(0,\sigma_{\SFA}^2/2)$, and then setting $A_t = A_{t,1} + A_{t,2}$.

\item The true latent reward function for a state-action pair is
$r(s,a) = r_0  +r_1  a +  r_2  s a + r_3 a^2$, where
$r_0 $, $r_1, r_2 $, and $r_3 $ are some constants.
\item  The observed proxy reward is
$R(S_t,A_t) = r(S_t, A_t) + \epsilon_t$,
where $\epsilon_t= b A_{t,2}^2 + o_t$, $b$ is some constant, and $o_t$ is  i.i.d. noise with mean $0$.
\item The value of $A_{t,1}$ is observable and taken as an IV.

\end{enumerate}

This MDP problem can be solved analytically:
the optimal Q-function is linear in the basis functions $\phi(s,a)=(1, s, a, sa, s^2 , a^2)^\intercal$ and the optimal policy is a linear policy in the form of $\pi^*(s) = \theta_{\SFA}^{*,(0)} + \theta_{\SFA}^{*,(1)} s $.
The goal here is to learn $\theta_{\SFA}^* = (\theta_{\SFA}^{*,(0)}, \theta_{\SFA}^{*,(1)})^\intercal $ without using the knowledge of the state transition or the true reward function.

Because the agent  knows only that $\epsilon_t$ does not depend on $A_{t,1}$, transformations of $A_{t,1}$ and $S_t$ can serve as IVs, provided that $A_{t,2}$ is independent of $S_t$.
Note that $\{A_t, S_tA_t, A_t^2\}$ are potentially endogenous.
Thus, we  construct three IVs $\{A_{t,1}, S_{t}A_{t,1}, A_{t,1}^2\}$ for them and set
$Z_t=(1,S_t, A_{t,1}, S_{t}A_{t,1}, S_t^2, A_{t,1}^2)^\intercal$, where $\{1, S_t,   S_t^2\}$ are the IVs for themselves.

In addition to the two IV-RL algorithms, we consider a ``non-RL'' learning method based on inverse propensity weighting (IPW). 
They are compared based on the inference on the optimal policy, particularly $\pi^*(s)$ for representative values of the state $s$.
\begin{enumerate}[label=(\roman*), noitemsep, topsep=0pt]
    \item IV-Q-learning: The behavior policy is fixed at $\pi_{\theta_{\SFA,0}}$ for some $\theta_{\SFA,0} = (\theta_{\SFA,0}^{(0)}, \theta_{\SFA,0}^{(1)})^\intercal$.
    Use $\widehat{\pi}^*(s) = \argmax_{a} (\phi(s,a)^\intercal \theta_{\SFQ,K})$ to estimate $\pi^*(s)$, where $K=\lfloor T/B \rfloor$.
    \item IV-AC: The parameterized policy class is $\pi_{\theta_{\SFA}}$ and the initial behavior policy is $\pi_{\theta_{\SFA,0}}$.
    Use $\widehat{\pi}^*(s) = \theta_{\SFA,K}^{(0)} +  \theta_{\SFA,K}^{(1)} s$.

    \item 2SLS-IPW:
    First, follow policy $\pi_{\theta_{\SFA,0}}$ to generate a sample path $\{(S_t,A_t, R_t):0\leq t\leq T\}$, and construct an estimate $\widehat{r}(s,a)$ of the true observed reward function using the full sample via 2SLS with the basis functions $\phi(s,a)$. 
    Second, perform off-policy evaluation of $\pi_{\theta_{\SFA}}$, given exploration policy $\pi_{\theta_{\SFA,0}}$:
    for each $\theta_{\SFA}$, estimate $Q^{\pi_{\theta_{\SFA}}}$ via
    $\widehat{Q}^{\pi_{\theta_{\SFA}}}(S_t,A_t) \approx  \rho_t  \widehat{Q}^{\pi_{\theta_{\SFA,0}}}(S_t,A_t)$, where with some appropriate integer $L>0$,
        \begin{gather*}
             \widehat{Q}^{\pi_{\theta_{\SFA,0}}}(S_t,A_t) = \sum_{l=0}^{L-1} \gamma^l \widehat{r}(S_{t+l}, A_{t+l}) \qq{and}
            \rho_t = \frac{\Pi_{l=0}^{L-1}\pi_{\theta_{\SFA}}(A_{t+l}|S_{t+l}) }{\Pi_{l=0}^{L-1}\pi_{\theta_{\SFA,0}}(A_{t+l}|S_{t+l})}.
        \end{gather*}
    Last, compute $\widehat{\theta}_{\SFA}^* = \argmax\limits_{\theta_{\SFA} \in \Theta_{\SFA}}\sum_{t=0}^{T-L}\widehat{Q}^{\pi_{\theta_{\SFA}}}(S_t,A_t)$, and use  $\widehat{\pi}^*(s) = \widehat{\theta}_A^{*,(0)}+\widehat{\theta}_A^{*,(1)}s $.
\end{enumerate}

For the two IV-RL algorithms,
we set the learning rates to be proportional $k^{-1}$ to achieve the parametric rate of convergence.
We consider two cases for the behavior policy $\pi_{\SFA,0}$:
in one case, $\pi_{\SFA,0}$ is close to $\pi^*$, making it ``easy'' for the algorithms to learn the optimal policy,
whereas in the other case, $\pi_{\SFA,0}$ is clearly distinct from $\pi^*$ and we call it a ``hard'' policy.
Specifications of the experimental design are detailed in Appendix~\ref{app:experiment}.
The results are reported in Tables~\ref{table:gamma-0.85} and \ref{table:gamma-0.95} for different discount factors, respectively.

\begin{table}[t]
\begin{center}
\caption{Inference on Optimal Policy $\pi^*(s)$: The Case of
$\gamma=0.85$} \label{table:gamma-0.85}
\footnotesize{
\begin{tabular}{l c c c c c c c c c }
\toprule
&\multicolumn{9}{c}{$\widehat{\pi}^*(s)$ } \\
\cmidrule(lr){2-10}
\multirow{2}{*}{\shortstack[l]{(Easy) $\pi_{\SFA,0}$:\\ $0.6 s + 0.4 \mathcal{N}(0, 1)$ }}  &\multicolumn{3}{c}{$s=0.5$} &\multicolumn{3}{c}{$s=1.0$} &\multicolumn{3}{c}{$s=1.5$} \\
\cmidrule(lr){2-4}
\cmidrule(lr){5-7}
\cmidrule(lr){8-10}
 & Bias & RMSE & CP & Bias & RMSE & CP & Bias & RMSE & CP  \\
\cmidrule(lr){1-1}
\cmidrule(lr){2-4}
\cmidrule(lr){5-7}
\cmidrule(lr){8-10}
IV-Q-learning & -0.002 & 0.018 & 97.6\% &  -0.001 &  0.017 & 99.2\% & -0.003 & 0.036 & 98.8\% \\
IV-AC & \phantom{-}0.002 &  0.021 & 91.2\%  &  -0.003 & 0.026 & 96.8\% &  -0.008 &  0.035  & 97.0\% \\
2SLS-IPW & -0.011 & 0.062 & 76.8\%  & -0.022 &0.080 & 65.2\% & -0.034 &0.117 & 70.2\% \\
\midrule
&\multicolumn{9}{c}{$\widehat{\pi}^*(s) $} \\
\cmidrule(lr){2-10}
\multirow{2}{*}{\shortstack[l]{(Hard) $\pi_{\SFA,0}$:\\ $0.8 s + 0.4 \mathcal{N}(0, 1)$}}  &\multicolumn{3}{c}{$s=0.5$} &\multicolumn{3}{c}{$s=1.0$} &\multicolumn{3}{c}{$s=1.5$} \\
\cmidrule(lr){2-4}
\cmidrule(lr){5-7}
\cmidrule(lr){8-10}
& Bias & RMSE & CP & Bias & RMSE & CP & Bias & RMSE & CP \\
\cmidrule(lr){1-1}
\cmidrule(lr){2-4}
\cmidrule(lr){5-7}
\cmidrule(lr){8-10}
IV-Q-learning & \phantom{-}0.001 & 0.049 & 93.6\%  & \phantom{-}0.004 & 0.037 & 97.0\%  & 0.002 & 0.037 &97.4\% \\
IV-AC & -0.003 &  0.029 &93.2\%  &  \phantom{-}0.001 &0.026 & 97.2\% &  0.003 &0.037 & 95.2\%  \\
2SLS-IPW & -0.006 & 0.087 & 98.4\% & -0.001 &0.092 &73.8\% & 0.007 & 0.139 & 71.0\% \\
\bottomrule
\end{tabular}
}
\end{center}
\footnotesize{
\emph{Note.}  The optimal policy is
$\pi^*(s) = 0.048 + 0.527 s$.
The target for CP (coverage percentage) is 95\%.
The computational time (sec.) with $T=10^4$ is 0.049 for IV-Q-learning, 0.040 for IV-AC, and 4.465 for 2SLS-IPW.
}
\end{table}

\begin{table}[t]
\begin{center}
\caption{Inference on Optimal Policy $\pi^*(s)$: The Case of
$\gamma=0.95$}\label{table:gamma-0.95}
\footnotesize{
\begin{tabular}{l c c c c c c c c c}
\toprule
&\multicolumn{9}{c}{$\widehat{\pi}^*(s) $} \\
\cmidrule(lr){2-10}
\multirow{2}{*}{\shortstack[l]{ (Easy) $\pi_{\SFA,0}$: \\ $0.6 s + 0.4 \mathcal{N}(0, 1)$}}  &\multicolumn{3}{c}{$s=0.5$} &\multicolumn{3}{c}{$s=1.0$} &\multicolumn{3}{c}{$s=1.5$}\\
\cmidrule(lr){2-4}
\cmidrule(lr){5-7}
\cmidrule(lr){8-10}
& Bias & RMSE & CP & Bias & RMSE & CP & Bias & RMSE & CP \\
\cmidrule(lr){1-1}
\cmidrule(lr){2-4}
\cmidrule(lr){5-7}
\cmidrule(lr){8-10}
IV-Q-learning & \phantom{-}0.011 & 0.047 & 93.0\%  & \phantom{-}0.015 & 0.050 & 93.8\% & \phantom{-}0.019 & 0.063 & 93.2\% \\
IV-AC & \phantom{-}0.007 & 0.051 & 95.8\% & \phantom{-}0.006 & 0.068 & 96.6\% & \phantom{-}0.005 & 0.087  & 97.0\% \\
2SLS-IPW & -0.022 & 0.063 & 92.6\% & -0.034 &0.082 & 84.4\% & -0.045 & 0.120 & 88.4\% \\
\midrule
&\multicolumn{9}{c}{$\widehat{\pi}^*(s) $} \\
\cmidrule(lr){2-10}
\multirow{2}{*}{\shortstack[l]{(Hard) $\pi_{\SFA,0}$: \\ $0.8 s + 0.4 \mathcal{N}(0, 1)$}}  &\multicolumn{3}{c}{$s=0.5$} &\multicolumn{3}{c}{$s=1.0$} &\multicolumn{3}{c}{$s=1.5$}\\
\cmidrule(lr){2-4}
\cmidrule(lr){5-7}
\cmidrule(lr){8-10}
 & Bias & RMSE & CP & Bias & RMSE & CP & Bias & RMSE & CP \\
\cmidrule(lr){1-1}
\cmidrule(lr){2-4}
\cmidrule(lr){5-7}
\cmidrule(lr){8-10}
IV-Q-learning & \phantom{-}0.048 & 0.109 & 65.0\% & \phantom{-}0.050 & 0.116 & 72.0\% & \phantom{-}0.042 & 0.108 & 73.6\% \\
IV-AC & -0.001 & 0.057 &96.8\%  & \phantom{-}0.018 & 0.071 & 96.0\% & \phantom{-}0.034 & 0.095  & 94.8\%  \\
2SLS-IPW & -0.023 & 0.092 & 99.8\%  & -0.021 & 0.103 & 89.0\% & -0.019 & 0.150 & 83.0\% \\
\bottomrule
\end{tabular}
}
\end{center}
\footnotesize{\emph{Note.} The optimal policy is
$\pi^*(s) = 0.061 + 0.532 s$. The target for CP (coverage percentage) is 95\%.
The computational time (sec.) with $T=10^4$ is 0.044 for IV-Q-learning, 0.042 for IV-AC, and 4.690 for 2SLS-IPW.
}
\end{table}

The tables show that IV-AC performs well in all the cases.
The initial behavior policy $\pi_{\SFA,0}$ has little impact on its performance.
This is because of policy improvement, as it allows the distribution of the generated data to be increasingly similar to the distribution corresponding to the optimal policy, making learning more efficient.

IV-Q-learning has a performance comparable to IV-AC in general.
The former is even slightly better when the behavior policy $\pi_{\SFA,0}$ is close to $\pi^*$ (the ``easy'' case), which is because IV-Q-learning has fewer parameters to estimate than IV-AC.
When $\pi_{\SFA,0}$ is changed from ``easy'' to ``hard'',
or when $\gamma$  increases from 0.85 to 0.95,
IV-Q-learning's performance deteriorates, exhibiting larger RMSEs and lower CPs.
This suggests that IV-Q-learning tends to work well either when the behavior policy is close to the optimal policy or when the discount factor is relatively small. Note that
when $\pi_{\SFA,0}$ is far from $\pi^*$ (the hard case) and the discount factor is large ($\gamma=0.95$),
IV-Q-learning's performance is unsatisfactory, confirming the drawback
of Q-learning with linear value functions (see Remark~\ref{comment:Q-convergence}).
In contrast, IV-AC continues to perform well because it uses policy improvement.

Last, 2SLS-IPW  has a moderate bias but large RMSE in all the cases, especially when
$\pi_{\SFA,0}$ is far away from $\pi^*$. In addition, 2SLS-IPW  generally yields significant under-coverage.
This is because the weighted data  $\rho_t  \widehat{Q}^{\pi_{\theta_{\SFA,0}}}(S_t,A_t)$ used for estimating $Q^{\pi_{\SFA}}$ are severely heavy-tailed.
This happens because the inverse propensity weight $\rho_t$ tends to explode, if the path $\{(S_{t+l}, A_{t+l}):0\leq l < L\}$ has a low likelihood under  $\pi_{\theta_{\SFA,0}}$ (i.e., the policy that generates the path), but has a high likelihood under the policy $\pi_{\theta_{\SFA}}$ (i.e., the policy to evaluate), which is often the case when the two policies are significantly different. See \cite{HadadHirshbergZhanWagerAthey21} and \cite{ZhanHadadHirshbergAthey21} for similar findings when data are adaptively collected from multi-armed bandit algorithms.

\section{Application: Corporate Share Repurchases}\label{sec:empapplication}

In this section, we apply IV-Q-learning to corporate share repurchase decisions, demonstrating how our methodology corrects for R-bias in a real-world scenario.

\subsection{Background}

To return cash to shareholders, companies can use two payout policies: dividends and share repurchases. Dividends provide direct payments to shareholders, while share repurchases involve the company buying back its own shares from existing shareholders on the secondary market.
Unlike dividends, which distribute cash equally to all shareholders, share repurchases (or buybacks) only provide cash to shareholders who choose to sell. Share repurchases are often viewed favorably by the market, as they signal management's confidence in future earnings and belief that the stock is undervalued. This typically leads to an increase in share price \citep{GrullonMichaely04}, benefiting all shareholders.

Share repurchases also serve as a valuable tool for earnings management, whereby companies strategically employ accounting techniques to enhance their financial reports.
Earnings management often occurs when firms face pressure to adjust their earnings-per-share (EPS) to meet pre-determined targets established in analysts' forecast reports of quarterly financial performance.
Share repurchases offer an immediate mechanism to boost short-term EPS by retiring repurchased shares, thereby mechanically reducing the EPS denominator.
Consider a scenario where a company anticipates reporting an actual EPS of \$2.99, falling slightly short of the target EPS of \$3.00.
In such cases, the firm has a strong incentive to repurchase shares to elevate its post-repurchase EPS to meet the target, thereby avoiding potential stock price declines typically associated with missed market expectations.

This mechanical EPS adjustment through repurchases assumes fixed earnings from the previous quarter. In practice, post-repurchase EPS calculations are complicated by various market factors, such as negative market sentiment. Companies must therefore dynamically adjust their repurchase decisions based on evolving EPS figures to maintain positive surprises and prevent adverse market reactions.

\cite{HribarJenkinsJohnson06} first documented a discontinuity in share repurchase likelihood driven by earnings management considerations. Building on this discovery, \cite{AlmeidaFosKronlund16} developed an IV to examine how EPS-motivated repurchases affect other corporate policies. Their IV for share repurchase volume is an indicator variable that identifies whether a firm would report a negative pre-repurchase EPS surprise in the absence of repurchases, leveraging the observation that firms with marginally negative pre-repurchase EPS surprises demonstrate a higher likelihood to conduct accretive share repurchases.

The repurchases triggered by narrowly missing target EPS connect to findings that firms conduct repurchases when their stock is undervalued \citep{IkenberryLakonishokVermaelen95,BrockmanChung01,PeyerVermaelen08}.
A stock price decline following a marginal EPS miss likely reflects market irrationality or pessimism rather than rational forecasting of the company's future prospects.
When market pessimism drives stock prices lower, even companies meeting EPS targets may experience price drops.
In such scenarios, companies have increased motivation to execute multiple rounds of repurchases following their initial announcement, adjusting both timing and volume until EPS returns to expected levels.
This highlights how repurchases represent endogenous decisions influenced by various corporate state variables and behaviors.

\subsection{Problem Formulation}

We find that the decisions of initiating and iteratively adjusting share repurchases as recurring decisions in response to external conditions, particularly changes in company EPS, meets all the following requirements and conditions, therefore is a good scenario, for applying IV-Q-learning.

\subsubsection{Decision Variable ($A$)}
Share repurchasing constitutes an endogenous decision variable with a recurring nature. Following initial repurchases, companies make subsequent repurchase decisions based on how other variables evolve over time. This share repurchase decision process bears similarity to the incentive cost concept in the platform incentive scheme design problem discussed in the introduction.

\subsubsection{Latent Reward $r$ and Observed Proxy $R$.}

Share repurchasing decisions affect the firm's latent reward $r(S_t,A_t)$, which aggregates contemporaneous monetary payoffs and intangible value creation; firm value is then obtained via intertemporal aggregating $r$. Firm value, equivalently as per-period value creation, serves as the decision maker's objective (i.e., true reward) since it's typically the primary metric boards use to evaluate CEO performance. CEOs' compensation packages, including stock options, and their tenure prospects are heavily tied to both market value and liability status. Our objective is to understand how share repurchase decisions affect the latent reward $r$ as well as firm value, the discounted aggregation of $r$, and to identify the optimal repurchase strategy along three dimensions: the initial decision to repurchase, frequency of repurchases, and repurchase volume.

In the MDP formulation, we interpret the latent reward $r(S_t,A_t)$ as \emph{per-period value creation} rather than accounting profit. Specifically, $r(S_t,A_t)$ aggregates both (i) contemporaneous monetary contributions (e.g., operating cash flows or earnings components) and (ii) contemporaneous \emph{non-monetary} value created in that period that is relevant for the firm's long-run fundamentals but is not directly observable. The latter includes, for example, technological and innovative gains (e.g., R\&D or basic research that may not translate into current income, see \cite{chan2001stock} and \cite{eberhart2004examination}), reputational capital accumulation (e.g., ESG investments, stakeholder engagement, and charitable activities, see \cite{edmans2011does} and \cite{eccles2014impact}), political and regulatory capital (e.g., political connections and influence activities, see \cite{heitz2023corporate}), as well as other forms of intangible asset accumulation that can affect future profitability, financing conditions, and growth opportunities.

Under this interpretation, firm value corresponds to the discounted expected sum of future value creation, which is captured by the value function:
\[
V^\pi(s)=\mathbb{E}\!\left[\sum_{t=0}^{\infty}\gamma^t r(S_t,A_t)\,\middle|\,S_0=s\right].
\]
The decision maker in the data does not observe $r(S_t,A_t)$ directly. Instead, we observe a short-horizon market-based signal $R_t$, measured as the observed reward using the short-term abnormal returns around repurchase-related events, which is also equivalent to the abnormal change in the firm’s market capitalization, defined as the firm’s realized change in market capitalization minus the change implied by the market-wide return over the same period. As emphasized in Section~6.2.2, we use $R_t$ as a proxy for the latent  reward $r(S_t,A_t)$.
This proxy is noisy because market returns incorporate not only changes in fundamentals but also time-varying market sentiment (``animal spirits'') and other sources of noise, so that $R_t = r(S_t,A_t) + \varepsilon_t$ with potential deviations.

\subsubsection{State Variable ($S$)}
EPS serves as the state variable that, alongside factors such as firm investment and cash holdings, influences repurchase decisions.
Conversely, repurchase decisions affect the subsequent period's EPS.
In the immediate term, repurchases mechanically increase EPS by reducing the number of outstanding shares.
In the short to medium term, repurchases send a positive signal to the market. When a company deploys its own funds for share repurchases, it demonstrates management's confidence in future prospects. This signaling effect can enhance investor confidence and market perception, potentially leading to improved operating performance and lower financing costs, which further boost EPS in subsequent periods. This bidirectional relationship makes EPS a recurring state variable.

\subsubsection{Instrumental Variable ($Z$)}
We seek an IV for the decision variable, and we adopt the IV proposed \cite{AlmeidaFosKronlund16}.
Their IV for share repurchases is an indicator of whether a firm would report a slightly negative pre-repurchase EPS surprise if without the buybacks.
The pre-repurchase EPS surprise is calculated as $S_t - T_t$, where $T_t$ represents the analyst's EPS forecast, considered the company's target EPS.

A valid IV must satisfy two key conditions: exogeneity and relevance.
The IV constructed in \cite{AlmeidaFosKronlund16}, which utilizes the ``just-miss'' discontinuity of the negative pre-repurchase EPS surprise, satisfies both conditions:

\begin{enumerate}[label=(\roman*), noitemsep, topsep=0pt]
\item
The negative pre-repurchase EPS surprise is exogenous. As discussed in the regression discontinuity literature, a marginal ``just-miss'' surprise can be viewed as a mere result of bad luck, provided we define the margin as sufficiently narrow.

\item As explained above, the companies with slightly negative pre-repurchase EPS surprises are more likely to engage in accretive share repurchases, therefore this IV is logically related to the decision variable. \footnote{We additionally conduct a diagnostic check
based on a static approximation of the empirical setting. Specifically, we abstract from dynamics and policy learning and estimate a static linear IV regression in which repurchase decisions are treated as a one-period endogenous variable instrumented by earnings-surprise–based instruments. Within this static approximation, the Kleibergen–Paap rk Wald F statistic is 12.19, which indicates that the instruments are not weak in this reduced-form setting.}
\end{enumerate}

\begin{figure}[ht]
    \centering
    \includegraphics[width=0.9\linewidth]{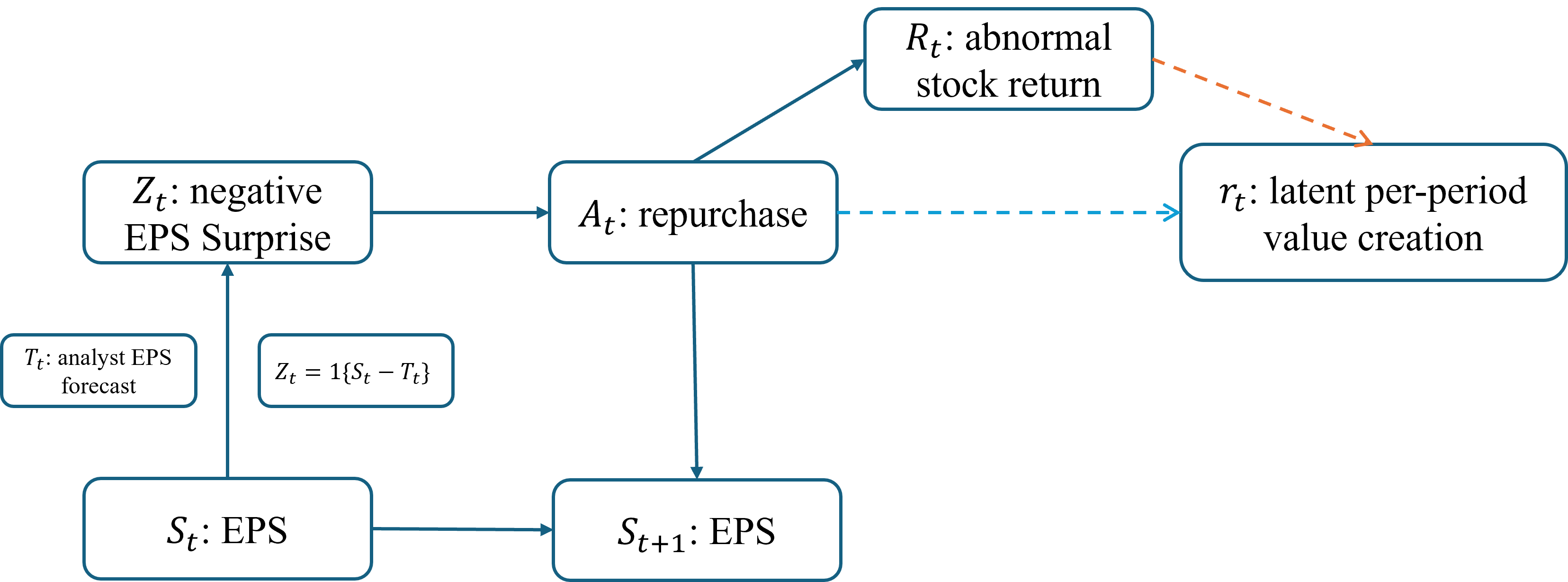}
    \caption{Dynamics of Variables}
    \label{fig:Dynamics of Variables}
\end{figure}

Figure \ref{fig:Dynamics of Variables} illustrates the dynamics of the above variables.\footnote{
We intentionally do not include a direct arrow from $S_t$ to $A_t$ in Figure~1. In the stock repurchase application, firm state variables such as earnings levels (e.g., EPS) are not assumed to mechanically trigger repurchase decisions. Consistent with the corporate finance literature, repurchases are instead driven by \emph{marginal incentive shocks}, such as negative earnings surprises relative to salient thresholds, rather than by the level of earnings or other state variables themselves. These marginal shocks are represented separately (e.g., by $Z_t$) and are the relevant source of exogenous variation affecting $A_t$. The absence of a direct $S_t \rightarrow A_t$ link therefore reflects the intended economic mechanism rather than a modeling omission.
}

\subsection{Data Sources and Sample Construction}

Our analysis integrates data from four primary sources.
We start with the Thomson Reuters Securities Data Company (SDC) transactions dataset, which provides 58 years of U.S. repurchase records.
This dataset covers case-level repurchase information and details the deal date, the board of directors’ initial authorization date, total authorized amount, and the number of shares the company has repurchased for each repurchase program. Next, we merge these observations with firm-quarter fundamental data from the S\&P Compustat dataset, stock-level data from the Center for Research in Security Prices (CRSP) dataset, and analysts forecast data of EPS from Institutional Brokers’ Estimate System (IBES) dataset.

The decision variable $A$ is defined as the repurchase amount normalized by the firm’s market value. Following \cite{AlmeidaFosKronlund16}, we measure the pre-repurchase earnings surprise as the difference between the adjusted pre-repurchase EPS and the median EPS forecast at the end of the quarter. We calculate adjusted pre-repurchase EPS as:
\[
EPS_{adj}=\frac{E+I}{SH+\Delta SH}
\]
where $E$ denotes reported quarterly earnings, $I$ denotes the after-tax interest that would be earned on the fund used to repurchase shares if it were invested in a 3-month T-bill alternatively, $SH$ denotes the number of shares at the end of the quarter, and $\Delta SH$ denotes the estimated number of shares repurchased measured as the repurchase amount divided by average daily stock price during the quarter.

The observable reward ($R$) is constructed as follows.
First, we define the treatment group as firms that conduct repurchases and meet or exceed EPS forecasts, but would have reported negative pre-repurchase EPS surprises if without the buybacks.
Then, for each firm in the treatment group, we identify a corresponding firm in the control group using one-to-one nearest neighbor matching based on total assets, ROA, and sales within the same Fama--French 48 industry classification\footnote{\url{https://mba.tuck.dartmouth.edu/pages/faculty/ken.french/Data\_Library/det\_48\_ind\_port.html}}.
The observable reward is measured as the matched firms' 5-day abnormal returns around earnings announcements (-2 to +2 days relative to announcement).

After excluding the observations with the missing value of the decision variable, our final sample includes 5,666 repurchase cases for 1,195 firms from 1996 to 2020. To mitigate the influence of outliers, the repurchase ratio and the reward variable are winsorized at the 1\% level in both tails.

\subsection{Empirical Results}

We treat the sample observed as an online learning environment with each repurchase case as a data point. More specifically, for each company's each repurchase announcement, we have multiple data points for its multiple times of repurchase actions that aim to achieve its goal of correcting the market value that is lowered by negative sentiment. Using the constructed sample described in the last subsection, we compute the the optimal repurchase amount normalized by the firm’s market value (or the ``repurchase ratio'') for each repurchase action.

Similar to the simulation experiment in Section \ref{sec:numerical}, we assume that the linear-quadratic specification is used:
$$\phi(s,a) = (1,s,a,s^2,sa,a^2)^\intercal.$$
As argued previously, the repurchase action is potentially endogenous, as company CEOs tries to manage EPS to meet the analysts' forecasts. We then adopt the following IV vector, denoted as $\widetilde{Z}$,in our analysis that:
$$\widetilde{Z} = (1,s,z,s^2,sz,s^2z)^\intercal.$$
The parameters of $Q_\theta =  \phi(s,a)^\intercal \theta_Q$ are initialized at the static two-stage least squares (2SLS) estimate of the reward equation, and the first-stage parameter $\Gamma$ is initialized as the full-sample regression coefficient of $\phi(s,a)$ on $\widetilde{Z}$. The SA updates use minibatches of size $B=64$ with learning rate $\alpha_t = \alpha_0 t^{-\delta}$, $\alpha_0 = 1$, $\delta = 1$, together with the preconditioning device described in Appendix D. Discount rate $\gamma$ is set to be $0.95$; the results are insensitive to this choice.\footnote{As a convergence check, we also solve the sample moment conditions in \eqref{eq:iv-1} exactly on the full sample by Newton's method; the resulting optimal policy is nearly identical to the SA output.}

\begin{remark} In the 
application the panel is treated as a collection of firm-level decision processes sharing common
parameters: each firm’s repurchase history under one authorization forms one trajectory, and
trajectories enter the algorithm as separate batches. To a first-order approximation, firms’
processes can be viewed as independent replications of the same underlying decision problem,
in which case the single-series theory applies when the MDP environment is strongly ergodic so that dependency decays fast. In such a case, our inference approach would still apply for analysis.
\end{remark}

\begin{figure}[ht]
    \centering
    \includegraphics[width=0.75\linewidth]{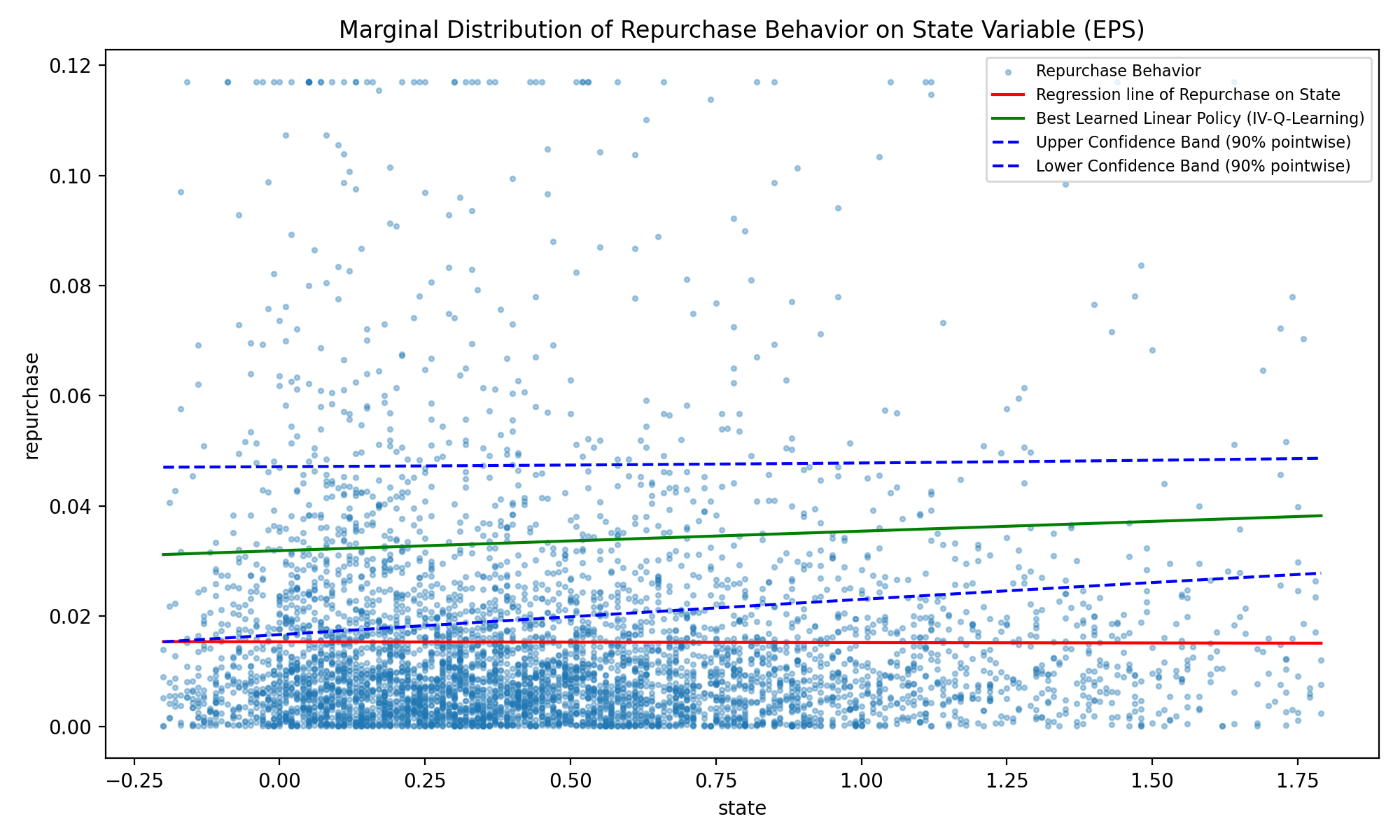}
    \caption{The Scatter Plot and Learned Optimal Policy of Repurchase Actions and States}
    \label{fig:repurchase on state}
\end{figure}

The standard linear regression of repurchase action on state demonstrates a weak and insignificant negative slope that:
\begin{equation*}
    Repurchase_t = 0.0153 - 0.0001 EPS_t +\epsilon_t,
\end{equation*}
with t-stats of $EPS_t$ as $-0.53$, which only reveals a weak negative relationship between repurchase behavior and the EPS. After applying IV-Q-learning, the optimal policy learned (the green line) in Figure \ref{fig:repurchase on state} is essentially flat in the EPS, i.e., its slope is slightly positive and statistically indistinguishable from zero, at a level of about $3$--$4\%$ of the firm's market value, with 90\% pointwise confidence bands (the blue lines) computed according to our asymptotic results in Theorem \ref{theo:asymp}.\footnote{In Figure \ref{fig:repurchase on state}, repurchase ratios are winsorized at the 1\% level for better presentation.} We observe that the actual repurchase behavior lies well below the learned optimal level over the entire range of the state variable.

In the following sections, we first quantify the degree of R-bias, measured as the deviation between observed firm actions and the optimal actions implied by our model. We then investigate the determinants of the magnitude of this bias.

\subsubsection{R-bias in Repurchases}

The optimal repurchase ratios calculated are close to the actual repurchases for some firms and deviate from the actual ratios for others. Here we calculate the difference between optimal and actual repurchase ratios for each repurchase action, and present the summary statistics below.

\begin{table}[htbp]
  \centering
  \caption{Bias Between Actual and Optimal Repurchases}
    \begin{tabular}{rrrr}
    \toprule
          & \multicolumn{1}{c}{Actual Policy} & \multicolumn{1}{c}{Optimal Policy} & \multicolumn{1}{c}{Difference } \\
    \midrule
    \multicolumn{1}{l}{Mean} & 0.015 & 0.034 & -0.018*** \\
          &       &       & (0.000) \\
\cmidrule{4-4}    \multicolumn{1}{l}{Median} & 0.009 & 0.033 & -0.024*** \\
          &       &       & (0.000) \\
    \bottomrule
    \end{tabular}%
  \label{table:repurchase-bias-summary}%
\end{table}%

As shown in Table \ref{table:repurchase-bias-summary}, the mean of the actual repurchase ratio is 0.015, significantly below the mean of the optimal repurchase ratio 0.034; the median of the actual repurchase ratio is 0.009, significantly below the median of the optimal repurchase ratio 0.033. The p-values presented under the differences are very close to zero.

\begin{figure}
    \centering
    \includegraphics[width=0.8\linewidth]{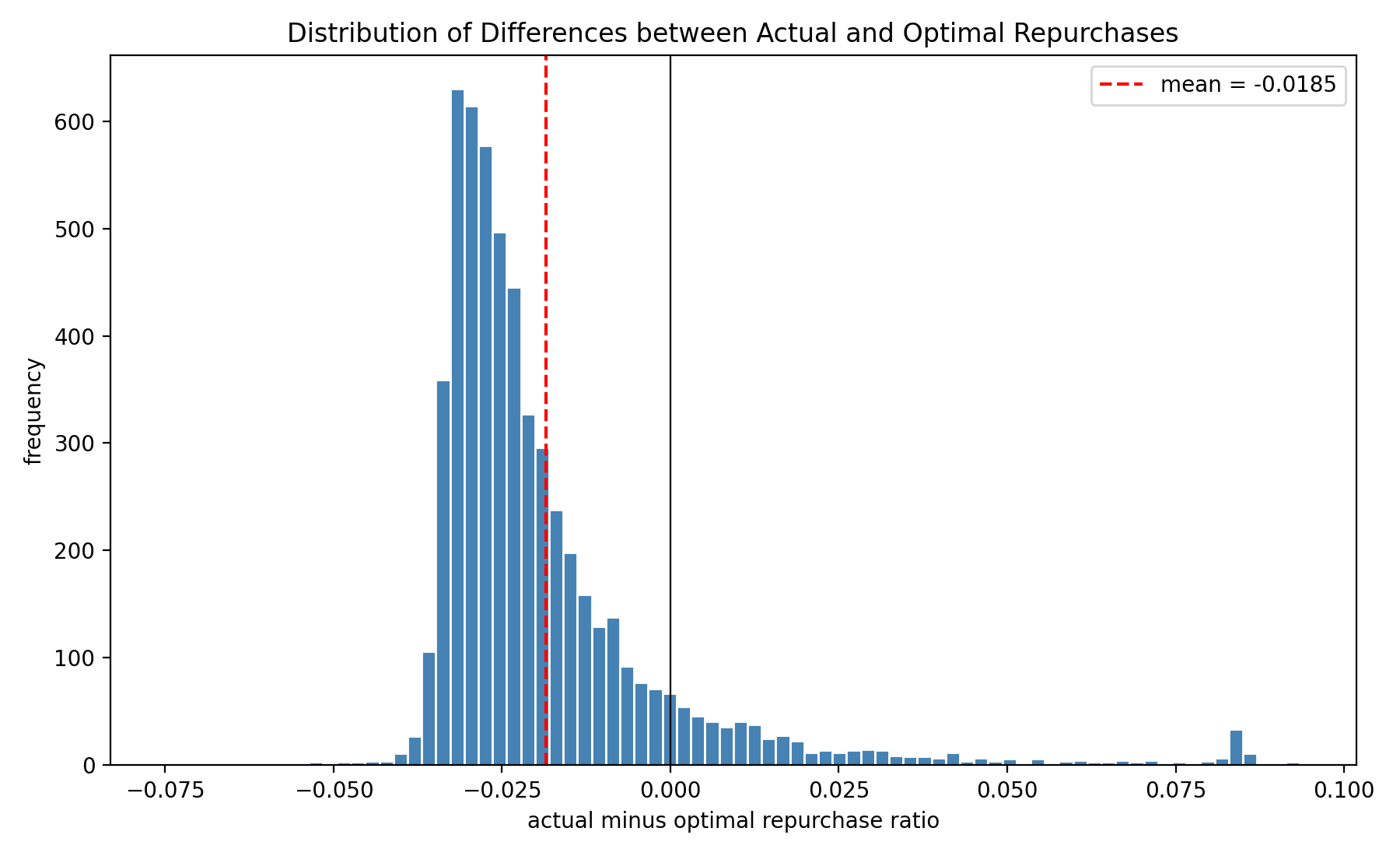}
    \caption{Distribution of Actual-Optimal Difference}
    \label{fig:repurchase-actual-optimal-diff}
\end{figure}

We present the distribution of each repurchase action's actual-optimal difference in Figure~\ref{fig:repurchase-actual-optimal-diff}. While the mean of the difference, or bias, is negative, there is a portion of repurchase cases in which the actual repurchase amount is higher than the optimal level calculated.

\subsubsection{Factors Related to R-bias}

Here we try to find some factors that may explain the R-bias. We average the difference between optimal and actual repurchase ratios by each firm and generate a firm-level R-bias, and use it as the dependent variable. Since R-bias can be positive or negative, we also create another dependent variable by taking the absolute value of R-bias, indicating merely the degree of deviation of actual repurchase from the optimal ratio. We collect each company's total assets in the repurchase year, years of operation (``firm age''), and the number of repurchase announcements that the company has done before this year (``repurchase experience'') as the explanatory variables.

Table \ref{table:bias-factors} presents the regression results. Columns (1) and (2) use the raw R-bias value and Columns (3) and (4) use the absolute value of R-bias as the dependent variables. A higher raw R-bias value represents a higher actual repurchase amount compared to the optimal level, while a higher absolute value of R-bias represents a further deviation of the actual repurchase from the optimal level. Columns (2) and (4) control for year fixed effects that absorb all other year-level variants, while the other two columns do not.

\begin{table}[htbp]
  \centering
  \caption{Factors Related to R-bias}
    \begin{tabular}{lllll}
    \toprule
          & \multicolumn{1}{c}{(1)} & \multicolumn{1}{c}{(2)} & \multicolumn{1}{c}{(3)} & \multicolumn{1}{c}{(4)} \\
    \midrule
          & \multicolumn{2}{c}{R-bias} & \multicolumn{2}{c}{Absolute Value of R-bias} \\
    \midrule
          &       &       &       &  \\
    Firm Age & \multicolumn{1}{c}{-0.199***} & \multicolumn{1}{c}{-0.191***} & \multicolumn{1}{c}{0.063***} & \multicolumn{1}{c}{0.059***} \\
          & \multicolumn{1}{c}{(0.030)} & \multicolumn{1}{c}{(0.030)} & \multicolumn{1}{c}{(0.018)} & \multicolumn{1}{c}{(0.018)} \\
    Repurchase Experience & \multicolumn{1}{c}{0.184} & \multicolumn{1}{c}{-0.315} & \multicolumn{1}{c}{-0.580***} & \multicolumn{1}{c}{-0.539***} \\
          & \multicolumn{1}{c}{(0.263)} & \multicolumn{1}{c}{(0.289)} & \multicolumn{1}{c}{(0.158)} & \multicolumn{1}{c}{(0.177)} \\
    Total Assets & \multicolumn{1}{c}{-0.000***} & \multicolumn{1}{c}{-0.000***} & \multicolumn{1}{c}{0.000} & \multicolumn{1}{c}{0.000} \\
          & \multicolumn{1}{c}{(0.000)} & \multicolumn{1}{c}{(0.000)} & \multicolumn{1}{c}{(0.000)} & \multicolumn{1}{c}{(0.000)} \\
    Tobin's Q & \multicolumn{1}{c}{-2.088***} & \multicolumn{1}{c}{-1.966***} & \multicolumn{1}{c}{0.331} & \multicolumn{1}{c}{0.271} \\
          & \multicolumn{1}{c}{(0.397)} & \multicolumn{1}{c}{(0.397)} & \multicolumn{1}{c}{(0.239)} & \multicolumn{1}{c}{(0.243)} \\
    Intercept & \multicolumn{1}{c}{-7.441***} & \multicolumn{1}{c}{-28.935} & \multicolumn{1}{c}{21.644***} & \multicolumn{1}{c}{33.093***} \\
          & \multicolumn{1}{c}{(0.993)} & \multicolumn{1}{c}{(20.641)} & \multicolumn{1}{c}{(0.599)} & \multicolumn{1}{c}{(12.632)} \\
          &       &       &       &  \\
    Year FE & \multicolumn{1}{c}{NO} & \multicolumn{1}{c}{YES} & \multicolumn{1}{c}{NO} & \multicolumn{1}{c}{YES} \\
          &       &       &       &  \\
    Observations & \multicolumn{1}{c}{2,456} & \multicolumn{1}{c}{2,456} & \multicolumn{1}{c}{2,456} & \multicolumn{1}{c}{2,456} \\
    R-square & \multicolumn{1}{c}{0.035} & \multicolumn{1}{c}{0.075} & \multicolumn{1}{c}{0.010} & \multicolumn{1}{c}{0.021} \\
    \midrule
    \multicolumn{5}{l}{Standard error in parentheses; *** p<0.01, ** p<0.05, * p<0.1} \\
    \multicolumn{5}{l}{R-bias is scaled by 1,000. The intercept in columns (2) and (4) refers to the omitted base year.} \\
    \end{tabular}%
  \label{table:bias-factors}%
\end{table}%

As Table \ref{table:bias-factors} shows, firms with more years of operation have a lower actual repurchase amount relative to the optimal level and deviate more from the optimal policy in absolute terms. Firms with more repurchase experience deviate less from the optimal policy in absolute terms, while their raw R-bias is statistically indistinguishable from zero. Firm size, as measured by total assets, has an economically negligible association with R-bias. Firm valuation, as measured by Tobin's Q, is negatively associated with the raw value of R-bias, and higher-valuation firms repurchase less relative to the optimal level, while its association with the absolute value of R-bias is statistically insignificant. Together, the results indicate that while firms on average repurchase less than the optimal level, more experienced repurchasers stay closer to the optimal policy.

\section{Conclusion}\label{sec:conclusion}

This paper examines causality problems in general MDPs when the observed rewards are biased. Because decisions depend on past data, this interaction generates R-bias. We propose IV-RL algorithms (IV-Q-learning and IV-AC) to correct R-bias and analyze them through the lens of SA with iterate-dependent Markovian structures.

Our paper is a step toward a rich research program of IV-RL algorithms. For example, we have not considered the financial cost of creating IVs. If firms can create different instruments, it will be useful to examine the ``optimal'' instruments (in terms of financial costs rather than in a statistical sense).
Exploring these topics would form a rich avenue for future research.

\begin{appendix}
\numberwithin{equation}{section}

\newpage

\bigskip
\begin{center}
    \LARGE {\textbf{Appendices}}
\end{center}

\section{The IV-AC Algorithm}\label{app:IV-AC}

\subsection{Actor-Critic}

Recall from  Remark~\ref{comment:Q-convergence}
that, when behavior policy is fixed, Q-learning with linear value functions requires the discount factor $\gamma$ to be sufficiently small. This issue can be addressed by policy improvement, i.e., alternating between \emph{evaluating a policy} and \emph{improving the policy}. The former means estimating the Q-function of a policy $\pi$, whereas the latter means finding a better policy $\pi'$ based on the estimate of $Q^\pi$ such that $Q^{\pi'}$ has a higher value than $Q^\pi$ on average. This pattern is commonly referred to as \emph{generalized policy iteration} (GPI), which is used by most RL algorithms in practice.

A leading class of RL algorithms with GPI is Actor-Critic, where two interdependent learning processes are executed by a ``critic'' and an ``actor''. The critic performs policy evaluation,
aiming to learn the Q-function associated with the policy that the actor currently undertakes;
meanwhile, the actor performs policy improvement,
aiming to learn a better policy based on the estimate  of the Q-function provided by the critic.

In Actor-Critic algorithms, both Q-functions and policies are parameterized.
For a behavior policy $\pi$ (we will discuss parameterization of $\pi$ later),
let the estimate of $Q^\pi(s,a)$ be $\widehat{Q}^\pi_t(s,a) =  \phi(s,a)^\intercal \theta_{\SFC,t}$ for some parameter value $\theta_{\SFC,t}$.
Because $Q^\pi$ satisfies the Bellman expectation equation
$Q^\pi(s,a) =  r(s,a)+\gamma \E [Q^\pi(S',A')]$, the critic therefore updates $\theta_{\SFC,t}$ in a way aiming to solve this equation.
An argument similar to that leading to
\eqref{eq:Q-learning-algorithm} implies that the updating rule for policy evaluation is given by
\begin{equation}\label{eq:critic}
    \theta_{\SFC,t+1} = \theta_{\SFC, t} +  {\alpha_{\SFC,t}}  \left[R_t + \gamma  \phi(S_{t+1},A_{t+1})^\intercal \theta_{\SFC,t} - \phi(S_t,A_t)^\intercal  \theta_{\SFC,t}\right]\phi(S_t,A_t).
\end{equation}

Next, for policy improvement, suppose that the actor selects behavior policies from a class of random policies $\pi_{\theta_{\SFA}}$, parameterized by $\theta_{\SFA} \in \Theta_{\SFA}$ for some compact set $\Theta_{\SFA}$. Let the behavior policy in period $t$ be $\pi_{\theta_{\SFA,t}}$ for some parameter value $\theta_{\SFA,t}$. The updating rule for $\theta_{\SFA,t}$ takes the form of \emph{stochastic gradient ascent}:
$\theta_{\SFA,t+1} = \theta_{\SFA,t} + \alpha_{\SFA,t} \widehat{\nabla J(\theta_{\SFA,t})}$,
where $\widehat{\nabla J(\theta_{\SFA,t})}$ is a stochastic estimate of the gradient of the performance objective $J(\theta_{\SFA})$.
In light of the MDP's objective \eqref{eq:Case2}, the performance objective is
the expected cumulative discounted reward, that is, $J(\theta_{\SFA}) = \E_{\nu_{\theta_\SFA}}[Q^{\pi_{\theta_\SFA}}(S_t,A_t)]$, where $\nu_{\theta_{\SFA}}$ denotes the stationary distribution of $(S_t,A_t)$ induced by policy $\pi_{\theta_{\SFA}}$.

To construct the stochastic gradient $\widehat{\nabla J(\theta_{\SFA,t})}$, note that because $\theta_{\SFA}$ affects $J(\theta_{\SFA})$ via the distribution $\nu_{\theta_{\SFA}}$ (through the behavior policy $\pi_{\theta_{\SFA}}$), a standard approach is the score-function method \citep{LEcuyer90}.
When applied to RL settings, this method results in the policy gradient theorem \citep[Section~13.2]{SuttonBarto18},
which asserts that
\begin{equation}\label{eq:J_gradient}
    \nabla J(\theta_{\SFA}) = \E_{\nu_{\theta_{\SFA}}}\biggl[Q^{\pi_{\theta_\SFA}}(S_t,A_t) \frac{\nabla_{\theta_{\SFA}} \pi_{\theta_{\SFA}}(A_t|S_t)}{\pi_{\theta_{\SFA}}(A_t|S_t)}\biggr].
\end{equation}
This leads to the following updating rule used by the actor:
\begin{equation}\label{eq:actor}
    \theta_{\SFA,t+1} = \theta_{\SFA,t} +
    {\alpha_{\SFA,t}} \phi(S_t,A_t)^\intercal\theta_{\SFC,t}\,\nabla_{\theta_{\SFA}}\ln \pi_{\theta_{\SFA,t}}(A_t|S_t),
\end{equation}
where the actor constructs the update by using the estimate provided by the critic in period $t$, namely, $\widehat{Q}^\pi_t(s,a) =  \phi(s,a)^\intercal \theta_{\SFC,t}$ for $\pi = \pi_{\theta_{\SFA,t}}$.

\begin{remark}\label{AC-mutual dependence}
The updating rules \eqref{eq:critic} and \eqref{eq:actor} demonstrate the mutual dependence  between the actor and the critic. The dependence of $\theta_{\SFA,t+1}$ on $\theta_{\SFC,t}$ is evident in \eqref{eq:actor}. To see the dependence in the other direction, notice that in \eqref{eq:critic}, the actor affects the critic through the data (i.e., state-action pairs), which is generated under the behavior policy  $\pi_{\theta_{\SFA,t}}$.
\end{remark}

Finally, similar to Q-learning, when $\{\theta_{\SFA,t}, \theta_{\SFC,t}\}$ converges, the limit $\{\theta_{\SFA}^*, \theta_{\SFC}^*\}$ satisfies
\begin{align}
& \E_{\nu_{\theta_\SFA}}\Bigl[ \Bigl(r(S_t, A_t) + \gamma  \phi(S_{t+1},A_{t+1})^\intercal \theta_{\SFC} - \phi(S_t,A_t)^\intercal \theta_{\SFC} \Bigr) \phi(S_t,A_t) \Bigr ] =0,  \label{eq:condition-critic} \\
& \E_{\nu_{\theta_{\SFA}}}[\phi(S_t,A_t)^\intercal  \theta_{\SFC} \nabla_{\theta_{\SFA}} \ln \pi_{\theta_{\SFA}}(A_t|S_t) ] = 0, \label{eq:condition-actor}
\end{align}
provided that $R_t = r(S_t, A_t) + \epsilon_t$ is unbiased.
Similar to that in Q-learning, equation~\eqref{eq:condition-critic} is the projected Bellman expectation equation for approximating $Q^{\pi_{\theta_{\SFA}}}$ with $\phi(S_t, A_t)^\intercal \theta_{\SFC}$.
Equation~\eqref{eq:condition-actor} stems from the fact that the actor seeks an zero of the policy gradient $\nabla J(\theta_{\SFA})$ given by \eqref{eq:J_gradient}, but with the guidance from the critic.

\subsection{Incorporating IVs into Actor-Critic}\label{subsec:IV-AC-algo}

Similar to IV-Q-learning in Section \ref{subsec:IV}, IVs can be incorporated into Actor-Critic by modifying the state-value function equations.
While maintaining the updating rules for the actor \eqref{eq:actor} and the IV coefficient \eqref{eq:IV-iterate}, we modify the critic's updating rule \eqref{eq:critic} as
\begin{equation}\label{eq:critic-debiased}
    \theta_{\SFC,t+1} = \theta_{\SFC, t} +  {\alpha_{\SFC,t}}  \left[R_t + \gamma  \phi(S_{t+1},A_{t+1})^\intercal \theta_{\SFC,t} - \phi(S_t,A_t)^\intercal  \theta_{\SFC,t}\right]\theta_{\SFI,t} Z_t.
\end{equation}
The complete procedure is detailed in Algorithm \ref{algo:IV-AC}.

\begin{algorithm}[t]
\caption{IV-AC}\label{algo:IV-AC}
\SetAlgoLined
\DontPrintSemicolon
\small{
\SetKwInOut{Input}{Input}
\SetKwInOut{Output}{Output}
\Input{Features $\phi$, policy class $\pi_{\theta_{\SFA}}$,  minibatch size $B$,
and learning rates $(\alpha_{\SFA,k},\alpha_{\SFC,k}, \alpha_{\SFI,k})$}
Initialize $\theta_{\SFA,0}$, $\theta_{\SFC,0}$, $\theta_{\SFI,0}$, and $S_0$ at random, and take action $A_0 \sim \pi_{\theta_{\SFA,0}}(\cdot|S_0)$.\;
\For{all $k=0,1,\ldots $}{
\For{all $t=kB,kB+1,...,(k+1)B-1$ }{
Observe $(Z_t, R_t, S_{t+1})$
and  take action $A_{t+1} \sim \pi_{\theta_{\SFA,k}}(\cdot|S_{t+1})$.\;
}
Update $\theta_{\SFA,k}$, $\theta_{\SFC,k}$, and $\theta_{\SFI,k}$ via
\vspace{-1ex}
\begin{align*}
    \theta_{\SFA,k+1} ={}& \theta_{\SFA,k} +
    \frac{\alpha_{\SFA,k}}{B}\sum_{t=kB}^{(k+1)B-1} \phi(S_t,A_t)^\intercal\theta_{\SFC,k}\,\nabla_{\theta_{\SFA}}\ln \pi_{\theta_{\SFA,k}}(A_t|S_t), \\
    \theta_{\SFC,k+1} ={}& \theta_{\SFC, k} + \frac{\alpha_{\SFC,k}}{B} \sum_{t=kB}^{(k+1)B-1}\bigl(R_t + \gamma  \phi(S_{t+1},A_{t+1})^\intercal\theta_{\SFC,k} - \phi(S_t,A_t)^\intercal  \theta_{\SFC,k}\bigr)  \theta_{\SFI,k} Z_t , \\
    \theta_{\SFI,k+1} ={}& \theta_{\SFI,k} + \frac{\alpha_{\SFI,k}}{B}\sum_{t=kB}^{(k+1)B-1} \bigl( \phi(S_t,A_t)-\theta_{\SFI,k} Z_t \bigr) Z_t^\intercal.
\end{align*}\;
\vspace{-3ex}
Perform projection: $\theta_{\SFX,k+1} \leftarrow \Pi_{\Theta}(\theta_{\SFX,k+1})$ for $\SFX\in \{\SFA,\SFC,\SFI\}$.
}
}
\end{algorithm}

Let $\{\theta_{\SFA}^*, \theta_{\SFC}^*, \theta_{\SFI}^*\}$ denote the limiting values of the IV-AC algorithm at convergence.
These values satisfy
\begin{equation}\label{eq:iv-2}
\left\{
\begin{aligned}
    & \E_{\nu_{\theta_{\SFA}}}\bigl[ \bigl(R_t  + \gamma \phi(S_{t+1},A_{t+1})^\intercal\theta_{\SFC} - \phi(S_t,A_t)^\intercal \theta_{\SFC}  \bigr) \theta_{\SFI} Z_t \bigr ] =0, \\
    & \E_{\nu_{\theta_{\SFA}}}[\phi(S_t,A_t)^\intercal  \theta_{\SFC} \nabla_{\theta_{\SFA}} \ln \pi_{\theta_{\SFA}}(A_t|S_t) ] = 0, \\
    & \E_{\nu_{\theta_{\SFA}}}[(\phi(S_t,A_t) - \theta_{\SFI} Z_t) Z_t^\intercal ] = 0,
\end{aligned}
\right.
\end{equation}
where $\nu_{\theta_{\SFA}}$ is the stationary distribution of $(S_t,A_t,Z_t)$ induced by the behavior policy $\pi_{\theta_{\SFA}}$.

\subsection{Theoretical Properties of IV-AC}\label{subsec:IV-AC-result}

\begin{assumption}\label{assump:phi-ac}
    \begin{enumerate}[label=(\roman*), noitemsep, topsep=0pt]
        \item  $\phi(S_t,A_t)$ is a vector of linearly independent random variables
            under the distribution $\nu_{\theta_{\SFA}}$ for all $\theta_{\SFA}\in\Theta_{\SFA}$ in the case of IV-AC (Algorithm~\ref{algo:IV-AC}).
        \item $\|\phi(s,a)\| \leq C(\norm{s}^{\ell_\phi}+\norm{a}^{\ell_\phi}+1)$ for some $C,\ell_\phi>0$, and  all $s\in\calS$, $a\in\calA$.
    \end{enumerate}
\end{assumption}

\begin{assumption}\label{assump:IV-AC}
Let
    $\theta \coloneqq (\theta_{\SFA}^\intercal, \theta_{\SFC}^\intercal, \mathrm{vec}{(\theta_{\SFI}})^\intercal)^\intercal$, $\Theta \coloneqq \Theta_{\SFA} \times \Theta_{\SFC}  \times \Theta_{\SFI} \subset \Real^{p+p+pq}$.
\begin{enumerate}[label=(\roman*), noitemsep, topsep=0pt]
\item  \label{asp:W_def_AC}
$W_t^{\mathsf{IVAC}}$ (defined in \eqref{eq:W_IVAC}) and $\Theta$ satisfy Assumption~\ref{assump:ergodic}.

\item \label{asp:growth-AC}
$\|\nabla_{\theta_{\SFA}} \ln \pi_{\theta_\SFA}(a|s)\| \leq C(1+\|s\|^{\ell_\pi}+\|a\|^{\ell_\pi})$ for some $C, \ell_\pi >0$ and all $s\in\calS$, $a\in\calA$, $\theta_{\SFA}\in\Theta_{\SFA}$.
\item \label{asp:contraction_AC}
    $H(\theta_\SFA) \coloneqq   \E_{\nu_{\theta_{\SFA}}} [
    \bar{\theta}_{\SFC}(\theta_{\SFA})^\intercal \phi(\theta_{\SFA}(S_t, A_t) \nabla_{\theta_{\SFA}} \ln \pi_{\theta_\SFA}(A_t|S_t)] $ is Lipschitz continuous, where $\bar{\theta}_{\SFC}(\theta_{\SFA}) $  is defined in \eqref{eq:theta_C_bar};  and for some $\psi_\SFA>0$ and all $\theta_{\SFA} \in \Theta_{\SFA}$,
    \begin{equation}\label{eq:contraction-AC}
        (H(\theta_\SFA) - H(\theta_\SFA^*))^\intercal (\theta_\SFA-\theta_{\SFA}^*)\leq -\psi_{\SFA}\|\theta_\SFA-\theta_{\SFA}^*\|^2.
    \end{equation}
\item \label{asp:bar_theta_c}
Both $\bar{\theta}_{\SFC}(\theta_{\SFA}) $  and  $\pi_{\theta_\SFA}(a|s)>0$ are twice continuously differentiable with respect to $\theta_{\SFA}$ for all $s\in\calS$,  $a\in\calA$, $\theta_{\SFA}\in\Theta_{\SFA}$.

\end{enumerate}
\end{assumption}

Assumption~\ref{assump:IV-AC} verifies the conditions of Theorem~\ref{theo:asymp} for IV-AC. Similar to Assumption~\ref{assump:IV-Q}, most conditions are self-explanatory except for condition~\eqref{eq:contraction-AC}, which is critical for verifying Assumption~\ref{assump:G-bar}\ref{asp:SA_nsd}.
(Here, $H(\theta_{\SFA})$ represents an approximate policy gradient.)
However, unlike the condition~\eqref{eq:contraction-Q} for IV-Q-learning,
the condition \eqref{eq:contraction-AC} does not involve $\gamma$. Therefore, IV-AC does not suffer from the drawback of IV-Q-learning concerning the discount factor and the behavior policy (see the numerical results in Section~\ref{sec:numerical}).

\begin{corollary}\label{corollary:CLT-AC}
In Algorithm~\ref{algo:IV-AC},
let $\alpha_{\SFX,k}=\alpha_{\SFX,0} k^{-\delta}$ for some $\alpha_{\SFX,0}>0$, $\delta\in(\frac{1}{2}, 1]$, and all $k\geq 1$, $\SFX\in \{\SFA,\SFC,\SFI\}$.
Suppose that (i) $\{\theta_{\SFA}^*,\theta_{\SFC}^*, \theta_{\SFI}^*\}$ is the unique solution to equation~\eqref{eq:iv-2} in the interior of $\Theta$ and $\bar{\theta}_{\SFC}(\theta_{\SFA}^*) = \theta_{\SFC}^*$,
(ii) Assumptions~\ref{assump:IV}, \ref{assump:phi-ac}, and \ref{assump:IV-AC}  hold,
(iii) $\min\{\psi_{\SFA} \alpha_{\SFA,0}, \psi_{\SFI} \alpha_{\SFI,0}\}>2$ when $\delta=1$, and (iv) $\frac{\alpha_{\SFC,0}}{\alpha_{\SFA, 0}}$ is  large enough.
If $B/T^\frac{\delta}{1+\delta}\rightarrow 0$,
then
\[(B^{1-\delta} T^\delta)^{\frac{1}{2}}\biggl(
\binom{\theta_{\SFA,\lfloor T/B\rfloor} }{\theta_{\SFC,\lfloor T/B\rfloor} }
- \binom{\theta_{\SFA}^*}{\theta_{\SFC}^*}
\biggr) \rightsquigarrow \calN(0,\Gamma_{\mathsf{AC}}^*)\quad \mbox{as } T\to\infty.\]
\end{corollary}

The asymptotic covariance matrix $\Gamma^*_{\mathsf{AC}}$ is defined in Appendix~\ref{app:Cov}.
The result that $\theta_{\SFA,k}$ and  $\theta_{\SFC,k}$ \emph{simultaneously} converge at the parametric rate (when $\delta=1$) contrasts with prior analysis of Actor-Critic algorithms \citep{WuZhangXuGu20}, which implies that the actor converges at a faster rate than the critic, so at most one of them can achieve the parametric rate.
This is because Actor-Critic algorithms usually adopt a two-timescale assumption. The actor uses a learning rate that  diminishes at a faster speed than the critic's learning rate;
for example, $\alpha_{\SFA,k} \propto k^{-\delta}$ and $\alpha_{\SFC,k}\propto k^{-\delta'}$ with $\frac{1}{2} < \delta' < \delta \leq 1$ \citep{KondaTsitsiklis03}. Under this two-timescale assumption, the actor effectively operates on a slower timescale than the critic does.
Thus, the actor's current policy appears essentially static from the critic's perspective, while the critic's iterative policy evaluation appears nearly converged from the actor's perspective.
Since the asymptotic performances of the actor and critic are determined by their respective learning rates,
$\theta_{\SFC,k}$ converges at a rate of $k^{-\frac{\delta'}{2}}$, which is slower than the parametric rate even if $\delta=1$.
In the IV-AC algorithm,
we instead assume that both $\alpha_{\SFA, k}$ and $\alpha_{\SFC,k}$ are proportional to $k^{-\delta}$ and that $\frac{\alpha_{\SFC,0}}{\alpha_{\SFA,0}}$ is sufficiently large.
The latter condition is crucial.
It serves a purpose similar to that of the two-timescale assumption (i.e., $\frac{\alpha_{\SFC,k}}{\alpha_{\SFA,k}} \to \infty$),
but without compromising the convergence rate of the critic.

\section{Embedding IV-RL Algorithms in SA Framework} \label{app:SA}

For this section, we discuss how to map the IV-Q-learning and IV-AC algorithms into the framework of SA defined in equation \eqref{eq:update-scheme-projection}, i.e., how the terms $G$ and $M$ are defined in \eqref{eq:update-scheme-projection} for the two IV-RL algorithms.

\textbf{IV-Q-learning}. Define $\alpha_k=\alpha_{\SFQ,k}$,
$G(W_t,\theta_k) = {\small \begin{pmatrix}
G_\SFQ(W_t,\theta_k)\\
\mathrm{vec}(G_{\SFI}(W_t,\theta_k))
\end{pmatrix}}$, and
$M_k = {\small \begin{pmatrix}
M_{\SFQ, k} \\
M_{\SFI, k}
\end{pmatrix}}$,
where $G_{\SFI}(W_t,\theta_k) = \frac{\alpha_{\SFI,0}}{\alpha_{\SFQ,0}}(\phi(S_t,A_t)-\theta_{\SFI,k} Z_t)Z_t^\intercal$, $M_{\SFI,k}=0$, and
\begin{align*}
    & G_{\SFQ}(W_t,\theta_k) = \Bigl(R_t+\gamma \max_{a'\in \mathcal{A}}\phi(S_{t+1},a')^\intercal \theta_{\SFQ}^*  - \phi(S_t,A_t)^\intercal \theta_{\SFQ}^* \Bigr)\theta_{\SFI,k} Z_t \nonumber\\
    & \qquad \qquad \qquad  + \Bigl(\gamma \max_{a'\in \mathcal{A}}\phi(S_{t+1},a')^\intercal\theta_{\SFQ,k} - \max_{a'\in \mathcal{A}}\phi(S_{t+1},a')^\intercal\theta_{\SFQ}^*\Bigr)\theta_{\SFI}^* Z_t \nonumber \\
    & \qquad \qquad \qquad  - \phi(S_t,A_t)^\intercal (\theta_{\SFQ,k}-\theta_{\SFQ}^*)\theta_{\SFI}^* Z_t, \nonumber\\
    & M_{\SFQ,k} =  \frac{\alpha_{\SFQ,k}}{B}\sum_{t=kB}^{(k+1)B-1} \Bigl(\gamma \max_{a'\in \mathcal{A}}\phi(S_{t+1},a')^\intercal (\theta_{\SFQ,k}-\theta_{\SFQ}^*) - \\ & \qquad \qquad \qquad \qquad \qquad \phi(S_t,A_t)^\intercal (\theta_{\SFQ,k}-\theta_{\SFQ}^*) \Bigr)   (\theta_{\SFI,k}-\theta_{\SFI}^*)Z_t.
\end{align*}

\textbf{IV-AC}.
Define
$\theta =
\small{
\begin{pmatrix}
\theta_{\SFA} \\  \theta_{\SFC} - \bar{\theta}_{\SFC}(\theta_{\SFA}) \\ \mathrm{vec}(\theta_{\SFI})
\end{pmatrix}}$,
$\theta^* =
{\small \begin{pmatrix}
\theta_{\SFA}^* \\  0 \\ \mathrm{vec}(\theta_{\SFI}^*)
\end{pmatrix}}$ and
$\theta_k =
{\small \begin{pmatrix}
\theta_{\SFA,k} \\ \theta_{\SFC,k}-\bar{\theta}_{\SFC,k} \\ \mathrm{vec}(\theta_{\SFI,k})
\end{pmatrix}}$,
where
\[
\bar{\theta}_{\SFC,k}\coloneqq \bar{\theta}_{\SFC}(\theta_{\SFA,k}) =  \E_{\nu_{\theta_{\SFA,k}}}[(\phi(S_t, A_t) \phi(S_t, A_t)^{-1}] \E_{{\nu_{\theta_{\SFA,k}}}}[Q^{\pi_{\theta_{\SFA,k}}}(S_t, A_t)\phi(S_t, A_t)].
\]

Define $\alpha_k=\alpha_{\SFA,k}$,
$
G(W_t,\theta_k) = {\small \begin{pmatrix}
G_\SFA(W_t,\theta_k)\\
G_\SFC(W_t,\theta_k)\\
\mathrm{vec}(G_{\SFI}(W_t,\theta_k))
\end{pmatrix}} $
and
$M_k = \small{\begin{pmatrix}
M_{\SFA, k} \\
M_{\SFC, k} \\
M_{\SFI, k}
\end{pmatrix}}$,
where
\begin{align}
    G_\SFA(W_t,\theta_k) ={}& \phi(S_t,A_t)\theta_{\SFC,k}\nabla_{\theta_{\SFA}}\ln \pi_{\theta_{\SFA,k}}(A_t|S_t),\nonumber\\
    G_\SFC(W_t,\theta_k) ={}& -\Lambda_G(\theta_{\SFA,k}) ( \phi(S_t,A_t)^\intercal \theta_{\SFC,k})\nabla_{\theta_{\SFA}}\ln \pi_{\theta_{\SFA,k}}(A_t|S_t)\nonumber \\
    &+ (R_t+\gamma \phi(S_{t+1},A_{t+1})^\intercal \bar{\theta}_{\SFC,k}-\phi(S_t,A_t)\bar{\theta}_{\SFC,k})\theta_{\SFI,k}Z_t \nonumber\\
    & +  \frac{\alpha_{\SFC,0}}{\alpha_{\SFA,0}}[(\gamma \phi(S_{t+1},A_{t+1})-\phi(S_t,A_t))^\intercal (\theta_{\SFC,k}-\bar{\theta}_{\SFC,k})\theta_{\SFI}^*]Z_t \nonumber\\
    G_\SFI(W_t,\theta_k) ={}& \frac{\alpha_{\SFI,0}}{\alpha_{\SFA,0}}(\phi(S_t,A_t) - \theta_{\SFI,k}Z_t)Z_t^\intercal,\nonumber
\end{align}
where $\Lambda_G(\theta_{\SFA})\coloneqq \frac{\partial}{\partial \theta_{\SFA}}\bar{\theta}_{\SFC}(\theta_{\SFA})$,
$M_{\SFA,k}= 0$, $M_{\SFI,k}=0$,
and
\begin{align*}
    M_{\SFC,k} ={}& \bar{\theta}_{\SFC,k+1} - \bar{\theta}_{\SFC,k} - \Lambda_G(\theta_{\SFA,k})(\theta_{\SFA,k}-\theta_{\SFA,k+1}) \\
    & + \frac{\alpha_{\SFC,k}}{B}\sum_{t=kB}^{(k+1)B-1}(\gamma \phi(S_{t+1},A_{t+1})-\phi(S_t,A_t))^\intercal (\theta_{\SFC,k}-\bar{\theta}_{\SFC,k}) (\theta_{\SFI,k}-\theta_\SFI^*)Z_t.
\end{align*}

Further, define
$\Lambda_\phi(\theta_{\SFA})\coloneqq \E_{\nu_{\theta_\SFA}}[\phi(S_t,A_t)\nabla_{\theta_\SFA}\ln \pi_{\theta_{\SFA}}(A_t|S_t)] $.

\section{Additional Details about Corollaries~\ref{corollary:CLT-Q} and \ref{corollary:CLT-AC}} \label{app:Cov}

Below, we define variables used in Section~\ref{sec:apply_SA}.
\begin{align}
W_t^{\mathsf{core}} \coloneqq{}& \{S_t, A_t, R_t, Z_t, S_{t+1}\}, \nonumber  \\
W_t^{\mathsf{IVQ}} \coloneqq{}& \{W_t^{\mathsf{core}},  R_t Z_t, Z_t Z_t^\intercal, (\|S_t\|^{\ell_\phi} + \|A_t\|^{\ell_\phi}+ \|S_{t+1}\|^{\ell_\phi}) \|Z_t\|\}, \label{eq:W_IVQ}\\
W_t^{\mathsf{IVAC}} \coloneqq{}&  \{W_t^{\mathsf{core}}, A_{t+1}, R_t Z_t, Z_t Z_t^\intercal, \norm{S_t}^{\ell_\pi+\ell_\phi}, \|A_t\|^{\ell_\pi+\ell_\phi}, \nonumber \\
& \quad (\|S_t\|^{\ell_\phi} + \|A_t\|^{\ell_\phi}+ \|S_{t+1}\|^{\ell_\phi} + \|A_{t+1}\|^{\ell_\phi}) \|Z_t\|\}, \label{eq:W_IVAC}  \\
\bar{\theta}_{\SFC}(\theta_{\SFA}) \coloneqq{}& \E_{\nu_{\theta_\SFA}}[(\phi(S_t,A_t)\phi(S_t,A_t)^\intercal)]^{-1}\E_{\nu_{\theta_\SFA}}[Q^{\pi_{\theta_\SFA}}(S_t,A_t)\phi(S_t,A_t)]. \label{eq:theta_C_bar}
\end{align}
Here, $\bar{\theta}_{\SFC}(\theta_{\SFA}) $ is  the  coefficient of the projection of $Q^{\pi_{\theta_{\SFA}}}$ onto $\mathsf{span}(\phi)$,  for a given $\theta_{\SFA}$.
Define $\Gamma_{\mathsf{AC}}^*:= {\small \begin{pmatrix}
I & 0 \\
\Lambda_G^* &  I
\end{pmatrix} \widetilde{\Gamma}_{\mathsf{AC}}^*\begin{pmatrix}
I & (\Lambda_G^*)^\intercal \\
0 & I
\end{pmatrix}}$.
Define $\Gamma^*_{\SFQ}$ (resp., $\widetilde{\Gamma}_{\mathsf{AC}}^*$) as \eqref{eq:Sigma} but with  $\Lambda^*$ being replaced by $\Lambda^*_{\SFQ}$ (reps.,  $\widetilde{\Lambda}^*_{\mathsf{AC}}$), where
\begin{gather*}
\Lambda^*_{\SFQ} \coloneqq  \left(\frac{\partial }{\partial \theta_{\SFQ}} \E_\nu\Bigl[\gamma\max_{a'\in \calA}\bigl(\phi(S_{t+1},a')^\intercal \theta_{\SFQ}\bigr) \theta_{\SFI}^* Z_t \Bigr]\Big|_{\theta_{\SFQ} = \theta_{\SFQ}^*} -\E_\nu[\phi(S_t,A_t)\theta_{\SFI}^* Z_t]\right), \\
\widetilde{\Lambda}^*_{\mathsf{AC}}\coloneq
\begin{pmatrix}
\Lambda^*_H  & \Lambda^*_\phi \\
- \Lambda^*_G \Lambda^*_H  & \quad \widetilde{\Lambda}^*_{\mathsf{AC}, 22}
\end{pmatrix},  \quad \Lambda^*_H \coloneqq \nabla H(\theta_\SFA^*), \quad \Lambda^*_G \coloneqq \nabla\bar{\theta}_\SFC(\theta_\SFA^*) \\
\Lambda^*_\phi \coloneqq \E_{\nu_{\theta_{\SFA}^*}}\left[\phi(S_t,A_t)\nabla_{\theta_\SFA}\ln \pi_{\theta_{\SFA}^*}(A_t|S_t)\right]\\
\widetilde{\Lambda}^*_{\mathsf{AC}, 22} \coloneqq  \frac{\alpha_{\SFC,0}}{\alpha_{\SFA,0}}\E_{\nu_{\theta_{\SFA}^*}}[\theta_\SFI ^*Z_t(\gamma\phi(S_{t+1},A_{t+1}) - \phi(S_t,A_t))^\intercal ] - \Lambda^*_\phi.
\end{gather*}

\section{Experimental Design} \label{app:experiment}

The asymptotic covariance matrix $\Gamma^*$ is estimated as follows.
For IV-Q-learning,
we estimate $\Gamma^*_{\SFQ}$ by plugging in consistent estimators of $\Sigma^*$ and $\Lambda_{\SFQ}^*$.
The former is estimated by
$\widehat{\Sigma}\coloneqq \widehat{\Sigma}(0)+ \sum_{l= 1}^\infty \omega_l (\widehat{\Sigma}(l)+\widehat{\Sigma}(l)^\intercal)$,
where
$\widehat{\Sigma}(l)\coloneqq\sum_{t=0}^{T-l}\widehat{G}(W_t,\theta_K)\widehat{G}(W_{t+l},\theta_K)^\intercal$,
$\widehat{G}(W_t,\theta_K)\coloneqq G(W_t,\theta_K)-\frac{1}{T}\sum_{u=0}^{T-1} G(W_u,\theta_K)$,
and the weight $\omega_l$ can be set according to the scheme proposed by
\cite{NeweyWest87}, namely, $\omega_l=(1-\frac{l}{h+1})\mathbb{I}(l\leq h)$, where $h=c T^{\frac{1}{3}}$ for some $c>0$.
The expressions of $G$ for the two IV-RL algorithms are available in Appendix~\ref{app:SA} (Online).
In addition, $\Lambda^*_{\SFQ}$ is estimated via local-linear regression with the SA iterates $\theta_{\SFQ,k}$.
For IV-AC, since $\Gamma^*_{\mathsf{AC}}$ is too involved to compute, we estimate it with weighted sample covariances based on the trajectory of $\theta_k$.
For 2SLS-IPW, we use standard methods for M-estimation to estimate the asymptotic covariance matrix.

The parameters used in the experiments are provided below.

\begin{enumerate}[label=(\roman*), noitemsep, topsep=0pt]
\item State transition: $c_0=0.5$, $c_1=0.4$, $c_2=0.2$, and $\eta_t\sim 0.3 \mathcal{N}(0,1)$.
\item Gaussian policy class: $\sigma_{\SFA} = 0.4$.
\item Reward function:
$r_0=r_1=0$, $r_2=1$, and $r_3=-1$.
\item Reward observation:
$b=0.8$ and $o_t\sim \mathsf{Uniform}[-0.5, 0.5]$. (Experiment results are not sensitive to the distributions of $\eta_t$ and $o_t$.)
\item Initial parameter values: $\theta_{\SFQ,0} = \theta_{\SFC, 0} = (1.0, 0.5, 0.5, 0.5, 0.5, -1.5)$,
and $\theta_{\SFI,0}$ is set as the 2SLS estimate based on 500 random samples drawn from $\pi_{\theta_A,0}$.
($\theta_{\SFQ,0} = \theta_{\SFC,0}$ is sufficiently far away from
$\theta^*_{\SFQ}$, which is identical to $\theta^*_{\SFC}$ in our numerical experiment.)
\item Learning rates: $\delta=1$,
$\alpha_{\SFQ,0} = \alpha_{\SFC,0} = 10.0$ and
$\alpha_{\SFA,0} = \alpha_{\SFI,0} = 1.0$. \footnote{In the simulation, the ratio of $\frac{\alpha_{0,A}}{\alpha_{0,C}}$ is initially set as $\frac{1}{10}$, and the ratio of $\frac{\alpha_{0,A}}{\alpha_{0,I}}$ is set as $1$. The setup of the ratio of these $\alpha_{0}$ parameters is important for the algorithm to be well-behaved. In general, we recommend to set $\alpha_{0,A}$ much smaller than $\alpha_{0,C}$.}
\item Minibatch size: $B=1$ for IV-Q-learning  and $B=2 $  for IV-AC. (Given $T=10000$, a larger $B$ tends to make IV-Q-learning more biased, but does not affect IV-AC much.)
\end{enumerate}

Under our experimental design, the matrix $\Lambda^* \coloneqq \nabla \E[G(W,\theta^*)]$ for IV-Q-learning  is nearly singular, resulting in a large $\alpha_{\SFQ,0}$ (e.g., 500) and a large $T$ (e.g., $4\times 10^5$) for the algorithm to converge.
To accelerate its convergence, we consider an equivalent root-finding problem $\E[PG(W,\theta)] = 0$, where $P$ is an invertible matrix.
Specifically, we use a pre-estimated matrix $P\approx c(\Lambda^*)^{-1}$ to regularize $\Lambda^*$ with some constant $c>0$.

\end{appendix}

\bibliographystyle{apalike}
\bibliography{CausalRL.bib}

\clearpage

\begin{appendix}
\numberwithin{equation}{section}
\renewcommand{\thesection}{\Roman{section}}

\bigskip
\begin{center}
    \LARGE {\textbf{Online Supplemental Material}}
\end{center}

\section{Technical Lemmas}

Define $C_{\nu, \ell} \coloneqq \sup_{\theta\in\Theta} \E_{\nu_\theta}[1 + \|W\|^\ell]$ and
$C_{w,\ell} = \max_{t\geq 0}\E_w [1+\|W_t\|^\ell|W_0=w,\theta_0]$. By Assumption~\ref{assump:ergodic}\ref{asp:boundedness}, we know that 
\begin{equation}\label{eq:constants}
    C_{\nu,\ell}<\infty \textrm{ and } C_{w,\ell}<\infty.
\end{equation}

We introduce the following lemmas in this section.
The proofs of these lemmas are documented in \cite{LiLuoZhang23sharp}.

\begin{lemma}\label{lemma:geo-ergodicity}
Let $\{W_t:t\geq 0\}$ be an irreducible and aperiodic Markov chain, having a unique stationary distribution $\nu$ on the state space $\calW$.
If the chain is geometrically ergodic with respect to $U(\cdot) \coloneqq 1+\|\cdot\|^\ell $,
then
there exist constants $\rho\in[0,1)$ and $C_U>0$ such that for all $t\geq 0$, $w\in \calW$, and function $f:\calW\mapsto \Real $ with $|f(\cdot)|\leq U(\cdot)$, we have
\begin{gather*}
    \biggl| \int_{\calW} f(w_t)  \bigl( \pr_w(W_t\in\dd{w_t}) - \nu(\dd{w_t}) \bigr) \biggr|
    \leq C_U\rho^t U(w).
\end{gather*}
\end{lemma}

\begin{lemma}\label{lemma:change-chain}
Suppose Assumption~\ref{assump:ergodic} holds.
For $t>0$,
let  $\{W_k: 0\leq k\leq t\}$ be
Markov chains governed by iterate-dependent parameters $\theta_0,\theta_1,...,\theta_{t-1}$, respectively.
Then, there exist constants $\rho\in [0,1)$ and $C_U>0$ such for all $w_0\in\calW$, and function $f:\calW\mapsto \Real$ with $|f(\cdot)|\leq U^\frac{1}{2}(\cdot)$, we have
   \begin{align*}
       & \bigl|\mathbb{E}[ f(w_t)|w_0,\theta_0] -\mathbb{E}_{w_l\sim \SFP_{\theta_0}(\cdot|w_{l-1}),l=1,2,...,t}[ f(w_t)|w_0,\theta_0] \bigr|\\
       \leq{}& C_U L_\nu\sum_{s=1}^{t-1}\rho^{t-s+1} \mathbb{E}[\|\theta_s-\theta_0\|(1+\|w_s\|)]
   \end{align*}
\end{lemma}

\begin{lemma}\label{lemma:mixing}
Suppose Assumption~\ref{assump:ergodic} holds.
Let $G:\calW\times \Real^d \mapsto \Real^d$ be a function satisfying  Assumption~\ref{assump:SA_Lip}.
Assume $B> \ln (2 C_U)/|\ln \rho|$.
Then, for all $k\geq 0$ and $w\in\calW$,
\begin{align*}
  \E_w \biggl[ \biggl\|
    \sum_{t=kB}^{(k+1)B-1} \bigl( G(W_t,\theta_k) - \bar{G}(\theta_k)\bigr) \biggr\|^2  \biggr] \leq \frac{2\widetilde{C}_{w,1}}{1-\rho} B,
\end{align*}
where
$\widetilde{C}_{w,1} \coloneqq \max\bigl\{d L^2 C_U ( 2 C_{w,2} + C_{\nu,1} C_{w,1}), \;  4 L^2 (C_{w,2} + C_{\nu,2})\bigr\}$.
\end{lemma}

\begin{lemma} \label{lemma:theta_k-diff-bounds}
For all $k\geq 1$, let $\alpha_k = \alpha_0 k^{-\delta}$ for some $\alpha_0>0$ and $0<\delta \leq 1$.
Then, under the same assumptions of Lemma~\ref{lemma:mixing},
$\E_w[\|\theta_{k+1} - \theta_k\|^2 ] \leq \widetilde{C}_{w, 2} \alpha_k^2$ for all $k\geq 0$ and $w\in\calW$,
where
$\widetilde{C}_{w, 2}  \coloneqq  4 L^2 C_{w,2} + 2 C_M (\alpha_0 + \alpha_0^2)$.
\end{lemma}

\begin{lemma}\label{lemma:expectation}
Suppose the assumptions of Lemma~\ref{lemma:uni-SA} hold.
Then, there exist positive constants $C_{\theta,1}$, $C_{\theta,2}$, and $K_0$ such that
for all $k\geq K_0$ and $w\in\calW$,
\begin{equation}\label{eq:risk-bound-1}
\E_w[\|\theta_k - \theta^*\|^2] \leq \frac{C_{\theta,1} \ln k}{B k^\delta}  + \frac{C_{\theta,2}\ln k}{k^{2\delta}}.
\end{equation}
Moreover, if $B \geq \ln T/ |\ln \rho|$, then the $\ln k$ term can be removed.
\end{lemma}

\begin{lemma}\label{lemma:uni-SA}
Fix $\theta_0\in\Theta$ and let $\{\theta_k: k\geq 1 \}$ be the SA iterates following \eqref{eq:update-scheme-projection}, where $\alpha_k = \alpha_0 k^{-\delta}$ for some $\alpha_0>0$ and $\delta \in (\frac{1}{2},  1]$.
Suppose that (i) Assumptions~\ref{assump:ergodic}, \ref{assump:SA_Lip}, \ref{assump:G-bar}\ref{asp:SA_unique}, \ref{assump:G-bar}\ref{asp:SA_nsd},
and  \ref{assump:SA_error}
hold,
(ii) $\psi\alpha_0>2$ in the case of $\delta=1$. 
Then, for any $w\in\calW$,  $C>0$, and $\delta_c\in[0, \delta-\frac{1}{2})$,
there exists a constant $C_\theta>0$ depending on  $(\delta,w, C, c,\delta_c)$ such that  for all  sufficiently large $k_0$,
with $\pr_w(\cdot) \coloneqq \pr(\cdot | W_0=w)$,
\[
\pr_w\bigl(\| \theta_k-\theta^*\|\leq C k^{-\delta_c},\; \forall k\geq k_0\bigr) \geq 1 - C_\theta  k_0^{-(2\delta-1-2\delta_c)} \max\biggl\{\Bigl(\frac{\ln k_0}{B}\Bigr)^2,  1\biggr\}.
\]
\end{lemma}

\section{Proof of Theorem~\ref{theo:asymp}} \label{app:proof-SA}

Let $K_0$ and $T_0$ be the constants defined in Lemma~\ref{lemma:expectation} and
let $K = \lfloor T/B \rfloor$.
By Lemma~\ref{lemma:uni-SA},  $\{\theta_{ \lfloor T / B \rfloor }: T  \geq T_0 \}$ is sufficiently close to $\theta^*$ with probability going to one as $T\to\infty$.
It then follows from Assumption~\ref{assump:G-bar}\ref{asp:SA_taylor} that the projection operator $\Pi_\Theta$ is the identity
for
all $k \geq K_0$.
Hence, $\|\theta_k\|\leq\SFR$ for some $\SFR>0$ (because $\Theta$ is a compact set)  and
\begin{align*}
    \theta_{k+1}-\theta^*  ={}& \theta_k-\theta^*+\alpha_k \Lambda^* (\theta_k-\theta^*)+\alpha_k r_G(\theta_k)+M_k
    +\frac{\alpha_k }{B}
    \sum_{t=kB}^{(k+1)B-1} \widetilde{G}(W_t,\theta_k) ,
\end{align*}
for
all $k \geq K_0$, where
$\Lambda^*\coloneqq \nabla\bar{G}(\theta^*)$,
$r_G(\theta) \coloneqq \bar{G}(\theta) - \Lambda^* (\theta - \theta^*)$, and
$\widetilde{G}(W_t,\theta_k)\coloneqq G(W_t,\theta_k)-\bar{G}(\theta_k)$.
It follows that
\begin{align}
    \theta_K-\theta^* ={}& \underbrace{Q_{K_0,K}(\theta_{K_0} - \theta^*)}_{\Psi_1}+\underbrace{\sum_{k=K_0}^{K-1}Q_{k,K}\alpha_k r_G(\theta_k)}_{\Psi_{2}}+\underbrace{\sum_{k=K_0}^{K-1}Q_{k,K} M_k}_{{\Psi_{3}}}
    \nonumber \\
    & + \sum_{k=K_0}^{K-1}Q_{k,K} \frac{\alpha_k}{B}\sum_{t=kB}^{(k+1)B-1}\widetilde{G}(W_t,\theta_k)
    = \sum_{i=1}^3 \Psi_i + \underbrace{\sum_{k=K_0}^{K-1} \bar{X}_k }_{\Psi_{4}}
    + \underbrace{\sum_{k=K_0}^{K-1} X^*_k }_{\Psi_{5}}, \label{eq:decomp-Q}
\end{align}
where
$Q_{k,K} \coloneqq \prod_{l=k}^{K-1} (I+\alpha_l \Lambda^*)$, $I$ is the identity matrix,
\[
X_k \coloneqq Q_{k,K} \frac{\alpha_k}{B}\sum_{t=kB}^{(k+1)B-1}\widetilde{G}(W_t,\theta_k),\quad
\bar{X}_k  \coloneqq \E[X_k|W_{kB-1}, \theta_k],
\qq{and} X_k^*\coloneqq X_k - \bar{X}_k.
\]

Our strategy to prove Theorem~\ref{theo:asymp} is to show: (i)  $(B^{1-\delta}T^\delta)^{\frac{1}{2}}\|\Psi_i\| \stackrel{p}{\to} 0$ as $T\to\infty$ for all $i=1,\ldots,4$, and (ii) $(B^{1-\delta}T^\delta)^{\frac{1}{2}} \Psi_5 \rightsquigarrow \mathcal{N}(0, \Gamma^*) $  as $T\to\infty$.

\subsection{Analysis of $\|Q_{k,K}\|$ }

By Assumptions~\ref{assump:G-bar}\ref{asp:SA_nsd} and \ref{assump:G-bar}\ref{asp:SA_taylor}, $\Lambda^* \prec 0$ and its eigenvalues are  bounded by $-\psi$, so
\begin{align}
\| Q_{k,K} \| \leq \biggl\|\exp\biggl(\Lambda^* \sum_{l=k}^{K-1} \alpha_l\biggr)  \biggr\| \leq{}&    \exp\biggl(-\psi \sum_{l=k}^{K-1} \alpha_l \biggr)
\leq  \exp\biggl( - \psi \alpha_0 \int_{k}^K u^{-\delta}\dd{u}\biggr). \label{eq:Q_estimate'}
\end{align}

If $\delta\in(\frac{1}{2},1)$, then
$\int_{k}^K u^{-\delta}\dd{u} = \frac{1}{1-\delta} \left(K^{1-\delta} - k^{1-\delta} \right)
    \geq  K^{-\delta} (K - k)$,
where the inequality can be shown via Taylor's expansion.
Moreover, if $\delta=1$, then
$
\int_{k}^K u^{-\delta}\dd{u} = \ln K  - \ln k
$.
Hence, it follows from \eqref{eq:Q_estimate'} that
\begin{align}
\| Q_{k,K} \| \leq
\left\{
\begin{array}{ll}
\exp \left(- \psi\alpha_0  K^{-\delta} (K - k) \right),     & \quad  \mbox{ if } \delta\in(\frac{1}{2},1), \\
\left(\frac{k}{K}\right)^{\psi \alpha_0},      &\quad  \mbox{ if } \delta = 1.
\end{array}
\right. \label{eq:Q_estimate}
\end{align}

The following two lemmas can be proved using basic calculus.

\begin{lemma}\label{lemma:sum-approx}
Fix positive constants $a$, $b$, and $\delta\in(0,1)$.
Then, for all sufficiently large $K$,
$
\sum_{k=K_0}^{K-1} k^{-a} \exp \left(- b K^{-\delta} (K - k) \right)\leq \frac{2}{b} K^{\delta-a}
$.
Further, when $\delta=1$, if $b-a>-1$, then for all $K > K_0$,
$
\sum_{k=K_0}^{K-1} k^{-a} \left(\frac{k}{K}\right)^b\leq  \frac{K^{1-a}}{b - a + 1}$.
\end{lemma}

\begin{lemma} \label{lemma:Q-sum-lim}
Under the setup of Theorem~\ref{theo:asymp}, for any $\delta\in(\frac{1}{2},1]$,
$
\lim_{T\to\infty}BK^{-\delta} = 0$
and
$\lim_{T\to\infty} (B^{1-\delta} T^{\delta})^{\frac{1}{2}}  \sum_{k=K_0}^{K-1} \|Q_{k,K}\| \frac{1}{k^{2\delta}}  = 0$.
\end{lemma}

\subsection{Analysis of $\Psi_1$, $\Psi_2$, and $\Psi_3$.}

For $\Psi_1$, we apply \eqref{eq:Q_estimate} to deduce that
\begin{equation*}
  \|\Psi_1\|
  \leq \|Q_{K_0,K} \| \|\theta_{K_0}-\theta^*\|
  \leq
  \left\{
    \begin{array}{ll}
     2\SFR  \exp(- \psi\alpha_0 K^{-\delta} (K  - K_0)),     & \quad  \mbox{ if } \delta\in(\frac{1}{2},1),\\
     2\SFR  \left(\frac{K_0}{K}\right)^{\psi \alpha_0},      &\quad  \mbox{ if } \delta = 1.
    \end{array}
    \right.
\end{equation*}

Let $\xi \coloneqq B/T^{\frac{\delta}{1+\delta}}$.
Then,
$\xi \to 0$ as $T\to \infty$ by assumption,
and thus,
$K = T/B = \xi^{-1} T^{\frac{1}{1+\delta}} \to \infty$  as $T\to \infty$.
Hence, if $\delta\in(\frac{1}{2},1)$, then
\begin{align*}
(B^{1-\delta}T^\delta)^{\frac{1}{2}} \|\Psi_1\|
\leq 2\SFR (K^\delta B)^{\frac{1}{2}} \exp(-\psi\alpha_0 K^{-\delta} (K -  K_0) )
\leq (K^{\delta} B)^{\frac{1}{2}}\cdot K^{-\delta} = \xi^{\frac{1+\delta}{2}}
\to 0,
\end{align*}
as $T\to\infty$, where the second inequality holds for all sufficiently large $T$.
The case of $\delta=1$ can be analyzed similarly.

For $\Psi_2$, we apply Assumption~\ref{assump:G-bar}\ref{asp:SA_taylor} and  Lemma~\ref{lemma:expectation} to deduce
\begin{align*}
\E_w[\|\Psi_2\|]
\leq C_r \alpha_0 \sum_{k=K_0}^{K-1}  \E_w[ \|Q_{k,K}\| \left(\frac{C_{\theta,1}}{B k^{2\delta}} + \frac{C_{\theta,2}}{k^{3\delta}}\right).
\end{align*}
Applying Lemma~\ref{lemma:Q-sum-lim} leads to
$(B^{1-\delta} T^{\delta})^{\frac{1}{2}} \E_w[\|\Psi_2\|] \to 0$ as $T\to\infty$,
which further implies (according to Markov's inequality) that $\delta\in(\frac{1}{2},1]$,
$(B^{1-\delta}T^\delta)^{\frac{1}{2}} \|\Psi_2\| \stackrel{p}{\to} 0 $ as $T\to\infty$.

By Assumption~\ref{assump:SA_error}, we have $\E_w[\|M_k\|]\leq C_M \alpha_k^2 $.
We can use analysis similar to that
for $\Psi_2$  to show that for any $\delta\in(\frac{1}{2},1]$,
$(B^{1-\delta}T^\delta)^{\frac{1}{2}} \|\Psi_3\| \stackrel{p}{\to} 0 $ as $T\to\infty$.

\subsection{Analysis of $\Psi_4$}

Now, we assume that $B\geq C_\rho \ln T$ for simplification of the proof. For $B < C_\rho \ln T$, we can redefine the minibatch size $\widetilde{B} := C_B B$ for some integer $C_B>0 $ such that $\widetilde{B}\in [C_\rho \ln T, 2C_\rho \ln T)$. Then, we can define $\widetilde{X}_j:= \sum_{k=I_j+1}^{I_{j+1}}Q_{k,K}\frac{\alpha_k}{B}\sum_{t=kB}^{(k+1)B-1}\widetilde{G}(W_t,\theta_k)$, with $I_j:=(j-1)\widetilde{B}$ for $j=1,2,...,\lfloor \frac{K}{C_B}\rfloor$. Since the data-generating process at $t\in [I_j+1,I_{j+1})$ only changes slightly, a martingale CLT can be established for $\{\widetilde{X}_j\}_{1\leq j\leq \frac{K}{C_B}}$ in a way similar to the case of  $\{X_j\}_{1\leq j\leq K}$ when $B\geq C_\rho \ln T$. We abbreviate the proof for that part.

We decompose $\bar{X}_k$ as two components:
\begin{align}
    \bar{X}_k ={}& \underbrace{Q_{k,K} \frac{\alpha_k}{B}\sum_{t=kB}^{(k+1)B  - 1}\E[\widetilde{G}(W_t,\theta_k) | W_{kB-1}, \theta_k]
}_{\bar{X}_{k,1}} \nonumber \\
     & +\underbrace{Q_{k,K} \frac{\alpha_k}{B}\sum_{t=kB }^{(k+1)B-1} \E[\widetilde{G}(W_t,\theta_{k})
     -\widetilde{G}(W_t,\theta_{k-1})| W_{kB-1}, \theta_k]
}_{\bar{X}_{k,2}}.\nonumber
\end{align}

For $\bar{X}_{k,1}$, note that according to Lemma \ref{lemma:change-chain},  for any $t=kB,\ldots,(k+1)B-1$ we have:
\begin{align}
& \|\E[\widetilde{G}(W_t,\theta_{k-1})| \mathscr{F}_{(k-1)B-1}]\| \nonumber \\
 \leq{}&
C_U^2 \rho^{t-(k-1)B+1}L(1+\|W_{(k-1)B-1}\|) +
(C_U + 1) L_{\nu} \E[\|\theta_{k-1}-\theta_k\| | \mathscr{F}_{(k-1)B-1}], \nonumber
\end{align}
which implies that
\begin{align}
& \|\E_w[\widetilde{G}(W_t,\theta_{k-1})]\|  \leq  \E[\|\E_w[\widetilde{G}(W_t,\theta_{k-1})| \mathscr{F}_{(k-1)B-1}]\|]\nonumber\\
\leq{}& C_U^2 \rho^{t-(k-1)B+1} L\E_w[ 1+\|W_{(k-1)B-1}\|] +
(C_U + 1)  L_{\nu} \E_w[\|\theta_{k-1}-\theta_k\| ]  \nonumber \\
\leq{}& C_U^2 \rho^{B} L C_{w,1} +
(C_U + 1)  L_{\nu}
\widetilde{C}_{w,2}^\frac{1}{2}  \alpha_{k-1}, \label{eq:G_tilde}
\end{align}
where the last inequality follows from \ref{lemma:theta_k-diff-bounds}.
Further, note that
$
    \rho^{C_\rho \ln T} =  T^{C_\rho \ln \rho} \leq k^{C_\rho \ln \rho} \leq k^{- \delta} = \alpha_0^{-1} \alpha_k,
$
where the second inequality holds because $C_\rho = 1/|\ln \rho|$ and $\delta\in (0,1]$.  Note that $\alpha_{k-1} = (\frac{k}{k-1})^\delta \alpha_k \leq 2 \alpha_k$ for $k\geq 2$.
It then follows from \eqref{eq:G_tilde} that
$
\|\E_w[\widetilde{G}(W_t,\theta_{k-1})]\|  \leq
(C_U^2 L C_{w,1} \alpha_0^{-1} + 2(C_U + 1)  L_\nu \widetilde{C}_{w,2}^{\frac{1}{2}} ) \alpha_k$.
Hence,
\begin{align*}
    \sum_{k=K_0}^{K-1} \|\E_w [\bar{X}_{k,1}]\|
    \leq{}& \sum_{k=K_0}^{K-1} \|Q_{k,K} \| \frac{\alpha_k}{B} \sum_{t=kB}^{(k+1)B - 1} \| \E_w[\widetilde{G}(W_t, \theta_{k-1}) ]  \|
    \nonumber\\
    \leq{}&  \sum_{k=K_0}^{K-1} \|Q_{k,K}\| \alpha_k^2  \left(C_U^2 L C_{w,1} \alpha_0^{-1} + 2 (C_U + 1)  L_\nu \widetilde{C}_{w,2}^{\frac{1}{2}} \right).
\end{align*}
It then follows from Lemma~\ref{lemma:Q-sum-lim} that
\begin{equation} \label{eq:X_k1_lim}
    \lim_{T\to\infty} (B^{1-\delta} T^{\delta})^{\frac{1}{2}} \sum_{k=K_0}^{K-1} \|\E_w [\bar{X}_{k,1}]\|  \to 0.
\end{equation}

For $\bar{X}_{k,2}$, we first note that
\begin{align}
\|\bar{G}(\theta_k) - \bar{G}(\theta_{k-1})\|
={}& \|\E_{W\sim\nu_{\theta_k}} [G(W,\theta_k)] - \E_{W\sim\nu_{\theta_{k-1}}} [G(W,\theta_{k-1})] \| \nonumber \\
\leq{}& \|\E_{W\sim\nu_{\theta_k}} [G(W,\theta_k)] - \E_{W\sim\nu_{\theta_k}} [G(W,\theta_{k-1})] \|\nonumber \\
{}&+ \|\E_{W\sim\nu_{\theta_k}} [G(W,\theta_{k-1})] - \E_{W\sim\nu_{\theta_{k-1}}} [G(W,\theta_{k-1})] \| \nonumber \\
\leq {}& \E_{W\sim\nu_{\theta_k}} [L(1+\|W\|) \|\theta_k - \theta_{k-1} \| ] + L_\nu \|\theta_k - \theta_{k-1} \| \nonumber \\
\leq{}&  L C_{\nu,1} \|\theta_k - \theta_{k-1} \| + L_\nu \|\theta_k - \theta_{k-1} \|, \label{eq:X_k2_decop-2}
\end{align}
where the second inequality follows from
Assumption~\ref{assump:ergodic}\ref{asp:ergodic_Lip}  and  Assumption~\ref{assump:SA_Lip}.
Therefore,
by the definition of $\widetilde{G}$,
\begin{align*}
    \| \widetilde{G}(W_t, \theta_k) - \widetilde{G}(W_t, \theta_{k-1})\|
\leq{}& \|G(W_t,\theta_{k})- G(W_t,\theta_{k-1})\| +  \| \bar{G}(\theta_{k}) - \bar{G}(\theta_{k-1})\| \\ \leq {}& (L + L C_{\nu,1} + L_{\nu})  (1 + \|W_t\|) \|\theta_k - \theta_{k-1}\|.  \label{eq:X_k2_decop}
\end{align*}
where the second inequality follows from Assumption~\ref{assump:SA_Lip} and \eqref{eq:X_k2_decop-2}.
Therefore,
\begin{align*}
    \E_w[\| \widetilde{G}(W_t, \theta_k) - \widetilde{G}(W_t, \theta_{k-1})\|]
    \leq{}& (L + L C_{\nu,1} + L_{\nu}) \E_w[(1+\|W_t\|)^2]^{\frac{1}{2}} \E_w[\|\theta_k - \theta_{k-1}\|^2]^{\frac{1}{2}} \nonumber \\
    \leq{}& (L + L C_{\nu,1} + L_{\nu}) \left(2 C_{w,2}  \right)^{\frac{1}{2}}  \widetilde{C}_{w,2}^{\frac{1}{2}} \alpha_{k-1},
\end{align*}
where the second inequality follows from  Lemma~\ref{lemma:theta_k-diff-bounds}.
Therefore,
\begin{align*}
    \sum_{k=K_0}^{K-1} \|\E_w [\bar{X}_{k,2}] \| \leq{}& \sum_{k=K_0}^{K-1} \|Q_{k,K} \| \frac{\alpha_k}{B}
    \sum_{t=kB}^{(k+1)B-1} \| \E_w[ \widetilde{G}(W_t, \theta_k) - \widetilde{G}(W_t, \theta_{k-1}) ] \|  \\
    \leq{}&  \sum_{k=K_0}^{K-1} \|Q_{k,K} \| \alpha_k^2
    (L + L C_{\nu,1} + L_{\nu}) \left(2 C_{w,2}  \right)^{\frac{1}{2}}  (2\widetilde{C}_{w,2}^{\frac{1}{2}}).
\end{align*}
It then follows from Lemma~\ref{lemma:Q-sum-lim} that
\begin{equation} \label{eq:X_k2_lim}
    \lim_{T\to\infty} (B^{1-\delta} T^{\delta})^{\frac{1}{2}} \sum_{k=K_0}^{K-1} \|\E_w [\bar{X}_{k,2}]\|  \to 0.
\end{equation}

Combining \eqref{eq:X_k1_lim} and \eqref{eq:X_k2_lim} yields
\begin{align}
    (B^{1-\delta} T^{\delta})^{\frac{1}{2}} \E_w[\|\Psi_4\|] ={}&  (B^{1-\delta} T^{\delta})^{\frac{1}{2}} \biggl\| \sum_{k=K_0}^{K-1} \E_w [\bar{X}_k]  \biggr\|  \nonumber  \\
    \leq{}&
(B^{1-\delta} T^{\delta})^{\frac{1}{2}} \sum_{k=K_0}^{K-1} (\|\E_w [\bar{X}_{k,1}]\| + \|\E_w [\bar{X}_{k,2}]\| ) \to 0,     \label{eq:Psi_4_bound'}
\end{align}
as $T\to\infty$.
Therefore, by Markov's inequality,
\begin{equation}\label{eq:Psi_4_bound}
    \lim_{T\to\infty} (B^{1-\delta}T^\delta)^{\frac{1}{2}} \|\Psi_4\| = 0 \quad \mbox{in probability}.
\end{equation}

\subsection{Analysis of $\Psi_5$ }

By the martingale CLT (Ethier and Kurtz, 1986, p.~339)\footnote{\textsc{Ethier, S. N. and  T. G. Kurtz} (1986): \textit{Markov Processes: Characterization and Convergence}, Wiley.}, to prove $(B^{1-\delta}T^\delta)^{\frac{1}{2}} \Psi_5 \rightsquigarrow \mathcal{N}(0, \Gamma^*) $  as $T\to\infty$, it suffices to verify the following two conditions:
\begin{gather}
\lim_{T\to\infty}(B^{1-\delta}T^\delta)^{\frac{1}{2}}  \E\left[\max_{K_0\leq k\leq K-1} \|X_k^*\|\right] = 0, \label{eq:MG_CLT_condition1} \\
\lim_{T\to\infty} B^{1-\delta}T^\delta \sum_{k=K_0}^{K-1} X^*_k (X^*_k)^\intercal = \Gamma^* \quad \mbox{in probability}. \label{eq:MG_CLT_condition2}
\end{gather}

\subsubsection{Verification of Condition~\eqref{eq:MG_CLT_condition1}.}
Let $X_{K, \max}\coloneqq \max\limits_{K_0 \leq k \leq K-1} \|X_{k}\|$ and $X_{K, \max}^* \coloneqq \max\limits_{K_0\leq k \leq K-1} \|X_{k}^*\|$.
Note that
\[
    \E_w \left[|X_{K, \max} - X_{K, \max}^*|\right]
    \leq \E_w \left[\max_{K_0\leq k\leq K-1} \bigl| \| X_k\| - \|X^*_k \| \bigr|\right]
    \leq \sum_{k=K_0}
    ^{K-1} \E_w\left[\|\bar{X}_k\|\right].
\]
It then follows from \eqref{eq:Psi_4_bound'} that
$(B^{1-\delta} T^\delta)^{\frac{1}{2}} \E_w [|X_{K, \max} - X_{K, \max}^*|] \to 0$ as $T\to\infty$.
Therefore, to verify condition~\eqref{eq:MG_CLT_condition1},
it suffices to show that $ (B^{1-\delta} T^\delta)^{\frac{1}{2}}  \E[X_{K,\max}]\rightarrow 0$ as $T\to\infty$.
To that end,
we will show
\begin{equation} \label{eq:4th-moment-lim}
   \lim_{T\to\infty}( B^{1-\delta} T^\delta)^{2+a} \sum_{K_0 \leq k\leq K}\E_w[\|X_{k}\|^4] = 0,
\end{equation}
for some $a>0$.
To see why, note that for any $b>0$,
\begin{align}
      \E_w[X_{K,\max}]= {}& \E_w[X_{K,\max}\ind(X_{K,\max}<b)]+\E_w[X_{K,\max}\ind(X_{K,\max}\geq b)] \nonumber \\
     \leq{} & b + \int_{x\geq b} \pr\biggl(\max_{K_0 \leq k\leq K} \|X_{k}\|\geq x\biggr)\dd{x}
    \leq  b + \sum_{K_0 \leq k\leq K} \int_{x\geq b} \pr(\|X_{k}\|\geq x)\dd{x}  \nonumber \\
     \leq{} &  b + \int_{x\geq b}\frac{1}{x^4} \sum_{K_0 \leq k\leq K}\E_w[\|X_{k}\|^4] =  b + \frac{1}{3b^3}\sum_{K_0 \leq k\leq K}\E_w[\|X_{k}\|^4], \label{eq:X_max_Markov}
\end{align}
where the third inequality follows from Markov's inequality.
If \eqref{eq:4th-moment-lim} holds, then we may set $b= ( B^\delta T^{1-\delta})^{-\frac{2+a}{4}}$ and use \eqref{eq:X_max_Markov} to deduce
$ (B^{1-\delta} T^\delta)^{\frac{1}{2}}  \E[X_{K,\max}]\rightarrow 0$ as $T\to\infty$.

We now prove \eqref{eq:4th-moment-lim}.
By the definition of $X_k$,
\begin{align}
    \E_w[\|X_k\|^4]\leq {} & \|Q_{k,K}\|^4 \frac{\alpha_k^4}{ B^{4} }\E\biggl[\biggl |\sum_{t=kB}^{(k+1)B-1}\widetilde{G}(W_t,\theta_k)\biggr |^4 \biggr] \nonumber \\
    \leq{}& d \|Q_{k,K}\|^4 \frac{\alpha_k^4}{ B^{4} }  \sum_{i=1}^d \E\biggl[\biggl |\sum_{t=kB}^{(k+1)B-1}\widetilde{G}_i(W_t,\theta_k)\biggr |^4 \biggr], \label{eq:4th-moment-bound}
\end{align}
where $\widetilde{G}_i$ denotes the $i^{th}$ entry of $\widetilde{G}$ for $i=1,\ldots,d$.
Note that for each $i$,
\begin{align}
    \E_w\biggl[\biggl|\sum_{t=kB}^{(k+1)B-1}\widetilde{G}_i(W_t,\theta_k)\biggr|^4\biggr]
     \leq 24\sum_{(t_1,t_2,t_3,t_4)\in\mathcal{M}} \biggl|\E_w\biggl[\prod_{j=1}^4\widetilde{G}_{i}(W_{t_j},\theta_k) \biggr]\biggr|,\label{eq:sum-4}
\end{align}
where
\begin{align*}
    \mathcal{M} \coloneqq {}& \{(t_1,t_2,t_3,t_4): kB \leq t_1 \leq t_2 \leq t_3 \leq t_4 \leq (k+1)B - 1\} \\
    \mathcal{M}_1 \coloneqq {}& \{(t_1,t_2,t_3,t_4) \in \mathcal{M}: t_4 - t_3 > C_\rho \ln T\} \\
    \mathcal{M}_2 \coloneqq {}& \{(t_1,t_2,t_3,t_4) \in \mathcal{M} : t_4 - t_3 \leq  C_\rho \ln T \mbox{ and } t_2 - t_1 > C_\rho \ln T\} \\
    \mathcal{M}_3 \coloneqq {}& \{(t_1,t_2,t_3,t_4) \in \mathcal{M} : t_4 - t_3 \leq  C_\rho \ln T \mbox{ and } t_2 - t_1 \leq C_\rho \ln T\}
\end{align*}
with $C_\rho = 1/|\ln \rho|$.
Then, we break the sum  on the right-hand side of \eqref{eq:sum-4} into three pieces, corresponding to $\mathcal{M}_1$, $\mathcal{M}_2$, and $\mathcal{M}_3$, respectively.

We first consider the case that $(t_1,t_2,t_3,t_4) \in \mathcal{M}_1$.  Note that
\begin{align*}
    \E[\widetilde{G}_i(W_{t_4},\theta_k)|\mathscr{F}_{t_3}] \leq {}& C_U \rho^{t_4 - t_3} L (1+\|W_{t_3}\|) \leq C_U T^{-1} L (1+\|W_{t_3}\|),
\end{align*}
where the first inequality follows from Lemma \ref{lemma:geo-ergodicity} and the second holds because $t_4 - t_3 > C_\rho \ln T$.
Therefore,
\begin{align}
    \biggr|\E_w\biggl[\prod_{j=1}^4\widetilde{G}_{i}(W_{t_j},\theta_k) \biggr]\biggr| \leq{}&  C_U T^{-1} L \biggl|\E_w\biggl[\prod_{j=1}^3\widetilde{G}_{i}(W_{t_j},\theta_k) (1+\|W_{t_3}\|) \biggr]\biggr| \nonumber \\
    \leq{}& \frac{1}{4} C_U T^{-1} L \E_w\biggl[  \sum_{j=1}^3 |\widetilde{G}_{i}(W_{t_j},\theta_k)|^4 +(1+\|W_{t_3}\|)^4 \biggr].   \label{eq:M_1_part1}
\end{align}
Note that for any $t=kB,\ldots, (k+1)B-1$,
\begin{align}
    \| \widetilde{G}(W_{t},\theta_k) \| \leq{}& \| G(W_{t}, \theta_k) \| + \| \bar{G}(\theta_k) \|
    \leq  (L + L  C_{\nu, 1}) (1+\|W_{t}\|),  \label{eq:G_tilde_bound}
\end{align}
where the second inequality follows from  Assumption~\ref{assump:SA_Lip}.
Hence,
\begin{align}
    \E_w[\|\widetilde{G}(W_t, \theta_k)\|^4]
    \leq{}&   (L + L C_{\nu,1})^4  \E_w [(1+\|W_t\|)^4] \nonumber  \\
    \leq{}& 16 (L + L C_{\nu,1})^4  \E_w [1+\|W_t\|^4]  \leq 16 (L + L C_{\nu,1})^4  C_{w,4}, \label{eq:M_1_part2}
\end{align}
where the last inequality follows from Assumption \ref{assump:ergodic} (iv), as this is stated in \eqref{eq:constants}.
Combining \eqref{eq:M_1_part1} and  \eqref{eq:M_1_part2}  yields
\begin{align}
    & \sum_{(t_1,t_2,t_3,t_4)\in \mathcal{M}_1}  \biggl|\E_w\biggl[\prod_{j=1}^4\widetilde{G}_{i}(W_{t_j},\theta_k) \biggr]\biggr| \nonumber\\
    \leq{}&
    \sum_{(t_1,t_2,t_3,t_4)\in \mathcal{M}_1}
     4 C_U T^{-1} L [3(L + L C_{\nu,1})^4 + 1] C_{w,4}  \nonumber\\
      \leq{}&  4 C_U L [3(L + L C_{\nu,1})^4 + 1] C_{w,4} \frac{B^4}{T}, \label{eq:M_1}
\end{align}
where the second step holds because the cardinality of $\mathcal{M}_1$ satisfies  $|\mathcal{M}_1| \leq |\mathcal{M}| \leq B^4$.

Now, we consider the case that $(t_1,t_2,t_3,t_4) \in \mathcal{M}_2$.
Note that
$\E[\prod_{j=2}^4 \widetilde{G}_i(W_{t_j},\theta_k)|\mathscr{F}_{t_2}]$ is a function of $W_{t_2}$ according to the Markov property and we denote it by $f(W_{t_2})$.
Then, it follows from \eqref{eq:G_tilde_bound} that
\begin{align}
    & |f(W_{t_2})| \leq  (L + L C_{\nu, 1})^3 \E\biggl[ \prod_{j=2}^4 (1+\|W_{t_j} \|)\bigg| W_{t_2}\biggr]  \nonumber \\
    \leq{}& \frac{1}{3} (L + L C_{\nu, 1})^3 \sum_{j=2}^4 \E[(1+\|W_{t_j}\|)^3 | W_{t_2} ]
    \leq \frac{8}{3} (L + L C_{\nu, 1})^3 \sum_{j=2}^4 \E[1+\|W_{t_j}\|^3 | W_{t_2} ]   \nonumber \\
    \leq{}&  \frac{8}{3} (L + L C_{\nu, 1})^3 [1 + 2(C_U + C_{\nu,3})] (1 + \|W_{t_2}\|^3) \coloneqq \check{C} (1 + \|W_{t_2}\|^3), \label{eq:f_W}
\end{align}
where the last inequality follows from Lemma~\ref{lemma:geo-ergodicity}.
Further, applying  Lemma~\ref{lemma:geo-ergodicity} to \eqref{eq:f_W},
\begin{align*}
    | \E[f(W_{t_2}) | \mathscr{F}_{t_1}] - \E_{\nu_{\theta_k}}[f(W_{t_2})] |
    \leq{}&   \check{C} C_U \rho^{t_2 - t_1}  (1 + \|W_{t_1}\|^3)
    \leq \check{C} C_U T^{-1} (1 + \|W_{t_1}\|^3),
\end{align*}
since $t_2 - t_1 > C_\rho \ln T$.
It then follows from \eqref{eq:G_tilde_bound} that
\begin{align*}
     & \E_w \Bigl[ \bigl|\widetilde{G}_i(W_{t_1},\theta_k) \bigr|  \bigl|  \E[f(W_{t_2}) | \mathscr{F}_{t_1}] - \E_{\nu_{\theta_k}}[f(W_{t_2})] \bigr| \Bigr] \nonumber \\
     \leq{}& (L+ L C_{\nu,1}) \check{C} C_U T^{-1} \E_w[(1+\|W_{t_1}\|)(1+\|W_{t_1}\|^3)]
     \leq 2 C_{w,4} (L+ L C_{\nu,1}) \check{C} C_U   T^{-1}. %
\end{align*}
Moreover, note that
\begin{align*}
    \biggl|\E_w \Bigl[ \widetilde{G}_i(W_{t_1},\theta_k)    \E_{\nu_{\theta_k}}[f(W_{t_2})]  \Bigr] \biggr|
    \leq \check{C} C_{\nu, 3} \biggl|\E_w \bigl[ \widetilde{G}_i(W_{t_1},\theta_k)   \bigr]\biggr|  \leq \check{C} C_{\nu, 3} C_U  L C_{w,1} \rho^{t_1 - kB} , %
\end{align*}
where the second inequality follows from Lemma~\ref{lemma:geo-ergodicity}.
Therefore,
\begin{align}
   &  \sum_{(t_1,t_2,t_3,t_4)\in \mathcal{M}_2}   \biggl|\E_w\biggl[\prod_{j=1}^4\widetilde{G}_{i}(W_{t_j},\theta_k) \biggr] \biggr|
   =   \sum_{(t_1,t_2,t_3,t_4)\in \mathcal{M}_2}  \biggl| \E_w \bigl[ \widetilde{G}_i(W_{t_1},\theta_k)  \E[ f(W_{t_2}) |\mathscr{F}_{t_1} ] \bigr] \biggr|\nonumber \\
   \leq{}&  \sum_{(t_1,t_2,t_3,t_4)\in \mathcal{M}_2}  \E_w\Bigl[  \bigl|\widetilde{G}_i(W_{t_1},\theta_k) \bigl|  \bigl|  \E[f(W_{t_2}) | \mathscr{F}_{t_1}] - \E_{\nu_{\theta_k}}[f(W_{t_2})] \bigr|   \Bigr] \nonumber \\ & + \biggl|\E_w\bigl[\widetilde{G}_i(W_{t_1},\theta_k) \E_{\nu_{\theta_k}}[f(W_{t_2})]\bigr]  \biggr|\nonumber \\
   \leq{}&  \sum_{(t_1,t_2,t_3,t_4)\in \mathcal{M}_2} \bigl[ 2 C_{w,4} (L+ L C_{\nu,1}) \check{C} C_U   T^{-1} +   \check{C} C_{\nu, 3} C_U  L C_{w,1} \rho^{t_1 - kB}\bigr]. \label{eq:M_2_decop}
\end{align}
Further, note that
\begin{gather}
     \sum_{(t_1,t_2,t_3,t_4)\in \mathcal{M}_2}  1
    \leq \sum_{kB\leq t_1 \leq  t_2\leq t_3 \leq (k+1)B-1}\sum_{t_4 = t_3}^{ t_3 + C_\rho \ln T} 1 \leq  B^3 C_\rho \ln T, \label{eq:M_2_cardinality} \\
     \sum_{(t_1,t_2,t_3,t_4)\in \mathcal{M}_2} \rho^{t_1 - kB} \leq  \sum_{kB\leq t_2 \leq t_3\leq (k+1)B-1}  \sum_{t_4 = t_3}^{ t_3 + C_\rho \ln T}  \sum_{kB \leq t_1 \leq (k+1)B - 1} \rho^{t_1 - kB} \leq \frac{B^2 C_\rho \ln T }{1-\rho}. \nonumber  %
\end{gather}
Hence, it follows from
\eqref{eq:M_2_decop} that
\begin{align}
    & \sum_{(t_1,t_2,t_3,t_4)\in \mathcal{M}_2}  \biggl|\E_w\biggl[\prod_{j=1}^4\widetilde{G}_{i}(W_{t_j},\theta_k) \biggr]\biggr| \nonumber \\
    & \leq 2 C_{w,4}C_\rho (L+ L C_{\nu,1}) \check{C} C_U  B^3 T^{-1} \ln T + \check{C} C_\rho C_{\nu, 3} C_U  L C_{w,1} \frac{1}{1-\rho} B^2  \ln T.
     \label{eq:M_2}
\end{align}

Last, we  consider the case that $(t_1,t_2,t_3,t_4) \in \mathcal{M}_3$.
It follows from \eqref{eq:M_1_part2} that
\begin{align*}
    \biggl|\E_w\biggl[\prod_{j=1}^4 \widetilde{G}_{i}(W_{t_j},\theta_k) \biggr]\biggr| \leq \frac{1}{4}\sum_{j=1}^4 \E_w[|\widetilde{G}_{i}(W_{t_j},\theta_k)|^4]  \leq 16 (L + L C_{\nu,1})^4 C_{w,4}.
\end{align*}
A calculation similar to \eqref{eq:M_2_cardinality}  leads to
$
|\mathcal{M}_3 | \leq B^2 (C_\rho \ln T)^2
$.  Therefore,
\begin{align}
    \sum_{(t_1,t_2,t_3,t_4)\in \mathcal{M}_3}  \biggl|\E_w\biggl[\prod_{j=1}^4\widetilde{G}_{i}(W_{t_j},\theta_k) \biggr] \biggr|
    \leq   16 (L + L C_{\nu,1})^4 C_{w,4} C_\rho^2  B^2 \ln^2 T. \label{eq:M_3}
\end{align}

Combining \eqref{eq:sum-4}, \eqref{eq:M_1}, \eqref{eq:M_2}, and \eqref{eq:M_3} yields
$\E_w\Bigl[\left|\sum_{t=kB}^{(k+1)B-1}\widetilde{G}_i(W_t,\theta_k)\right|^4\Bigr]
 = \mathcal{O}\bigl(\frac{B^4}{T} + B^2  \ln^2 T  \bigr)$.
It then follows from \eqref{eq:4th-moment-bound} that
to prove \eqref{eq:4th-moment-lim}, it suffices to show
\begin{align*}
    \lim_{T\to\infty}( B^{1-\delta} T^\delta)^{2+a} \sum_{K_0 \leq k\leq K} \|Q_{k,K}\|^4 \frac{\alpha_k^4}{ B^{4} }  \left(\frac{B^4}{T} + B^2  \ln^2 T  \right) = 0,
\end{align*}
for some $a>0$.
This can be done by setting $a\in(0, 1-\frac{1}{2\delta})$.

\subsubsection{Verification of Condition~\eqref{eq:MG_CLT_condition2}.}

By definition,
$X^*_k = X_k - \bar{X}_k$ and $\bar{X}_k = \E[X_k | \mathscr{F}_{kB- 1}]$, so
\begin{gather} \label{eq:gap_bet_X*_X}
    \|\E[X_{k}^*(X^*_{k})^\intercal|\mathscr{F}_{kB-1} ]-\E[X_{k}X_{k}^\intercal|\mathscr{F}_{kB-1}]\|
    = \|\bar{X}_k \bar{X}_k^\intercal\| =  \|\bar{X}_k\|^2.
\end{gather}
By \eqref{eq:Psi_4_bound},
$ (B^{1-\delta} T^{\delta})^{\frac{1}{2}} \sum_{k=K_0}^{K-1}  \|\bar{X}_k\| \stackrel{p}{\to} 0$.
Hence,  with probability approaching 1 as $T\to \infty$, we have
$\max_{k=K_0,\ldots,K-1}(B^{1-\delta} T^{\delta})^{\frac{1}{2}} \|\bar{X}_k\| < 1$,
which implies $B^{1-\delta} T^{\delta} \sum_{k=K_0}^{K-1}\|\bar{X}_k\|^2\leq (B^{1-\delta} T^{\delta})^{\frac{1}{2}}  \sum_{k=K_0}^{K-1}\|\bar{X}_k\|$.
Therefore,
$ B^{1-\delta} T^{\delta} \sum_{k=K_0}^{K-1}  \|\bar{X}_k\|^2 \stackrel{p}{\to} 0$ as $T\to\infty$.
It then follows from \eqref{eq:gap_bet_X*_X} that, to verify condition~\eqref{eq:MG_CLT_condition2},
it suffices to show
\begin{equation} \label{eq:MG_CLT_condition2_equiv}
    \lim_{T\to\infty} B^{1-\delta}T^\delta \sum_{k=K_0}^{K-1} X_k X_k^\intercal = \Gamma^* \quad \mbox{in probability}.
\end{equation}

For each $k\geq 0$ and $l< B$, define $\hat{\Sigma}_k(l) \coloneqq B^{-1}\sum_{t=kB}^{(k+1)B-l-1}\widetilde{G}(W_t,\theta_k)\widetilde{G}(W_{t+l},\theta_k)^\intercal $.
It follows from the definition of $\bar{X}_k$ that
\begin{align} \label{eq:X_k-X_k-transpose}
    X_k X_k^{\intercal}  = \frac{\alpha_k^2}{B} Q_{k,K} \biggl(\hat{\Sigma}_k(0)+\sum_{l=1}^{B-1} \bigl(\hat{\Sigma}_k(l)+\hat{\Sigma}_k(l)^\intercal\bigr)\biggr) Q_{k,K}^\intercal.
\end{align}

We now analyze $\E[\hat{\Sigma}_k(l)|\mathscr{F}_{kB-1}]$.
Define
\begin{align*}
   \bar{\Sigma}_k(l)\coloneqq{}&   \E_{\nu_{\theta_k}} [\widetilde{G}(W_t,\theta_k) \widetilde{G}(W_{t+l},\theta_k)^\intercal], \\
   r_{t,k}(l) \coloneqq {}& \E[\widetilde{G}(W_t,\theta_k) \widetilde{G}(W_{t+l},\theta_k)^\intercal|\mathscr{F}_{kB-1}] - \bar{\Sigma}_k(l).
\end{align*}

Note that $\E[ \widetilde{G}(W_{t+l},\theta_k) |\mathscr{F}_t]$ is a function of $W_t$ and we denote it by $h(W_t, l)$.
Then, $\bar{\Sigma}_k(l) = \E_{\nu_{\theta_k}} [\widetilde{G}(W_t,\theta_k) h(W_{t},l)^\intercal] $ and $r_{t,k}(l) =  \E[\widetilde{G}(W_t,\theta_k) h(W_t, l)^\intercal|\mathscr{F}_{kB-1}] - \bar{\Sigma}_k(l) $.
Moreover, let $h_i(W_t, l)$ denote the $i$-th component of $h(W_t, l)$.
By Lemma~\ref{lemma:geo-ergodicity},
\begin{align}
| h_i(W_t, l) |
\leq  |\E[G_i(W_{t+l},\theta_k) | W_{t}, \theta_k] - \bar{G}_i(\theta_k)|
\leq C_U \rho^l L(1 + \|W_t\|). \label{eq:h_norm_bound}
\end{align}
Therefore,
\begin{align}
    \|r_{t,k}(l)\| \leq{}& \sum_{i=1}^d \sum_{j=1}^d  | \E[\widetilde{G}_i(W_t,\theta_k) h_j(W_t, l) |\mathscr{F}_{kB-1}] - \E_{\nu_{\theta_k}}[\widetilde{G}_i(W_t,\theta_k) h_j(W_t, l) ] | \nonumber \\
    \leq{}& 2 d^2 C_U^2  L (L + L  C_{\nu, 1}) \rho^{t-(kB-1)}  \rho^l (1 + \|W_{kB-1}\|^2), \label{eq:bound_r_tk}
\end{align}
where the second inequality follows from \eqref{eq:G_tilde_bound}, \eqref{eq:h_norm_bound}, and  Lemma~\ref{lemma:geo-ergodicity}.
Therefore,
\begin{align*}
    & \E[\|\hat{\Sigma}_k(l) - B^{-1}(B-l) \Sigma^*(l)\| |\mathscr{F}_{kB-1}] \nonumber \\
    \leq{}&  \E[\|\hat{\Sigma}_k(l) - B^{-1}(B-l) \bar{\Sigma}_k(l)\| |\mathscr{F}_{kB-1}]  +B^{-1}(B-l)\|\bar{\Sigma}_k(l)-\Sigma^*(l)\|\nonumber\\
    \leq{}& \frac{1}{B}\sum_{t=kB}^{(k+1)B-l-1}\|r_{t,k}(l) \| + B^{-1}(B-l) L_{\nu} \|\theta_k - \theta^*\|
    \nonumber \\
     \leq{}&  \frac{2 d^2 C_U^2  L (L + L  C_{\nu, 1}) \rho^l}{B(1-\rho)} (1+\|W_{kB-1}\|^2) + L_{\nu} \biggl(\frac{C_{\theta,1}^{\frac{1}{2}}}{B^{\frac{1}{2}} k^{\frac{\delta}{2}}} + \frac{C_{\theta,2}^{\frac{1}{2}}}{k^{\delta}}\biggr),
\end{align*}
where the second inequality follows from Assumption~\ref{assump:ergodic}\ref{asp:ergodic_Lip}, and the third from \eqref{eq:bound_r_tk} and Lemma~\ref{lemma:expectation}.
So,
$\E_w[\|\hat{\Sigma}_k(l) - B^{-1}(B-l) \Sigma^*(l)\| ] \leq C_\Sigma (\rho^l B^{-1} + B^{-\frac{1}{2}} k^{-\frac{\delta}{2}} + k^{-\delta})$ where $C_\Sigma := \max\{2d^2 C_U^2 L (L+L C_{\nu,1})/(B(1-\rho)) C_{w,2}, L_\nu C_{\theta_1}^{\frac{1}{2}}, L_\nu C_{\theta_2}^{\frac{1}{2}}\}.$
Then,
\begin{align*}
    & \E_w\biggl[\biggl\|\sum_{k=K_0}^{K-1}\alpha_k^2 Q_{k,K}\hat{\Sigma}_k(l)Q_{k,K}^\intercal -  \sum_{k=K_0}^{K-1}\alpha_k^2 Q_{k,K}B^{-1}(B-l)\Sigma^*(l)Q_{k,K}^\intercal \biggr\| \biggr]\nonumber\\
    \leq{}& \sum_{k=K_0}^{K-1}\alpha_k^2 \|Q_{k,K}\|^2 \E_w[\| \hat{\Sigma}_k(l)-B^{-1}(B-l)\Sigma^*(l)\|]\nonumber\\
    \leq {}&  C_\Sigma \biggl( \sum_{k=K_0}^K \alpha_k^2 \|Q_{k,K}\|^2 \left(\rho^l B^{-1} + B^{-\frac{1}{2}} k^{-\frac{\delta}{2}} + k^{-\delta}\right) \biggr) \nonumber\\
    \leq {}& C_\Sigma \alpha_0^2 \frac{2}{\psi}\biggl( \rho^l B^{-1} K^{-\delta} + B^{-\frac{1}{2}}K^{-\frac{3\delta}{2}} + K^{-2\delta}\biggr),
\end{align*}
where the last step follows from \eqref{eq:Q_estimate} and  Lemma~\ref{lemma:sum-approx}.
It then follows from \eqref{eq:X_k-X_k-transpose} that
\begin{align*}
   & B^{1-\delta}T^\delta \E_w\biggl[ \biggl\|  \sum_{k=K_0}^{K-1} X_k X_k^\intercal -  \sum_{k=K_0}^{K-1}\frac{\alpha_k^2}{B} Q_{k,K} \biggl( \Sigma^*(0) +  \sum_{l=1}^{B-1} \frac{B-l}{B} (\Sigma^*(l) + \Sigma^*(l)^\intercal )\biggr) Q_{k,K}^\intercal \biggr\|\biggr] \nonumber \\
   \leq {}& B^{1-\delta}T^\delta C_\Sigma \alpha_0^2 \frac{2}{\psi}  \biggl( \frac{1}{B} \sum_{l=0}^{B-1} \left( \rho^l B^{-1} K^{-\delta} + B^{-\frac{1}{2}}K^{-\frac{3\delta}{2}} + K^{-2\delta}\right) \biggr) \nonumber \\
   \leq  {} &  C_\Sigma \alpha_0^2 \frac{2}{\psi} \left( B^{-1}  + B^{\frac{1}{2}}K^{-\frac{\delta}{2}} +  B K^{-\delta}\right) \to 0,
\end{align*}
as $T\to\infty$, because of Lemma~\ref{lemma:Q-sum-lim}.
Therefore, to prove \eqref{eq:MG_CLT_condition2_equiv}, it suffices to show
\begin{align}
\lim_{T\to\infty} B^{1-\delta}T^\delta \sum_{k=K_0}^{K-1}\frac{\alpha_k^2}{B} Q_{k,K} \biggl( \underbrace{\Sigma^*(0) +  \sum_{l=1}^{B-1} \frac{B-l}{B} (\Sigma^*(l) + \Sigma^*(l)^\intercal )}_{\widetilde{\Sigma}^*_B}\biggr) Q_{k,K}^\intercal = \Gamma^*.    \label{eq:MG_CLT_condition2_equiv_2}
\end{align}

Define $\Sigma^*_B \coloneqq     \Sigma^*(0) +  \sum_{l=1}^{B-1}  (\Sigma^*(l) + \Sigma^*(l)^\intercal )$ and
\begin{align*}
    E_{k,K} \coloneqq \left\{
    \begin{array}{ll}
         \displaystyle \exp \left(\alpha_0  K^{-\delta} (K - k) \Lambda^*\right),     & \quad  \mbox{ if } \delta\in(\frac{1}{2},1), \\
    \displaystyle \exp\left(\alpha_0 \ln(K/k) \Lambda^* \right),      &\quad  \mbox{ if } \delta = 1.
    \end{array}
    \right.
\end{align*}
By Assumptions~\ref{assump:G-bar}\ref{asp:SA_nsd} and \ref{assump:G-bar}\ref{asp:SA_taylor}, $\Lambda^* \prec 0$ and its eigenvalues are no greater than $-\psi$.
Hence,
\begin{align}
     \|E_{k,K}\| \leq  \left\{
    \begin{array}{ll}
         \displaystyle \exp \left(-\psi\alpha_0  K^{-\delta} (K - k) \right),     & \quad  \mbox{ if } \delta\in(\frac{1}{2},1), \\
    \displaystyle \exp\left(-\psi \alpha_0 \ln(K/k) \right),      &\quad  \mbox{ if } \delta = 1.
    \end{array}
    \right. \label{eq:E_k_K_bound}
\end{align}

To prove \eqref{eq:MG_CLT_condition2_equiv_2}, it suffices to show
\begin{gather}
    \lim_{T\to\infty} B^{1-\delta}T^\delta \biggl\| \sum_{k=K_0}^{K-1}\frac{\alpha_k^2}{B} \bigl( Q_{k,K} \widetilde{\Sigma}^*_B Q_{k,K}^\intercal - E_{k,K} \widetilde{\Sigma}^*_B E_{k,K}^\intercal \bigr)  \biggr\| = 0, \label{eq:MG_CLT_condition2_equiv_step1} \\
    \lim_{T\to\infty} B^{1-\delta}T^\delta \biggl\| \sum_{k=K_0}^{K-1}\frac{\alpha_k^2}{B} \bigl( E_{k,K}  \widetilde{\Sigma}^*_B E_{k,K}^\intercal -   E_{k,K} \Sigma^*_B E_{k,K}^\intercal  \bigr) \biggr\| = 0, \label{eq:MG_CLT_condition2_equiv_step2} \\
    \lim_{T\to\infty} B^{1-\delta}T^\delta \sum_{k=K_0}^{K-1}\frac{\alpha_k^2}{B} E_{k,K} \Sigma^*_B E_{k,K}^\intercal = \Gamma^*. \label{eq:MG_CLT_condition2_equiv_step3}
\end{gather}

It follows from the fact that $x - \frac{x^2}{2} \leq \ln(1+x) \leq x$ for all $x\in(-1,0)$ that
\begin{align*}
    \exp(\sum_{l=k}^{K-1} \biggl(\alpha_l \Lambda^* - \frac{1}{2} \alpha_l^2 (\Lambda^*)^2\biggr) ) \preceq Q_{k,K} \preceq \exp(\sum_{l=k}^{K-1} \alpha_l \Lambda^*)
\end{align*}
for all $k$ large enough.
A calculation similar to \eqref{eq:Q_estimate'} leads to
\[E_{k,K} \exp(-\frac{1}{2}\alpha_0(2\delta-1)(k^{1-2\delta} - K^{1-2\delta})(\Lambda^*)^2) \preceq  Q_{k,K} \preceq E_{k,K}\] for all $k$ large enough.
Therefore,
for any $c>0$, for $k>\ln T$ and $T$ large enough,
\begin{align} \label{eq:Q-E-gap}
    \|Q_{k,K} - E_{k,K} \| \leq c \|E_{k,K}\|.
\end{align}

For any $l$, by definition,
\begin{align}
     \|\Sigma^*(l)\| \leq{}&  \E_{\nu_{\theta^*}}\bigl[ \| \widetilde{G}(W_t, \theta^*) \| \E[\|\widetilde{G}(W_{t+l}, \theta^*) \| | \mathscr{F}_t] \bigr] \nonumber \\
     \leq{}&  \E_{\nu_{\theta^*}}\bigl[ \| \widetilde{G}(W_t, \theta^*) \| d C_U \rho^l L (1+ \|W_t\|) \bigr] \nonumber \\
     \leq{}& d C_U \rho^l L \E_{\nu_{\theta^*}} \bigl[(L+L C_{\nu,1}) (1+ \|W_t\|)^2 \bigr]
     \leq 2 d C_U L^2 (1 + C_{\nu,1}) C_{\nu,2} \rho^l, \label{eq:Sigma_l_decay}
\end{align}
where the second inequality follows from \eqref{eq:h_norm_bound}, and the third from \eqref{eq:G_tilde_bound}.
So,
\begin{align}
    \|\widetilde{\Sigma}^*_B\| \leq 2 \sum_{l=0}^{B-1} \|\Sigma^*(l)\|
    \leq 4 d C_U L^2 (1 + C_{\nu,1})C_{\nu,2}  \sum_{l=0}^{B-1} \rho^l
    \leq      C_{\Sigma} \label{eq:Sigma_B_bound}
\end{align}
for any $B$,
where $C_{\Sigma} \coloneqq  \frac{ 4 d C_U L^2 (1 + C_{\nu,1})C_{\nu,2}}{1-\rho}$.
To prove \eqref{eq:MG_CLT_condition2_equiv_step1}, note that
\begin{align}
    \bigl\| Q_{k,K} \widetilde{\Sigma}^*_B Q_{k,K}^\intercal - E_{k,K} \widetilde{\Sigma}^*_B E_{k,K}^\intercal \bigr\|
    \leq{}&  \|Q_{k,K} - E_{k,K} \|^2  \| \widetilde{\Sigma}^*_B \| + 2 \|Q_{k,K} - E_{k,K} \| \|E_{k,K}\|  \| \widetilde{\Sigma}^*_B \| \nonumber \\
    \leq{} & ( c^2 + 2c)  C_{\Sigma} \|E_{k,K}\|^2,
\end{align}
for $T$ large enough and $k>\ln T$,
where the last step follows from \eqref{eq:Q-E-gap} and \eqref{eq:Sigma_B_bound}.
Thus,
\begin{align}
    & B^{1-\delta}T^\delta \biggl\| \sum_{k=K_0}^{K-1}\frac{\alpha_k^2}{B} \bigl( Q_{k,K} \widetilde{\Sigma}^*_B Q_{k,K}^\intercal - E_{k,K} \widetilde{\Sigma}^*_B E_{k,K}^\intercal \bigr)  \biggr\| \nonumber  \\
    \leq{}& 2 K^\delta \biggl\| \sum_{k=K_0}^{\lfloor \ln T\rfloor} + \sum_{k=\lfloor \ln T\rfloor+1}^{K-1} \alpha_k^2 \bigl( Q_{k,K} \widetilde{\Sigma}^*_B Q_{k,K}^\intercal - E_{k,K} \widetilde{\Sigma}^*_B E_{k,K}^\intercal \bigr)  \biggr\|  \nonumber \\
    \leq{}& 4  K^\delta \ln T \alpha_0^2 \biggl(\frac{\ln T}{K}\biggr)^{\psi \alpha_0 }  + 2 K^{\delta} C_\Sigma \sum_{k=\lfloor \ln T\rfloor+1}^{K-1} \alpha_k^2 ( c^2 + 2c)  C_{\Sigma} \|E_{k,K}\|^2  \nonumber \\
    \leq{}& 4 C_\Sigma \frac{(\ln T)^{\psi \alpha_0  + 1}}{K^{\psi \alpha_0  - \delta}} + 2 \alpha_0^2 (c^2 + 2c) C_{\Sigma} \mathring{C} \to 2 \alpha_0^2 (c^2 + 2c) C_{\Sigma},\nonumber
\end{align}
as $T\to\infty$,
for some $\mathring{C}>0$, where   the third inequality follows from \eqref{eq:E_k_K_bound} and Lemma~\ref{lemma:sum-approx}.
As $c$ can be arbitrarily small, it must be that \eqref{eq:MG_CLT_condition2_equiv_step1} holds.

To prove \eqref{eq:MG_CLT_condition2_equiv_step2}, note that by the definitions of $\Sigma^*(l)$ and $\Sigma^*(l)$,
\begin{align*}
    & B^{1-\delta}T^\delta \biggl\| \sum_{k=K_0}^{K-1}\frac{\alpha_k^2}{B} \bigl( E_{k,K} \widetilde{\Sigma}^*_B E_{k,K}^\intercal - E_{k,K} \Sigma^*_B E_{k,K}^\intercal \bigr)  \biggr\|
    \leq 2 K^\delta \sum_{k=K_0}^{K-1}\alpha_k^2 \|E_{k,K}\|^2 \sum_{l=1}^{B-1} \frac{2 l}{B}  \|\Sigma^*(l)\| \\
    & \leq 4d C_U L^2(1+C_{\nu,1})C_{\nu,2} \cdot K^\delta \sum_{k=K_0}^K \alpha_k^2 \|E_{k,K}\|^2 \frac{2}{B} \sum_{l=1}^{B-1}  l \rho^l  \to 0,
\end{align*}
as $T\to\infty$, where  the second inequality follows from \eqref{eq:Sigma_l_decay}, and the convergence to zero follows from \eqref{eq:E_k_K_bound} and Lemma~\ref{lemma:sum-approx}.
Last,  \eqref{eq:MG_CLT_condition2_equiv_step3} can be proved using elementary calculus.

\end{appendix}

\end{document}